\documentclass[10pt,final]{article}
\usepackage[english]{babel}
\usepackage{indentfirst}
\usepackage[utf8x]{inputenc}
\usepackage[T1]{fontenc}
\usepackage{amsmath}
\usepackage{amssymb}
\usepackage{makecell}
\usepackage{array}
\usepackage{titlesec}
\usepackage{subcaption}
\usepackage{graphicx}
\usepackage[edges]{forest}
\usepackage{tikz}
\usepackage{pgfkeys}
\usepackage{ulem}
\usepackage[square,sort,comma,numbers]{natbib}
\usepackage{ctable}
\usepackage{multirow}
\usepackage{tcolorbox}
\usepackage{cuted,lipsum}
\usepackage{psfrag}
\usepackage{epsfig}
\usepackage{wrapfig}
\usetikzlibrary{arrows.meta}
\DeclareMathOperator{\E}{\mathbb{E}}
\newcommand{\argmin}{\operatornamewithlimits{argmin}}
\newcommand{\argmax}{\operatornamewithlimits{argmax}}
\newcommand{\ie}{\textit{i}.\textit{e}., }
\newcommand{\eg}{\textit{e}.\textit{g}. }
\usepackage[a4paper,top=3cm,bottom=2cm,left=3cm,right=3cm,marginparwidth=1.75cm]{geometry}

\usepackage[colorinlistoftodos]{todonotes}
\usepackage[colorlinks=true, allcolors=blue]{hyperref}
\titleformat{\paragraph}
{\normalfont\normalsize\bfseries}{\theparagraph}{1em}{}
\titlespacing*{\paragraph}
{0pt}{3.25ex plus 1ex minus .2ex}{1.5ex plus .2ex}

\title{How Generative Adversarial Networks and \\ Their Variants Work: An Overview}
\author{Yongjun Hong, Uiwon Hwang, Jaeyoon Yoo and Sungroh Yoon \\
Department of Electrical and Computer Engineering \\
Seoul National University, Seoul, Korea \\
\texttt{\{yjhong, uiwon.hwang, yjy765, sryoon\}@snu.ac.kr} 
}
\date{}

\begin{document}
\maketitle

\begin{abstract}
Generative Adversarial Networks (GAN) have received wide attention in the machine learning field for their potential to learn high-dimensional, complex real data distribution. Specifically, they do not rely on any assumptions about the distribution and can generate real-like samples from latent space in a simple manner. This powerful property leads GAN to be applied to various applications such as image synthesis, image attribute editing, image translation, domain adaptation and other academic fields. In this paper, we aim to discuss the details of GAN for those readers who are familiar with, but do not comprehend GAN deeply or who wish to view GAN from various perspectives. In addition, we explain how GAN operates and the fundamental meaning of various objective functions that have been suggested recently. We then focus on how the GAN can be combined with an autoencoder framework. Finally, we enumerate the GAN variants that are applied to various tasks and other fields for those who are interested in exploiting GAN for their research.     
\end{abstract}


\section{Introduction} \label{intro}
Recently, in the machine learning field, generative models have become more important and popular because of their applicability in various fields. Their capability to represent complex and high-dimensional data can be utilized in treating images \cite{wang2018perceptual,DiscoGAN,CycleGAN,Unsupervised,SRGAN,3DGAN,DANN}, videos \cite{MocoGAN,Pose-GAN,VGAN}, music generation \cite{SeqGAN,ORGAN,leeseqgan}, natural languages \cite{RANKGAN,VAW-GAN} and other academic domains such as medical images \cite{GANCS,DI2IN,SCAN} and security \cite{SteganographyGAN,SSGAN}. Specifically, generative models are highly useful for image to image translation (See Figure~\ref{fig:cycle}) \cite{CycleGAN,DiscoGAN,DistanceGAN,DualGAN} which transfers images to another specific domain, image super-resolution \cite{SRGAN}, changing some features of an object in an image \cite{conditioncycle,AGEGAN,ICGAN,ViGAN,GeneGAN} and predicting the next frames of a video \cite{MocoGAN,VGAN,Pose-GAN}. In addition, generative models can be the solution for various problems in the machine learning field such as semisupervised learning \cite{CatGAN,triple,ssl-gan,Improved}, which tries to address the lack of labeled data, and domain adaptation \cite{DANN,Unsupervised,ARDA,AUNDA,CyCADA,dirt-t}, which leverages known knowledge for some tasks in other domains where only little information is given.

Formally, a generative model learns to model a real data probability distribution $p_{\mathrm{data}}(x)$ where the data $x$ exists in the $d$-dimensional real space $R^d$, and most generative models, including autoregressive models~\cite{pixelcnn,pixelrnn}, are based on the maximum likelihood principle with a model parametrized by parameters $\theta$. With independent and identically distributed (i.i.d.) training samples $x^i$ where $i \in \{1,2,\cdots n\}$, the likelihood is defined as the product of probabilities that the model gives to each training data: $\prod_{i=1}^n p_{\mathrm{\theta}}(x^i)$ where $p_{\mathrm{\theta}}(x)$ is the probability that the model assigns to $x$. The maximum likelihood principle trains the model to maximize the likelihood that the model follows the real data distribution.

From this point of view, we need to assume a certain form of $p_{\mathrm{\theta}}(x)$ explicitly to estimate the likelihood of the given data and retrieve the samples from the learned model after the training. In this way, some approaches \cite{pixelcnn,pixelrnn,Wavenet} successfully learned the generative model in various fields including speech synthesis. However, while the explicitly defined probability density function brings about computational tractability, it may fail to represent the complexity of real data distribution and learn the high-dimensional data distributions~\cite{dcmam}.

Generative Adversarial Networks (GANs) \cite{GAN} were proposed to solve the disadvantages of other generative models. Instead of maximizing the likelihood, GAN introduces the concept of adversarial learning between the generator and the discriminator. The generator and the discriminator act as adversaries with respect to each other to produce real-like samples. The generator is a continuous, differentiable transformation function mapping a prior distribution $p_z$ from the latent space $\mathcal{Z}$ into the data space $\mathcal{X}$ that tries to fool the discriminator. The discriminator distinguishes its input whether it comes from the real data distribution or the generator. The basic intuition behind the adversarial learning is that as the generator tries to deceive the discriminator which also evolves against the generator, the generator improves. This adversarial process gives GAN notable advantages over the other generative models.

GAN avoids defining $p_{\mathrm{\theta}}(x)$ explicitly, and instead trains the generator using a binary classification of the discriminator. Thus, the generator does not need to follow a certain form of $p_{\mathrm{\theta}}(x)$. In addition, since the generator is a simple, usually deterministic feed-forward network from $\mathcal{Z}$ to $\mathcal{X}$, GAN can sample the generated data in a simple manner unlike other models using the Markov chain \cite{smolensky1986information} in which the sampling is computationally slow and not accurate. Furthermore, GAN can parallelize the generation, which is not possible for other models such as PixelCNN \cite{pixelcnn}, PixelRNN \cite{pixelrnn}, and WaveNet \cite{Wavenet} due to their autoregressive nature.

For these advantages, GAN has been gaining considerable attention, and the desire to use GAN in many fields is growing. In this study, we explain GAN \cite{GOODFELLOWNIPS,GAN} in detail which generates sharper and better real-like samples than the other generative models by adopting two components, the generator and the discriminator. We look into how GAN works theoretically and how GAN has been applied to various applications. 



\begin{figure} [t]
    \centering    
        \includegraphics[width=0.7\textwidth]{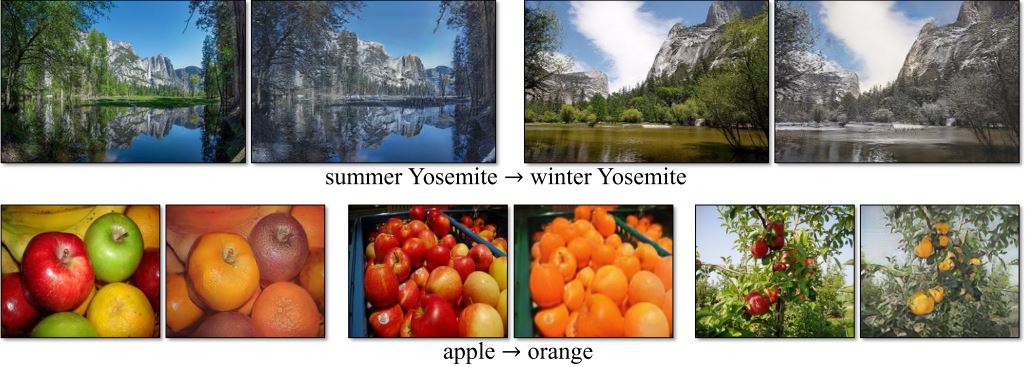}
        \label{fig:cycle}
    \caption{Examples of unpaired image to image translation from CycleGAN \cite{CycleGAN}. CycleGAN use a GAN concept, converting contents of the input image to the desired output image. There are many creative applications using a GAN, including image to image translation, and these will be introduced in Section~\ref{section4}. Images from CycleGAN \cite{CycleGAN}.}\label{fig:cycle}
\end{figure}

\subsection{Paper Organization}

Table~\ref{tab:overview1} shows GAN and GAN variants which will be discussed in Section~\ref{section2} and \ref{section3}. In Section~\ref{section2}, we first present a standard objective function of a GAN and describe how its components work. After that, we present various objective functions proposed recently, focusing on their similarities in terms of the feature matching problem. We then explain the architecture of GAN extending the discussion to dominant obstacles caused by optimizing a minimax problem, especially a mode collapse, and how to address those issues.  

In Section~\ref{section3}, we discuss how GAN can be exploited to learn the latent space where a compressed and low dimensional representation of data lies. In particular, we emphasize how the GAN extracts the latent space from the data space with autoencoder frameworks. Section~\ref{section4} provides several extensions of the GAN applied to other domains and various topics as shown in Table~\ref{tab:overview_app}. In Section~\ref{section5}, we observe a macroscopic view of GAN, especially why GAN is advantageous over other generative models. Finally, Section~\ref{section6} concludes the paper.

\forestset{
  L1/.style={fill=green,edge={black, line width=1pt}},
  L2/.style={fill=orange,edge={orange,line width=2pt}},
  L3/.style={fill=yellow,edge={yellow,line width=2pt}},
  L4/.style={fill=pink,edge={pink,line width=2pt}},
}

\ctable[
	caption = {An overview of GANs discussed in Section~\ref{section2} and \ref{section3}.},
	label = tab:overview1,
	star,
    width = \textwidth,
	doinside = \footnotesize,
	mincapwidth=\textwidth,
]{lll}{
}{
	\toprule
Subject & Topic  & Reference 

                                                        \\                 	
\midrule

 Object    & f-divergence      & GAN \cite{GAN}, f-GAN \cite{f-GAN}, LSGAN \cite{LSGAN}
						 	    \\
  functions                             &  IPM     & WGAN \cite{WGAN}, WGAN-GP \cite{WGAN-GP}, FISHER GAN \cite{FisherGAN}, McGAN \cite{McGAN}, MMDGAN \cite{MMDGAN}  
						 	    \\ \cline{2-3}
                                
 \multirow{3}{*}{Architecture}          &       DCGAN              & DCGAN \cite{DCGAN}
                                   \\ 
                        &  Hierarchy        & StackedGAN \cite{StackedGAN}, GoGAN \cite{GoGAN}, Progressive GAN \cite{karras2017progressive}
                                          \\
                        &       Auto encoder               & BEGAN \cite{BEGAN}, EBGAN \cite{EBGAN}, MAGAN \cite{MAGAN}
                                  \\ \cline{2-3}
                      \multirow{3}{*}{Issues}                                                    &       Theoretical analysis                & Towards principled methods for training GANs \cite{Towards}
                                   \\
                                                                             &                        & Generalization and equilibrium in GAN \cite{MIXGAN}
                                 \\ 
                                                                                                                                  &       Mode collapse                & MRGAN \cite{MRGAN}, DRAGAN \cite{DRAGAN}, MAD-GAN \cite{MAD-GAN}, Unrolled GAN \cite{UnrolledGAN}                                    \\
           \midrule

\multirow{4}{*}{Latent space} & Decomposition & CGAN \cite{CGAN}, ACGAN \cite{AC-GAN}, InfoGAN \cite{InfoGAN}, ss-InfoGAN \cite{ss-InfoGAN}
                                    \\ \cline{2-3} 
                                    
            &   Encoder      & ALI \cite{ALI}, BiGAN \cite{BiGAN}, Adversarial Generator-Encoder Networks \cite{AGE}
                                    \\ \cline{2-3} 
                        & VAE        & VAEGAN \cite{VAEGAN}, $\alpha$-GAN \cite{alpha-GAN}
                        		\\  
	\bottomrule
\vspace{-6mm}
}

\ctable[
	caption = {Categorization of GANs applied for various topics.},
	label = tab:overview_app,
	star,
    width = \textwidth,
    pos = ht,
	doinside = \footnotesize,
    mincapwidth=\textwidth,
]{lll}{
}{
	\toprule
Domain & Topic  & Reference 

                                                        \\                            	
\midrule

\multirow{10}{*}{Image}  & Image translation      & Pix2pix \cite{pix2pix}, PAN \cite{wang2018perceptual}, CycleGAN \cite{CycleGAN}, DiscoGAN \cite{DiscoGAN} 
						 	    \\  

                        &        Super resolution     & SRGAN \cite{SRGAN}
                                           \\ 
                        &       Object detection                  & SeGAN \cite{SeGAN}, Perceptual GAN for small object detection \cite{Perceptual}
                                   \\ 
                        &  Object transfiguration        & GeneGAN \cite{GeneGAN}, GP-GAN \cite{GP-GAN}
                                  \\ 
                                                                             &       Joint image generation                 & Coupled GAN \cite{CoupledGAN}
                                   \\ 
                                                                             &       Video generation                 & VGAN \cite{VGAN}, Pose-GAN \cite{Pose-GAN}, MoCoGAN \cite{MocoGAN}
                                 \\ 
                                                                                                                                  &       Text to image                 & Stack GAN \cite{StackedGAN}, TAC-GAN \cite{TAC-GAN}
                                    \\ 
                                        &      Change facial attributes                & SD-GAN \cite{SD-GAN}, SL-GAN \cite{SL-GAN}, DR-GAN \cite{DR-GAN}, AGEGAN \cite{AGEGAN}
                                    \\ 
           \midrule

\multirow{3}{*}{Sequential data} & Music generation  & C-RNN-GAN \cite{C-RNN-GAN}, SeqGAN \cite{SeqGAN}, ORGAN \cite{ORGAN}
						 				  \\                   
            &      Text generation          & RankGAN \cite{RANKGAN}
                                    \\ 
                        & Speech conversion        & VAW-GAN \cite{VAW-GAN}
                        		\\ \midrule       
                                                     
 & Semi-supervised learning      & SSL-GAN \cite{Improved}, CatGAN \cite{CatGAN}, Triple-GAN \cite{triple}
						 		 \\ 
\multirow{4}{*}{Others}              & Domain adaptation        & DANN \cite{DANN}, CyCADA \cite{CyCADA}
                        		  \\ 
                                                 &     & Unsupervised pixel-level domain adaptation \cite{Unsupervised}
                        		  \\  

                        & Continual learning       & Deep generative replay \cite{Continual} 
                                 \\
                                                     
  & Medical image segmentation    & DI2IN \cite{DI2IN}, SCAN \cite{SCAN}, SegAN \cite{Medical} 						   			  \\  
                        &     Steganography        & Steganography GAN \cite{SteganographyGAN}, Secure steganography GAN \cite{SSGAN}
                        	   \\ 
	\bottomrule
}

\section{Generative Adversarial Networks} \label{section2}
As its name implies, GAN is a generative model that learns to make real-like data adversarially \cite{GAN} containing two components, the generator $G$ and the discriminator $D$. $G$ takes the role of producing real-like fake samples from the latent variable $z$, whereas $D$ determines whether its input comes from $G$ or real data space. $D$ outputs a high value as it determines that its input is more likely to be real. $G$ and $D$ compete with each other to achieve their individual goals, thus generating the term adversarial. This adversarial learning situation can be formulated as Equation~\ref{eq:gan} with parametrized networks $G$ and $D$. $p_{\mathrm{data}}(x)$ and $p_{z}(z)$ in Equation~\ref{eq:gan} denote the real data probability distribution defined in the data space $\mathcal{X}$ and the probability distribution of $z$ defined on the latent space $\mathcal{Z}$, respectively.
\begin{equation} \label{eq:gan}
\min_G \max_D V(G,D) = \min_G \max_D \E_{x\sim p_{\mathrm{data}}}\left[\log D(x)\right] + \E_{z\sim p_{z}}\left[\log (1-D(G(z))\right].
\end{equation}
$V(G,D)$ is a binary cross entropy function that is commonly used in binary classification problems \cite{LSGAN}. Note that $G$ maps $z$ from $\mathcal{Z}$ into the element of $\mathcal{X}$, whereas $D$ takes an input $x$ and distinguishes whether $x$ is a real sample or a fake sample generated by $G$.

\begin{figure} [t]
\centering
\includegraphics[width=0.7\textwidth]{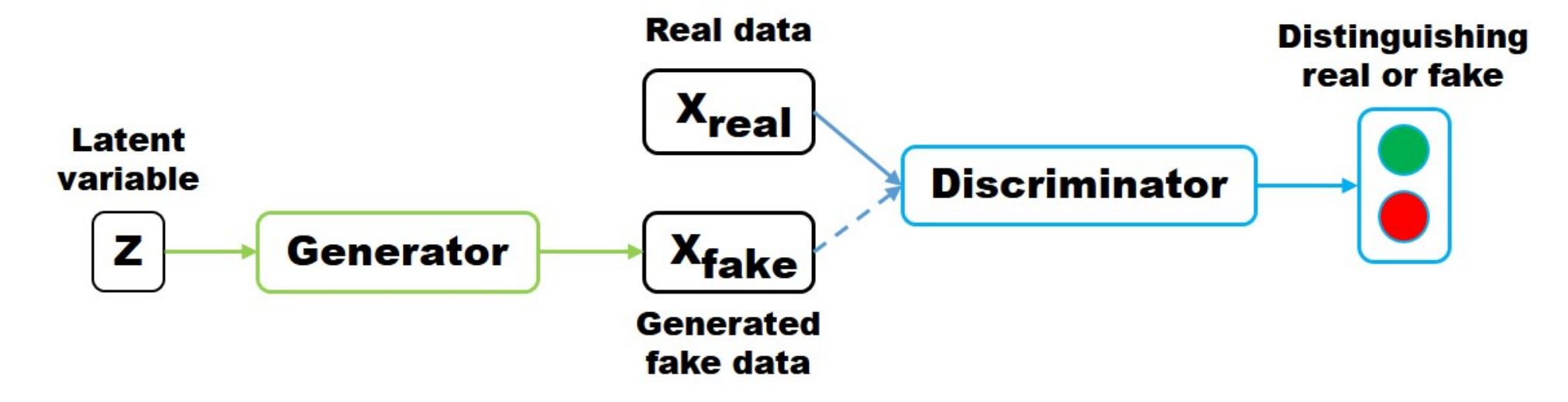}
\caption{\label{fig:ganoriginal}Generative Adversarial Network}
\end{figure}

As $D$ wants to classify real or fake samples, $V(G,D)$ is a natural choice for an objective function in aspect of the classification problem. From $D$'s perspective, if a sample comes from real data, $D$ will maximize its output, while if a sample comes from $G$, $D$ will minimize its output; thus, the $\log(1-D(G(z)))$ term appears in Equation~\ref{eq:gan}. 
Simultaneously, $G$ wants to deceive $D$, so it tries to maximize $D$'s output when a fake sample is presented to $D$. Consequently, $D$ tries to maximize $V(G,D)$ while $G$ tries to minimize $V(G,D)$, thus forming the minimax relationship in Equation~\ref{eq:gan}. Figure~\ref{fig:ganoriginal} shows an illustration of the GAN.  

Theoretically, assuming that the two models $G$ and $D$ both have sufficient capacity, the equilibrium between $G$ and $D$ occurs when $p_\mathrm{data}(x)=p_g(x)$ and $D$ always produces $\frac{1}{2}$ where $p_g(x)$ means a probability distribution of the data provided by the generator \cite{GAN}. Formally, for fixed $G$ the optimal discriminator $D^\star$ is $D^\star(x) = \frac{p_\mathrm{g}(x)}{p_\mathrm{g}(x)+p_{\mathrm{data}}(x)}$ which can be shown by differentiating Equation~\ref{eq:gan}. If we plug in the optimal $D^\star$ into Equation~\ref{eq:gan}, the equation becomes the Jensen Shannon Divergence (JSD) between $p_{\mathrm{data}}(x)$ and $p_{\mathrm{g}}(x)$. Thus, the optimal generator minimizing JSD$(p_{\mathrm{data}}||p_{\mathrm{g}})$ is the data distribution $p_\mathrm{data}(x)$ and $D$ becomes $\frac{1}{2}$ by substituting the optimal generator into the optimal $D^\star$ term. 

Beyond the theoretical support of GAN, the above paragraph leads us to infer two points. First, from the optimal discriminator in the above, GAN can be connected into the density ratio trick \cite{alpha-GAN}. That is, the density ratio between the data distribution and the generated data distribution as follows:
\begin{equation} \label{eq:densityratio}
Dr(x)=\frac{p_{\mathrm{data}}(x)}{p_{\mathrm{g}}(x)}=\frac{p(x|y=1)}{p(x|y=0)}=\frac{p(y=1|x)}{p(y=0|x)}=\frac{D^\star(x)}{1-D^\star(x)}
\end{equation}
where $y=0$ and $y=1$ indicate the generated data and the real data, respectively and $p(y=1)=p(y=0)$ is assumed. This means that GAN addresses the intractability of the likelihood by just using the relative behavior of the two distributions~\cite{alpha-GAN}, and transferring this information to the generator to produce real-like samples. Second, GAN can be interpreted to measure the discrepancy between the generated data distribution and the real data distribution and then learn to reduce it. The discriminator is used to implicitly measure the discrepancy.

Despite the advantage and theoretical support of GAN, many shortcomings has been found due to the practical issues and inability to implement the assumption in theory including the infinite capacity of the discriminator. There have been many attempts to solve these issues by changing the objective function, the architecture, etc. Holding the fundamental framework of GAN, we assess variants of the object function and the architectures proposed for the development of GAN. We then focus on the crucial failures of GAN and how to address those issues.

\subsection{Object Functions} \label{objective function}
The goal of generative models is to match the real data distribution $p_{\mathrm{data}}(x)$ from $p_{\mathrm{g}}(x)$. Thus, minimizing differences between two distributions is a crucial point for training generative models. As mentioned above, standard GAN \cite{GAN} minimizes JSD$(p_{\mathrm{data}}||p_{\mathrm{g}})$ estimated by using the discriminator. Recently, researchers have found that various distances or divergence measures can be adopted instead of JSD and can improve the performance of the GAN. In this section, we discuss how to measure the discrepancy between $p_{\mathrm{data}}(x)$ and $p_{\mathrm{g}}(x)$ using various distances and object functions derived from these distances.
\subsubsection{f-divergence} \label{sec:f}

The f-divergence $D_f(p_{\mathrm{data}}||p_{\mathrm{g}})$ is one of the means to measure differences between two distributions with a specific convex function $f$. Using the ratio of the two distributions, the f-divergence for $p_{\mathrm{data}}$ and $p_{\mathrm{g}}$ with a function $f$ is defined as follows: 
\begin{equation} \label{eq:fganoriginal}
D_f(p_{\mathrm{data}}||p_{\mathrm{g}}) = \int_{\mathcal{X}}p_{\mathrm{g}}(x)f\left(\frac{p_{\mathrm{data}}(x)}{p_{\mathrm{g}}(x)}\right)dx  
\end{equation}
It should be noted that $D_f(p_{\mathrm{data}}||p_{\mathrm{g}})$ can act as a divergence between two distributions under the conditions that $f$ is a convex function and $f(1)=0$ is satisfied. Because of the condition $f(1)=0$, if two distributions are equivalent, their ratio becomes 1 and their divergence goes to 0. Though $f$ is termed a generator function \cite{f-GAN} in general, we call $f$ an f-divergence function to avoid confusion with the generator $G$.

f-GAN \cite{f-GAN} generalizes the GAN objective function in terms of f-divergence under an arbitrary convex function $f$. As we do not know the distributions exactly, Equation~\ref{eq:fganoriginal} should be estimated through a tractable form such as an expectation form. By using the convex conjugate $f(u)=\sup_{t\in dom f^{\star}}(tu - f^{\star}(t))$, Equation~\ref{eq:fganoriginal} can be reformulated as follows: 
\begin{align} 
D_f(p_{\mathrm{data}}||p_{\mathrm{g}}) &= \int_{\mathcal{X}}p_{\mathrm{g}}(x) \sup_{t\in dom f^{\star}}\left(t\frac{p_{\mathrm{data}}(x)}{p_{\mathrm{g}}(x)} - f^{\star}(t)\right)dx \label{fganfirst}\\
&\geq \sup_{T \in \mathcal{T}} \left(\int_{\mathcal{X}}T(x)p_{\mathrm{data}}(x) - f^{\star}\left(T(x)\right)p_{\mathrm{g}}(x)\right)dx \label{fgansecond}\\
&= \sup_{T \in \mathcal{T}}\left(\E_{x\sim p_{\mathrm{data}}}[T(x)] - \E_{x\sim p_{\mathrm{g}}}[f^{\star}(T(x))]\right) \label{fganlast}
\end{align}
where $f^{\star}$ is a Fenchel conjugate \cite{fenchel} of a convex function $f$ and $dom f^{\star}$ indicates a domain of $f^{\star}$.

Equation~\ref{fgansecond} follows from the fact that the summation of the maximum is larger than the maximum of the summation and $\mathcal{T}$ is an arbitrary function class that satisfies $\mathcal{X}\rightarrow \mathbb{R}$. Note that we replace $t$ in Equation~\ref{fganfirst} by $T(x): \mathcal{X} \rightarrow dom f^{\star}$ in Equation~\ref{fgansecond} to make $t$ involved in $\int_{\mathcal{X}}$. If we express $T(x)$ in the form of $T(x)=a(D_\omega(x))$ with $a(\cdot): \mathbb{R}\rightarrow dom f^{\star}$ and $D_\omega(x): \mathcal{X}\rightarrow \mathbb{R}$, we can interpret $T(x)$ as the parameterized discriminator with a specific activation function $a(\cdot)$.

We can then create various GAN frameworks with the specified generator function $f$ and activation function $a$ using the parameterized generator $G_\theta$ and discriminator $\mathcal{T}_\omega$. Similar to the standard GAN \cite{GAN}, f-GAN first maximizes the lower bound in Equation~\ref{fganlast} with respect to $T_\omega$ to make the lower bound tight to $D_f(p_{\mathrm{data}}||p_{\mathrm{g}})$, and then minimizes the approximated divergence with respect to $G_\theta$ to make $p_{\mathrm{g}}(x)$ similar to $p_{data}(x)$. In this manner, f-GAN tries to generalize various GAN objectives by estimating some types of divergences given $f$ as shown in Table~\ref{tab:f-divergence}

\begin{table} [t]
\centering
\caption{\label{tab:f-divergence}GANs using f-divergence. Table reproduced from \cite{f-GAN}.}
\begin{tabular}{c|c|c}
GAN & Divergence & Generator $f(t)$ \tabularnewline
\Xhline{3\arrayrulewidth}
& KLD & $t\log t$\\
GAN \cite{GAN} & JSD - $2\log$2 & $t\log t-(t+1)\log(t+1)$ \\
LSGAN \cite{LSGAN} & Pearson $\mathcal{X}^2$ & $(t-1)^2$ \\
EBGAN \cite{EBGAN} & Total Variance & $|t-1|$ \\
\end{tabular}
\vspace{-4mm}
\end{table}

Kullback-Leibler Divergence (KLD), reverse KLD, JSD and other divergences can be derived using the f-GAN framework with the specific generator function $f$, though they are not all represented in Table~\ref{tab:f-divergence}. Among f-GAN based GANs, LSGAN is one of the most widely used GANs due to its simplicity and high performance; we briefly explain LSGAN in the next paragraph. To summarize, f-divergence $D_f(p_{\mathrm{data}}||p_{\mathrm{g}})$ in Equation~\ref{eq:fganoriginal} can be indirectly estimated by calculating expectations of its lower bound to deal with the intractable form with the unknown probability distributions. f-GAN generalizes various divergences under an f-divergence framework and thus it can derive the corresponding GAN objective with respect to a specific divergence.

\paragraph{Least square GAN}
The standard GAN uses a sigmoid cross entropy loss for the discriminator to classify whether its input is real or fake. However, if a generated sample is well classified as real by the discriminator, there would be no reason for the generator to be updated even though the generated sample is located far from the real data distribution. A sigmoid cross entropy loss can barely push such generated samples towards real data distribution since its classification role has been achieved.

Motivated by this phenomenon, least-square GAN (LSGAN) replaces a sigmoid cross entropy loss with a least square loss, which directly penalizes fake samples by moving them close to the real data distribution. Compared to Equation~\ref{eq:gan}, LSGAN solves the following problems:
\begin{align} \label{eq:lsgan}
\min_D V_{LSGAN}(D) &= \min_D \frac{1}{2}\E_{x\sim p_{\mathrm{data}}}[(D(x)-b)^2] + \E_{z\sim p_{z}}[(D(G(z))-a)^2] \\ \label{eq:lsgan1}
\min_{G} V_{LSGAN}(G) &= \min_G \frac{1}{2}\E_{z\sim p_z}[(D(G(z))-c)^2]
\end{align}
where $a,b$ and $c$ refer to the baseline values for the discriminator.

Equations~\ref{eq:lsgan} and \ref{eq:lsgan1} use a least square loss, under which the discriminator is forced to have designated values ($a,b$ and $c$) for the real samples and the generated samples, respectively, rather than a probability for the real or fake samples. Thus, in contrary to a sigmoid cross entropy loss, a least square loss not only classifies the real samples and the generated samples but also pushes generated samples closer to the real data distribution. In addition, LSGAN can be connected to an f-divergence framework as shown in Table~\ref{tab:f-divergence}.

\subsubsection{Integral probability metric} \label{sec:ipm}

Integral probability metric (IPM) defines a critic function $f$, which belongs to a specific function class $\mathcal{F}$, and IPM is defined as a maximal measure between two arbitrary distributions under the frame of $f$. In a compact space $\mathcal{X} \subset R^d$, let $\mathcal{P}(\mathcal{X})$ denote the probability measures defined on $\mathcal{X}$. IPM metrics between two distributions $p_{\mathrm{data}}$, $p_{\mathrm{g}} \in \mathcal{P}(\mathcal{X})$ is defined as follows:
\begin{equation} \label{eq:IPM}
d_{\mathcal{F}}(p_{\mathrm{data}}, p_{\mathrm{g}}) =sup_{f \in \mathcal{F}} \;\E_{x\sim p_{\mathrm{data}}}[f(x)] - \E_{x\sim p_{\mathrm{g}}}[f(x)]
\end{equation}

As shown in Equation~\ref{eq:IPM}, IPM metric $d_\mathcal{F}(p_{\mathrm{data}}, p_{\mathrm{g}})$ defined on $\mathcal{X}$ determines a maximal distance between $p_{\mathrm{data}}(x)$ and $p_{\mathrm{g}}(x)$ with functions belonging to $\mathcal{F}$ which is a set of measurable, bounded, real-valued functions. It should be noted that $\mathcal{F}$ determines various distances and their properties. Here, we consider the function class $\mathcal{F}_{v,w}$ whose elements are the critic $f$, which can be represented as a standard inner product of parameterized neural networks $\Phi_w$ and a linear output activation function $v$ in real space as described in Equation~\ref{eq:featurematching}. $w$ belongs to parameter space $\Omega$ that forces the function space to be bounded. Under the function class in Equation~\ref{eq:featurematching}, we can reformulate Equation~\ref{eq:IPM} as the following equations:
\begin{equation} \label{eq:featurematching}
\mathcal{F}_{v,w} = \{{f(x)=<v, \Phi_w(x)>|v\in R^m, \Phi_w(x):\mathcal{X}\rightarrow R^m}\}
\end{equation}
\begin{align} 
d_{\mathcal{F}_{v,w}}(p_{\mathrm{data}}, p_{\mathrm{g}}) &= sup_{f\in \mathcal{F}_{v,w}}\E_{x\sim p_{\mathrm{data}}}f(x)-\E_{x\sim p_{\mathrm{g}}}f(x) \label{ipfirst} \\
&= \max_{w\in\Omega, v}<v, \E_{x\sim p_{\mathrm{data}}}\Phi_w(x)-\E_{x\sim p_{\mathrm{g}}}\Phi_w(x)> \label{ipsecond} \\
&= \max_{w\in\Omega}\max_{v}<v, \E_{x\sim p_{\mathrm{data}}}\Phi_w(x)-\E_{x\sim p_{\mathrm{g}}}\Phi_w(x)> \label{iplast}
\end{align}

In Equation~\ref{iplast}, the range of $v$ determines the semantic meanings of the corresponding IPM metrics. From now on, we discuss IPM metric variants such as the Wasserstein metric, maximum mean discrepancy (MMD), and the Fisher metric based on Equation~\ref{iplast}.

\paragraph{Wasserstein GAN} \label{wagan}
Wasserstein GAN (WGAN) \cite{WGAN} presents significant studies regarding the distance between $p_{\mathrm{data}}(x)$ and $p_{\mathrm{g}}(x)$. GAN learns the generator function $g_\theta$ that transforms a latent variable $z$ into $p_{\mathrm{g}}(x)$ rather than directly learning the probability distribution $p_{\mathrm{data}}(x)$ itself. A measure between $p_{\mathrm{g}}(x)$ and $p_{\mathrm{data}}(x)$ thus is required to train $g_\theta$. WGAN suggests the Earth-mover (EM) distance which is also called the Wasserstein distance, as a measure of the discrepancy between the two distributions. The Wasserstein distance is defined as follows: 
\begin{equation} \label{eq:wasserstein}
W(p_{\mathrm{data}}, p_{\mathrm{g}}) = inf_{\gamma\in\sqcap(p_{\mathrm{data}}, p_{\mathrm{g}})} \E_{(x,y) \sim \gamma}\left[\Vert x-y\Vert\right]
\end{equation}
where $\sqcap(p_{\mathrm{data}}, p_{\mathrm{g}})$ denotes the set of all joint distributions where the marginals of $\gamma(x,y)$ are $p_{\mathrm{data}}(x)$ and $p_{\mathrm{g}}(x)$ respectively. 

Probability distributions can be interpreted as the amount of mass they place at each point, and EM distance is the minimum total amount of work required to transform $p_{\mathrm{data}}(x)$ into $p_{\mathrm{g}}(x)$. From this view, calculating the EM distance is equivalent to finding a transport plan $\gamma(x, y)$, which defines how we distribute the amount of mass from $p_{\mathrm{data}}(x)$ over $p_{\mathrm{g}}(y)$. 
Therefore, a marginality condition can be interpreted that $p_{\mathrm{data}}(x)=\int_{y}\gamma(x,y)dy$ is the amount of mass to move from point $x$ and $p_{\mathrm{g}}(y)=\int_{x}\gamma(x,y)dx$ is the amount of mass to be stacked at the point $y$. Because work is defined as the amount of mass times the distance it moves, we have to multiply the Euclidean distance $\Vert x-y\Vert$ by $\gamma(x,y)$ at each point $x$, $y$ and the minimum amount of work is derived as Equation~\ref{eq:wasserstein}.

The benefit of the EM distance over other metrics is that it is a more sensible objective function when learning distributions with the support of a low-dimensional manifold. The article on WGAN shows that EM distance is the weakest convergent metric in that the convergent sequence under the EM distance does not converge under other metrics and it is continuous and differentiable almost everywhere under the Lipschitz condition, which standard feed-forward neural networks satisfy. Thus, EM distance results in a more tolerable measure than do other distances such as KLD and total variance distance regarding convergence of the distance.

As the $\inf$ term in Equation~\ref{eq:wasserstein} is intractable, it is converted into a tractable equation via Kantorovich-Rubinstein duality with the Lipschitz function class \cite{kr}, \cite{kantorovich}; \ie $f:X\rightarrow R$, satisfying $d_R(f(x_1), f(x_2)) \leq 1\times d_X(x_1, x_2)$, $\forall x_1,x_2 \in X$ where $d_X$ denotes the distance metric in the domain $X$. A duality of Equation~\ref{eq:wasserstein} is as follows:
\begin{equation} \label{eq:kantro}
W(p_{\mathrm{data}}, p_{\mathrm{g}}) = sup_{|f|_{L}\leq 1}\E_{x\sim p_{\mathrm{data}}}[f(x)] - \E_{x\sim p_{\mathrm{g}}}[f(x)]
\end{equation}
Consequently, if we parametrize the critic $f$ with $w$ to be a 1-Lipschitz function, the formulation becomes to a minimax problem in that we train $f_w$ first to approximate $W(p_{\mathrm{data}}, p_{\mathrm{g}})$ by searching for the maximum as in Equation~\ref{eq:kantro}, and minimize such approximated distance by optimizing the generator $g_\theta$. To guarantee that $f_w$ is a Lipschitz function, weight clipping is conducted for every update of $w$ to ensure that the parameter space of $w$ lies in a compact space. It should be noted that $f(x)$ is called the critic because it does not explicitly classify its inputs as the discriminator, but rather scores its input.

\paragraph{Variants of WGAN} \label{wgangp}
WGAN with gradient penalty (WGAN-GP) \cite{WGAN-GP} points out that the weight clipping for the critic while training WGAN incurs a pathological behavior of the discriminator and suggests adding a penalizing term of the gradient's norm instead of the weight clipping. It shows that guaranteeing the Lipschitz condition for the critic via weight clipping constraints the critic in a very limited subset of all Lipschitz functions; this biases the critic toward a simple function. The weight clipping also creates a gradient problem as it pushes weights to the extremes of the clipping range. Instead of the weight clipping, adding a gradient penalty term to Equation~\ref{eq:kantro} for the purpose of implementing the Lipschitz condition by directly constraining the gradient of the critic has been suggested~\cite{WGAN-GP}.

Loss sensitive GAN (LS-GAN) \cite{LS-GAN} also uses a Lipschitz constraint but with a different method. It learns loss function $L_\theta$ instead of the critic such that the loss of a real sample should be smaller than a generated sample by a data-dependent margin, leading to more focus on fake samples whose margin is high. Moreover, LS-GAN assumes that the density of real samples $p_{\mathrm{data}}(x)$ is Lipschitz continuous so that nearby data do not abruptly change. The reason for adopting the Lipschitz condition is independent of WGAN's Lipschitz condition. The article on LS-GAN discusses that the nonparametric assumption that the model should have infinite capacity proposed by \citet{GAN} is too harsh a condition to satisfy even for deep neural networks and causes various problems in training; hence, it constrains a model to lie in Lipschitz continuous function space while WGAN's Lipschitz condition comes from the Kantorovich-Rubinstein duality and only the critic is constrained. In addition, LS-GAN uses a weight-decay regularization technique to impose the weights of a model to lie in a bounded area to ensure the Lipschitz function condition.


\paragraph{Mean feature matching}

From Equation~\ref{eq:featurematching}, we can generalize several IPM metrics under the measure of the inner product. If we constrain $v$ with $p$ norm where $p$ is a nonnegative integer, we can derive Equation~\ref{qnormlast} as a feature matching problem as follows by adding the $\Vert v\Vert_p\leq 1$ condition where $\Vert v\Vert_p=\{\Sigma_{i=1}^{m}v_i^p\}^{1/p}$. It should be noted that, with conjugate exponent $q$ of $p$ such that $\frac{1}{p}+\frac{1}{q}=1$, the dual norm of norm $p$ satisfies $\Vert x\Vert_q=sup\{{<v,x>:\Vert v\Vert_p\leq 1}\}$ by Holder's inequality \cite{holder}. Motivated by this dual norm property \cite{McGAN}, we can derive a $l_q$ mean matching problem as follows:   
\begin{align} 
d_{\mathcal{F}}(p_{\mathrm{data}}, p_{\mathrm{g}}) &= \max_{w\in\Omega}\max_{\Vert v\Vert_p\leq 1}<v, \E_{x\sim p_{\mathrm{data}}}\Phi_w(x)-\E_{x\sim p_{\mathrm{g}}}\Phi_w(x)> \label{qnormfirst} \\
&= \max_{w\in\Omega}\Vert\E_{x\sim p_{\mathrm{data}}}\Phi_w(x)-\E_{x\sim p_{\mathrm{g}}}\Phi_w(x)\Vert_q \label{qnormsecond} \\
&= \max_{w\in\Omega}\Vert\mu_w(p_{\mathrm{data}})-\mu_w(p_{\mathrm{g}})\Vert_q  \label{qnormlast}
\end{align}
where $\E_{x\sim \mathcal{P}}\Phi_w(x)=\mu_w(\mathcal{P})$ denotes an embedding mean from distribution $\mathcal{P}$ represented by a neural network $\Phi_w$.

In terms of WGAN, WGAN uses the 1-Lipschitz function class condition by the weight clipping. The weight clipping indicates that the infinite norm $\Vert v\Vert_\infty=max_i |v_i|$ is constrained. Thus, WGAN can be interpreted as an $l_1$ mean feature matching problem. Mean and covariance feature matching GAN (McGAN) \cite{McGAN} extended this concept to match not only the $l_q$ mean feature but also the second order moment feature by using the singular value decomposition concept; it aims to also maximize an embedding covariance discrepancy between $p_{\mathrm{data}}(x)$ and $p_{\mathrm{g}}(x)$. Geometric GAN \cite{GeometricGAN} shows that the McGAN framework is equivalent to a support vector machine (SVM) \cite{SVM}, which separates the two distributions with a hyperplane that maximizes the margin. It encourages the discriminator to move away from the separating hyperplane and the generator to move toward the separating hyperplane. However, such high-order moment matching requires complex matrix computations. MMD addresses this problem with a kernel technique which induces an approximation of high-order moments and can be analyzed in a feature matching framework. 

\paragraph{Maximum mean discrepancy (MMD)}

Before describing MMD methodology, we must first list some mathematical facts. A Hilbert space $\mathcal{H}$ is a complete vector space with the metric endowed by the inner product in the space. Kernel $k$ is defined as $k:\mathcal{X}\times\mathcal{X}\rightarrow \mathbb{R}$ such that $k(y,x)=k(x,y)$. Then, for any given positive definite kernel $k(\cdot,\cdot)$, there exists a unique space of functions $f:\mathcal{X}\rightarrow \mathbb{R}$ called the reproducing kernel Hilbert space (RKHS), $\mathcal{H}_k$ which is a Hilbert space and satisfies $<f,k(\cdot,x)>_\mathcal{H_K}=f(x),\,\forall x \in \mathcal{X}, \forall f \in \mathcal{H}_k$ (so called reproducing property).

MMD methodology can be seen as feature matching under the RKHS space for some given kernel with an additional function class restriction such that $\mathcal{F}=\{{f|\Vert f\Vert_{\mathcal{H}_k}}\leq 1\}$. It can be related to Equation~\ref{eq:featurematching} in that the RKHS space is a completion of $\{f|f(x)=\Sigma_{i=1}^n a_i\Phi_{x_i}(x), a_i\in\mathbb{R},x_i\in\mathcal{X}\}$ where $\Phi_{x_i}(x)=k(x,x_i)$. Therefore, we can formulate MMD methodology as another mean feature matching from Equation~\ref{ipfirst} as follows:
\begin{align} 
d_\mathcal{F}(p_{\mathrm{data}}, p_{\mathrm{g}}) &= \sup_{\Vert f\Vert_{\mathcal{H}_k}\leq 1}{\E_{x\sim p_{\mathrm{data}}}f(x)-\E_{x\sim p_{\mathrm{g}}}f(x)} \label{mmdfirst} \\
&= \sup_{\Vert f\Vert_{\mathcal{H}_k}\leq 1}\{<f, \E_{x\sim p_{\mathrm{data}}}\Phi_x(\cdot)-\E_{x\sim p_{\mathrm{g}}}\Phi_x(\cdot)>_{\mathcal{H}_k}\} \label{mmdsecond} \\
&= \Vert\mu(p_{\mathrm{data}})-\mu(p_{\mathrm{g}})\Vert_{\mathcal{H}_k}
\label{mmdlast}
\end{align}
where $\mu(\mathcal{P})=\E_{x\sim\mathcal{P}}\Phi_x(\cdot)$ denotes a kernel embedding mean and Equation~\ref{mmdsecond} comes from the reproducing property. 

From Equation~\ref{mmdlast}, $d_\mathcal{F}$ is defined as the maximum kernel mean discrepancy in RKHS. It is widely used in statistical tasks, such as a two sample test to measure a dissimilarity. Given $p_{\mathrm{data}}(x)$ with mean $\mu_{p_{\mathrm{data}}}$, $\,p_{\mathrm{g}}(x)$ with mean $\mu_{p_{\mathrm{g}}}$ and kernel $k$, the square of the MMD distance $M_k(p_{\mathrm{data}}, p_{\mathrm{g}})$ can be reformulated as follows:
\begin{equation}
M_k(p_{\mathrm{data}}, p_{\mathrm{g}}) = {\Vert \mu_{p_{\mathrm{data}}} - \mu_{p_{\mathrm{g}}} \Vert}_{\mathcal{H}_k}^2 = \E_{x,x' \sim p_{\mathrm{data}}}[k(x,x')] - 2\E_{x\sim p_{\mathrm{data}},y\sim p_{\mathrm{g}}}[k(x,y)]+\E_{y,y'\sim p_{\mathrm{g}}}[k(y,y')] 
\end{equation}

Generative moment matching networks (GMMN) \cite{GMMN} suggest directly minimizing MMD distance with a fixed Gaussian kernel $k(x,x')=exp(-{\Vert x-x'\Vert}^2)$ by optimizing $\min_\theta M_k(p_{\mathrm{data}}, p_{\mathrm{g}})$. It appears quite dissimilar to the standard GAN \cite{GAN} because there is no discriminator that estimates the discrepancy between two distributions. Unlike GMMN, MMDGAN \cite{MMDGAN} suggests adversarial kernel learning by not fixing the kernel but learning it itself. They replace the fixed Gaussian kernel with a composition of a Gaussian kernel and an injective function $f_\phi$ as follows: $\tilde{k}(x,x') = exp(-{\Vert f_\phi(x)-f_\phi(x')\Vert}^2)$ and learn a kernel to maximize the mean discrepancy. An objective function with optimizing kernel $k$ then becomes $\min_\theta \max_\phi M_{\tilde{k}}(p_{\mathrm{data}}, p_{\mathrm{g}})$ and now is similar to the standard GAN objective as in Equation~\ref{eq:gan}. To enforce $f_\phi$ modeled with the neural network to be injective, an autoencoder satisfying $f_{decoder} \approx f^{-1}$ is adopted for $f_\phi$. 

Similar to other IPM metrics, MMD distance is continuous and differentiable almost everywhere in $\theta$. It can also be understood under the IPM framework with function class $\mathcal{F} = \mathcal{H}_{K}$ as discussed above. By introducing an RKHS with kernel $k$, MMD distance has an advantage over other feature matching metrics in that kernel $k$ can represent various feature space by mapping input data $x$ into other feature space. In particular, MMDGAN can also be connected with WGAN when $f_\phi$ is composited to a linear kernel with an output dimension of 1 instead of a Gaussian kernel. The moment matching technique using the Gaussian kernel also has an advantage over WGAN in that it can match even an infinite order of moments since exponential form can be represented as an infinite order via Taylor expansion while WGAN can be treated as a first-order moment matching problem as discussed above. However, a great disadvantage of measuring MMD distance is that computational cost grows quadratically as the number of samples grows \cite{WGAN}.

Meanwhile, CramerGAN \cite{CramerGAN} argues that the Wasserstein distance incurs biased gradients, suggesting the energy distance between two distributions. In fact, it measures energy distance indirectly in the data manifold but with transformation function $h$. However, CramerGAN can be thought of as the distance in the kernel embedded space of MMDGAN, which forces $h$ to be injective by the additional autoencoder reconstruction loss as discussed above.

\paragraph{Fisher GAN}

In addition to standard IPM in Equation~\ref{eq:IPM}, Fisher GAN \cite{FisherGAN} incorporates a data-dependent constraint by the following equations: 
\begin{align} 
d_\mathcal{F}(p_{\mathrm{data}}, p_{\mathrm{g}}) &= \sup_{f\in\mathcal{F}}\frac{\E_{x\sim p_{\mathrm{data}}}f(x)-\E_{x\sim p_{\mathrm{g}}}f(x)}{\sqrt[]{\frac{1}{2}(\E_{x\sim p_{\mathrm{data}}}f^2(x)+\E_{x\sim p_{\mathrm{g}}}f^2(x))}} \label{fisherfirst} \\
&= \sup_{f\in\mathcal{F}, \frac{1}{2}[\E_{x\sim p_{\mathrm{data}}}f^2(x)+\E_{x\sim p_{\mathrm{g}}}f^2(x)]=1}\E_{x\sim p_{\mathrm{data}}}f(x)-\E_{x\sim p_{\mathrm{g}}}f(x) \label{fisherlast}
\end{align}

Equation~\ref{fisherfirst} is motivated by fisher linear discriminant analysis (FLDA) \cite{FLDA} which not only maximizes the mean difference but also reduces the total with-in class variance of two distributions. Equation~\ref{fisherlast} follows from the constraining numerator of Equation~\ref{fisherfirst} to be 1. It is also, as are other IPM metrics, interpreted as a mean feature matching problem under the somewhat different constraints. Under the definition of Equation~\ref{eq:featurematching}, Fisher GAN can be converted into another mean feature matching problem with second order moment constraint. A mean feature matching problem derived from the FLDA concept is as follows:
\begin{align} 
d_{\mathcal{F}}(p_{\mathrm{data}}, p_{\mathrm{g}})&=\max_{w\in\Omega}\max_v\frac{<v, \E_{x\sim p_{\mathrm{data}}}\Phi(x)-\E_{x\sim p_{\mathrm{g}}}\Phi(x)>}{\sqrt[]{\frac{1}{2}(\E_{x\sim p_{\mathrm{data}}}f^2(x)+\E_{x\sim p_{\mathrm{g}}}f^2(x))}} \label{fisherexfirst} \\
&= \max_{w\in\Omega}\max_v\frac{<v, \E_{x\sim p_{\mathrm{data}}}\Phi(x)-\E_{x\sim p_{\mathrm{g}}}\Phi(x)>}{\sqrt[]{\frac{1}{2}(\E_{x\sim p_{\mathrm{data}}}v^T\Phi(x)\Phi(x)^Tv+\E_{x\sim p_{\mathrm{g}}}v^T\Phi(x)\Phi(x)^Tv)}} \label{fisherexsecond} \\
&= \max_{w\in\Omega}\max_v\frac{<v, \mu_w(p_{\mathrm{data}})-\mu_w(p_{\mathrm{g}})>}{\sqrt[]{v^T(\frac{1}{2}\sum_w(p_{\mathrm{data}})+\frac{1}{2}\sum_w(p_{\mathrm{g}})+\gamma I_m)v}} \label{fisherexthird}\\
&= \max_{w\in\Omega}\max_{v,v^T(\frac{1}{2}\sum_w(p_{\mathrm{data}})+\frac{1}{2}\sum_w(p_{\mathrm{g}})+\gamma I_m)v=1}<v, \mu_w(p_{\mathrm{data}})-\mu_w(p_{\mathrm{g}})> \label{fisherexlast}
\end{align}
where $\mu_w(\mathcal{P})=\E_{x\sim\mathcal{P}}\Phi_w(x)$ denotes an embedding mean and $\sum_w(\mathcal{P})=\E_{x\sim\mathcal{P}}\Phi_w(x)\Phi_w^T(x)$ denotes an embedding covariance for the probability $\mathcal{P}$.

Equation~\ref{fisherexsecond} can be induced by using the inner product of $f$ defined as in Equation~\ref{eq:featurematching}. $\gamma I_m$ of Equation~\ref{fisherexthird} is an $m$ by $m$ identity matrix that guarantees a numerator of the above equations not to be zero. In Equation~\ref{fisherexlast}, Fisher GAN aims to find the embedding direction $v$ which maximizes the mean discrepancy while constraining it to lie in a hyperellipsoid as $v^T(\frac{1}{2}\sum_w(p_{\mathrm{data}})+\frac{1}{2}\sum_w(p_{\mathrm{g}})+\gamma I_m)v=1$ represents. It naturally derives the Mahalanobis distance \cite{maha} which is defined as a distance between two distributions given a positive definite matrix such as a covariance matrix of each class. More importantly, Fisher GAN has advantages over WGAN. It does not impose a data independent constraint such as weight clipping which makes training too sensitive on the clipping value, and it has computational benefit over the gradient penalty method in WGAN-GP \cite{WGAN-GP} as the latter must compute gradients of the critic while Fisher GAN computes covariances.

\paragraph{Comparison to f-divergence}

The f-divergence family, which can be defined as in Equation~\ref{eq:fganoriginal} with a convex function $f$, has restrictions in that as the dimension $d$ of the data space $x\in\mathcal{X}=R^d$ increases, the f-divergence is highly difficult to estimate, and the supports of two distributions tends to be unaligned, which leads a divergence value to infinity \cite{IPMf-divergence}. Even though Equation~\ref{fganlast} derives a variational lower bound of Equation~\ref{eq:fganoriginal} which looks very similar to Equation~\ref{eq:IPM}, the tightness of the lower bound to the true divergence is not guaranteed in practice and can incur an incorrect, biased estimation. 

\citet{IPMf-divergence} showed that the only non-trivial intersection between the f-divergence family and the IPM family is total variation distance; therefore, the IPM family does not inherit the disadvantages of f-divergence. They also proved that IPM estimators using finite i.i.d. samples are more consistent in convergence whereas the convergence of f-divergence is highly dependent on data distributions.  

Consequently, employing an IPM family to measure distance between two distributions is advantageous over using an f-divergence family because IPM families are not affected by data dimension and consistently converge to the true distance between two distributions. Moreover, they do not diverge even though the supports of two distributions are disjointed. In addition, Fisher GAN \cite{FisherGAN} is also equivalent to the Chi-squared distance \cite{chi}, which can be covered by an f-divergence framework. However, with a data dependent constraint, Chi-squared distance can use IPM family characteristics, so it is more robust to unstable training of the f-divergence estimation.

\subsubsection{Auxiliary object functions}
In Section~\ref{sec:f} and Section~\ref{sec:ipm}, we demonstrated various objective functions for adversarial learning. Concretely, through a minimax iterative algorithm, the discriminator estimates a specific kind of distance between $p_{\mathrm{g}}(x)$ and $p_{\mathrm{data}}(x)$. The generator reduces the estimated distance to make two distributions closer. Based on this adversarial objective function, a GAN can incorporate with other types of objective functions to help the generator and the discriminator stabilize during training or to perform some kinds of tasks such as classification. In this section, we introduce auxiliary object functions attached to the adversarial object function, mainly a reconstruction objective function and a classification objective function.

\paragraph{Reconstruction object function} \label{rof}
Reconstruction is to make an output image of a neural network to be the same as an original input image of a neural network. The purpose of the reconstruction is to encourage the generator to preserve the contents of the original input image \cite{DiscoGAN,CycleGAN,alpha-GAN,AGE,MRGAN} or to adopt auto-encoder architecture for the discriminator \cite{EBGAN,BEGAN,MAGAN}. For a reconstruction objective function, mostly the L1 norm of the difference of the original input image and the output image is used.

When the reconstruction objective term is used for the generator, the generator is trained to maintain the contents of the original input image. In particular, this operation is crucial for tasks where semantic and several modes of the image should be maintained, such as image translation \cite{DiscoGAN,CycleGAN} (detailed in Section~\ref{unpaired}) and auto-encoder reconstruction \cite{alpha-GAN,MRGAN,AGE} (detailed in Section~\ref{waa}). The intuition of using reconstruction loss for the generator is that it guides the generator to restore the original input in a supervised learning manner. Without a reconstruction loss, the generator is to simply fool the discriminator, so there is no reason for the generator to maintain crucial parts of the input. As a supervised learning approach, the generator treats the original input image as label information, so we can reduce the space of possible mappings of the generator in a way we desire.

There are some GAN variants using a reconstruction objective function for the discriminator, which is naturally derived from auto-encoder architecture for the discriminator \cite{BEGAN,MAGAN,EBGAN}. These GANs are based on the aspect that views the discriminator as an energy function, and will be detailed in Section~\ref{aea}.

\paragraph{Classification object functions}
A cross entropy loss for a classification is widely added for many GAN applications where labeled data exists, especially semi-supervised learning and domain adaptation. Cross entropy loss can be directly applied to the discriminator, which gives the discriminator an additional role of classification \cite{AC-GAN,CatGAN}. Other approaches \cite{triple,DANN,Unsupervised,AUNDA} adopt classifier explicitly, training the classifier jointly with the generator and the discriminator through a cross entropy loss (detailed in Section~\ref{ssl} and \ref{domainadaptation}).
\subsection{Architecture}

An architecture of the generator and the discriminator is important as it highly influences the training stability and performance of GAN. Various papers adopt several techniques such as batch normalization, stacked architecture, and multiple generators and discriminators to promote adversarial learning. We start with deep convolutional GAN (DCGAN) \cite{DCGAN}, which provides a remarkable benchmark architecture for other GAN variants.

\subsubsection{DCGAN}

DCGAN provides significant contributions to GAN in that its suggested convolution neural network (CNN) \cite{cnn} architecture greatly stabilizes GAN training. 
DCGAN suggests an architecture guideline in which the generator is modeled with a transposed CNN \cite{transposedc}, and the discriminator is modeled with a CNN with an output dimension 1. It also proposes other techniques such as batch normalization and types of activation functions for the generator and the discriminator to help stabilize the GAN training. As it solves the instability of training GAN only through architecture, it becomes a baseline for modeling various GANs proposed later. 
For example, \citet{GRAN} uses a recurrent neural network (RNN) \cite{rnn} to generate images motivated by DCGAN. By accumulating images of each time step output of DCGAN and combining several time step images, it produces higher visual quality images.  





\subsubsection{Hierarchical architecture}

\begin{figure} [t]
\centering
\includegraphics[width=0.9\textwidth]{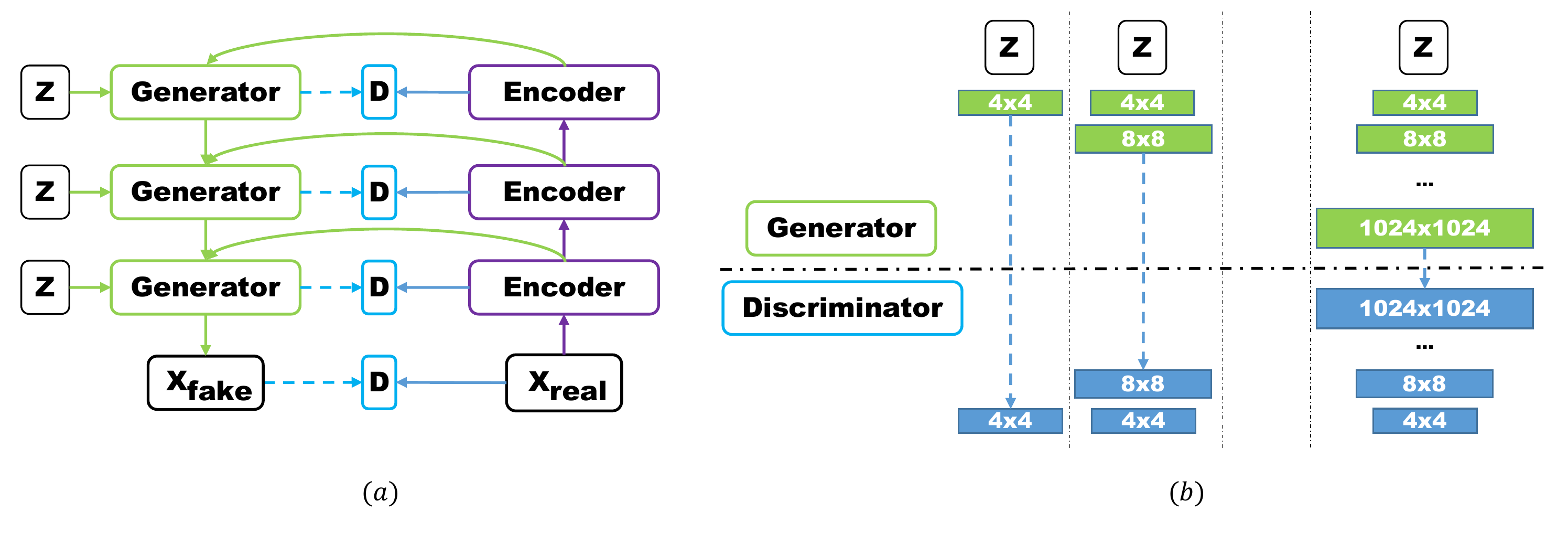}
\caption{\label{fig:archi} Illustrations of (a) StackedGAN \cite{StackedGAN} and (b) Progressive GAN \cite{karras2017progressive}.}
\end{figure}

In this section, we describe GAN variants that stack multiple generator-discriminator pairs. Commonly, these GANs generate samples in multiple stages to generate large-scale and high-quality samples. The generator of each stage is utilized or conditioned to help the next stage generator to better produce samples as shown in Figure~\ref{fig:archi}. 

\paragraph{Hierarchy using multiple GAN pairs}

StackedGAN \cite{StackedGAN} attempts to learn a hierarchical representation by stacking several generator-discriminator pairs. For each layer of a generator stack, there exists the generator which produces level-specific representation, the corresponding discriminator training the  generator adversarially at each level and an encoder which generates the semantic features of real samples. Figure \textcolor{red}{3a} shows a flowchart of StackedGAN. Each generator tries to produce a plausible feature representation that can deceive the corresponding discriminator, given previously generated features and the corresponding hierarchically encoded features.

In addition, Gang of GAN (GoGAN) \cite{GoGAN} proposes to improve WGAN \cite{WGAN} by adopting multiple WGAN pairs. For each stage, it changes the Wasserstein distance to a margin-based Wasserstein disitance as $[D(G(z))-D(x)+m]^{+}$ so the discriminator focuses on generated samples whose gap $D(x)-D(G(z))$ is less than $m$. In addition, GoGAN adopts ranking loss for adjacent stages which induces the generator in later stages to produce better results than the former generator by using a smaller margin at the next stage of the generation process. By progressively moving stages, GoGAN aims to gradually reduce the gap between $p_{\mathrm{data}}(x)$ and $p_{\mathrm{g}}(x)$.


\paragraph{Hierarchy using a single GAN}

Generating high resolution images is highly challenging since a large scale generated image is easily distinguished by the discriminator, so the generator often fails to be trained. Moreover, there is a memory issue in that we are forced to set a low mini-batch size due to the large size of neural networks. Therefore, some studies adopt hierarchical stacks of multiple generators and discriminators \cite{MAD-GAN,huang2017stacked,durugkar2016generative}. This strategy divides a large complex generator's mapping space step by step for each GAN pair, making it easier to learn to generate high resolution images. However, Progressive GAN \cite{karras2017progressive} succeeds in to generating high resolution images in a single GAN, making training faster and more stable. 

Progressive GAN generates high resolution images by stacking each layer of the generator and the discriminator incrementally as shown in Figure \textcolor{red}{3b}. It starts training to generate a very low spatial resolution (\eg 4$\times$4), and progressively doubles the resolution of generated images by adding layers to the generator and the discriminator incrementally. In addition, it proposes various training techniques such as pixel normalization, equalized learning rate and mini-batch standard deviation, all of which help GAN training to become more stable.


\subsubsection{Auto encoder architecture} \label{aea}
An auto encoder is a neural network for unsupervised learning, It assigns its input as a target value, so it is trained in a self-supervised manner. The reason for self-reconstruction is to encode a compressed representation or features of the input, which is widely utilized with a decoder. Its usage will be detailed in Section~\ref{waa}.

In this section, we describe GAN variants which adopt an auto encoder as the discriminator. These GANs view the discriminator as an energy function, not a probabilistic model that distinguishes its input as real or fake. An energy model assigns a low energy for a sample lying near the data manifold (a high data density region), while assigning a high energy for a contrastive sample lying far away from the data manifold (a low data density region). These variants are mainly Energy Based GAN (EBGAN), Boundary Equilibrium GAN (BEGAN) and Margin Adaptation GAN (MAGAN), all of which frame GAN as an energy model.

Since an auto encoder is utilized for the discriminator, a pixelwise reconstruction loss between an input and an output of the discriminator is naturally adopted for the discriminator's energy function and is defined as follows:
\begin{align} \label{rec}
D(v)&=\Vert v-AE(v)\Vert
\end{align}
where $AE:R^{N_x}\Rightarrow R^{N_x}$ denotes an auto encoder and $R^{N_x}$ represents the dimension of an input and an output of an auto encoder. It is noted that $D(v)$ in Equation~\ref{rec} is the pixelwise L1 loss for an autoencoder which maps an input $v\in R^{N_x}$ into a positive real number $R^{+}$.

A discriminator with an energy $D(v)$ is trained to give a low energy for a real $v$ and a high energy for a generated $v$. From this point of view, the generator produces a contrastive sample for the discriminator, so that the discriminator is forced to be regularized near the data manifold. Simultaneously, the generator is trained to generate samples near the data manifold since the discriminator is encouraged to reconstruct only real samples. 
Table~\ref{tab:Autoencoder} presents the summarized details of BEGAN, EBGAN and MAGAN. 
$L_G$ and $L_D$ indicate the generator loss and the discriminator loss, respectively, and 
$[t]^+=\max(0,t)$ represents a maximum value between $t$ and $0$, which act as a hinge.

\begin{table}[t]
\begin{center}
\caption{\label{tab:Autoencoder}An autoencoder based GAN variants (BEGAN, EBGAN and MAGAN).}
\begin{tabular}{c|p{6cm}|p{4cm}}
 & Objective function & Details \tabularnewline
\Xhline{3\arrayrulewidth}
BEGAN \cite{BEGAN} & $L_D=D(x)-k_tD(G(z))$ \par $L_G=D(G(z))$ \par $k_{t+1}=k_t+\alpha(\gamma D(x)-D(G(z)))$ & Wasserstein distance \par between loss distributions \\\hline
EBGAN \cite{EBGAN} & $L_D=D(x)+[m-D(G(z))]^{+}$ \par $L_G=D(G(z))$ & Total Variance($p_{\mathrm{data}}, p_{\mathrm{\theta}}$) \\\hline
MAGAN \cite{MAGAN} & $L_D=D(x)+[m-D(G(z))]^{+}$ \par $L_G=D(G(z))$ & Margin $m$ is adjusted in \par EBGAN's training 
\end{tabular}
\end{center}
\vspace{-2mm}
\end{table}

\begin{figure} [t]
    \centering    
    \includegraphics[width=0.7\textwidth]{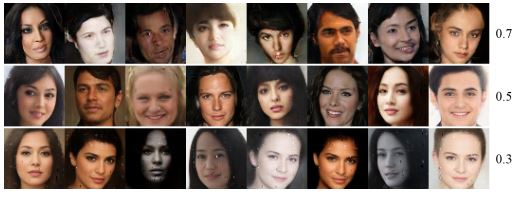}
    \caption{Random images sampled from the generator at varying $\gamma \in \{0.3, 0.5, 0.7\}$ of BEGAN \cite{BEGAN}. Samples at lower $\gamma$ shows similar images. At high $\gamma$ values, image diversity seems to increase, but have some artifacts. Images from BEGAN \cite{BEGAN}.}\label{fig:began}
\end{figure}

Boundary equilibrium GAN (BEGAN) \cite{BEGAN} uses the fact that pixelwise loss distribution follows a normal distribution by CLT. It focuses on matching loss distributions through Wasserstein distance and not on directly matching data distributions. In BEGAN, the discriminator has two roles: one is to reconstruct real samples sufficiently and the other is to balance the generator and the discriminator via an equilibrium hyperparameter $\gamma=\E[L(G(z))]/\E[L(x)]$. $\gamma$ is fed into an objective function to prevent the discriminator from easily winning over the generator; therefore, this balances the power of the two components. Figure~\ref{fig:began} shows face images at varying $\gamma$ of BEGAN.

Energy-based GAN (EBGAN) \cite{EBGAN} interprets the discriminator as an energy agent, which assigns low energy to real samples and high energy to generated samples. Through the $[m-L(G(z))]^{+}$ term in an objective function, the discriminator ignores generated samples with higher energy than $m$ so the generator attempts to synthesize samples that have lower energy than $m$ to fool the discriminator, which allows that mechanism to stabilize training. Margin adaptation GAN (MAGAN) \cite{MAGAN} takes a similar approach to EBGAN, where the only difference is that MAGAN does not fix the margin $m$. MAGAN shows empirically that the energy of the generated sample fluctuates near the margin $m$ and that phenomena with a fixed margin make it difficult to adapt to the changing dynamics of the discriminator and generator. MAGAN suggests that margin $m$ should be adapted to the expected energy of real data, thus, $m$ is monotonically reduced, so the discriminator reconstructs real samples more efficiently.  

In addition, because the total variance distance belongs to an IPM family with the function class $\mathcal{F}=\{f:\Vert f\Vert_\infty=\sup_x|f(x)|\leq 1\}$, it can be shown that EBGAN is equivalent to optimizing the total variance distance by using the fact that the discriminator's output for generated samples is only available for $0\leq D\leq m$ \cite{WGAN}. Because the total variance is the only intersection between IPM and f-divergence \cite{IPMf-divergence}, it inherits some disadvantages for estimating f-divergence as discussed by \citet{WGAN} and \citet{IPMf-divergence}.
\subsection{Obstacles in Training GAN} \label{obstacle}

In this section, we discuss theoretical and practical issues related to the training dynamics of GAN. We first evaluate a theoretical problem of the standard GAN, which is incurred from the fact that the discriminator of GAN aims to approximate the JSD \cite{GAN} between $p_{\mathrm{data}}(x)$ and $p_{\mathrm{\theta}}(x)$ and the generator of GAN tries to minimize the approximated JSD, as discussed in Section~\ref{theoissue}. We then discuss practical issues, especially a mode collapse problem where the generator fails to capture the diversity of real samples, and generates only specific types of real samples, as discussed in Section~\ref{pracissue}. 

\subsubsection{Theoretical issues} \label{theoissue}

As mentioned in Section~\ref{intro}, the traditional generative model is to maximize a likelihood of $p_{\mathrm{g}}(x)$. It can be shown that maximizing the log likelihood is equivalent to minimizing the Kullback-Leibler Divergence (KLD) between $p_{\mathrm{data}}(x)$ and $p_{\mathrm{\theta}}(x)$ as the number of samples $m$ increases: 
\begin{align} 
\theta^{\star} &= \argmax_\theta\lim_{m\to\infty}\frac{1}{m}\sum_{i=1}^{m}\log p_{\mathrm{g}}(x^{i})\label{mlklfirst} \\ 
&= \argmax_\theta \int_{x} p_{\mathrm{data}}(x)\log p_{\mathrm{g}}(x)dx \label{mlklsecond}\\
&= \argmin_\theta \int_{x} -p_{\mathrm{data}}(x)\log p_{\mathrm{g}}(x) dx +  p_{\mathrm{data}}(x)\log p_{\mathrm{data}}(x) dx \label{mlklthird}\\
&= \argmin_\theta \int_{x} p_{\mathrm{data}}(x) \log\frac{p_{\mathrm{data}}(x)}{p_{\mathrm{g}}(x)} dx \label{mlklfourth}\\
&= \argmin_\theta KLD(p_{\mathrm{data}}||p_{\mathrm{g}}) \label{mlkllast}
\end{align}
We note that we need to find an optimal parameter $\theta^{\star}$ for maximizing likelihood; therefore, $\argmax$ is used instead of $\max$. In addition, we replace our model's probability distribution $p_{\mathrm{\theta}}(x)$ with $p_{\mathrm{g}}(x)$ for consistency of notation.

Equation~\ref{mlklsecond} is established by the central limit theorem (CLT) \cite{clt} in that as $m$ increases, the variance of the expectation of the distribution decreases. Equation~\ref{mlklthird} can be induced because $p_{\mathrm{data}}(x)$ is not dependent on $\theta$, and Equation~\ref{mlkllast} follows from the definition of KLD. Intuitively, minimizing KLD between these two distributions can be interpreted as approximating $p_{\mathrm{data}}(x)$ with a large number of real training data because the minimum KLD is achieved when $p_{\mathrm{data}}(x) = p_{\mathrm{g}}(x)$. 

\begin{figure} [t]
    \centering
    \begin{subfigure}[b]{0.25\textwidth}
        \includegraphics[width=\textwidth]{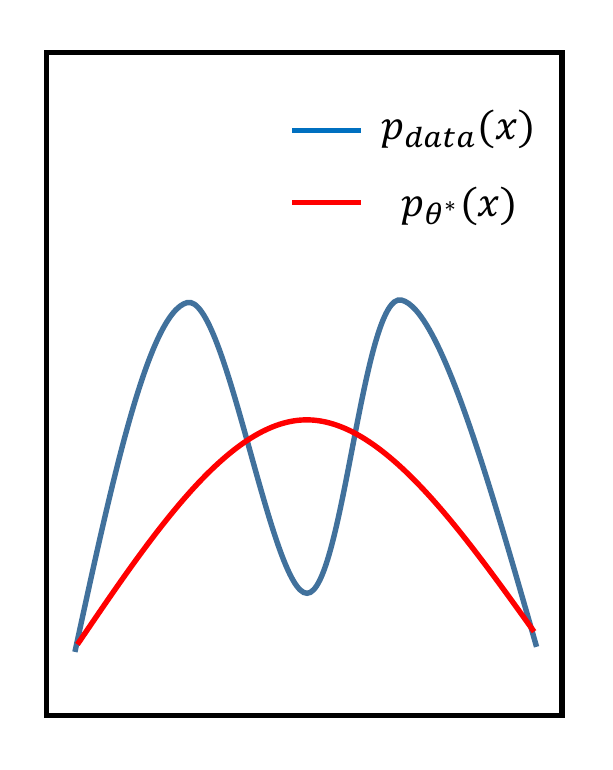}
        \caption{$KLD(p_{\mathrm{data}}||p_{\mathrm{g}})$}
        \label{fig:kla}
    \end{subfigure}
    \begin{subfigure}[b]{0.25\textwidth}
        \includegraphics[width=\textwidth]{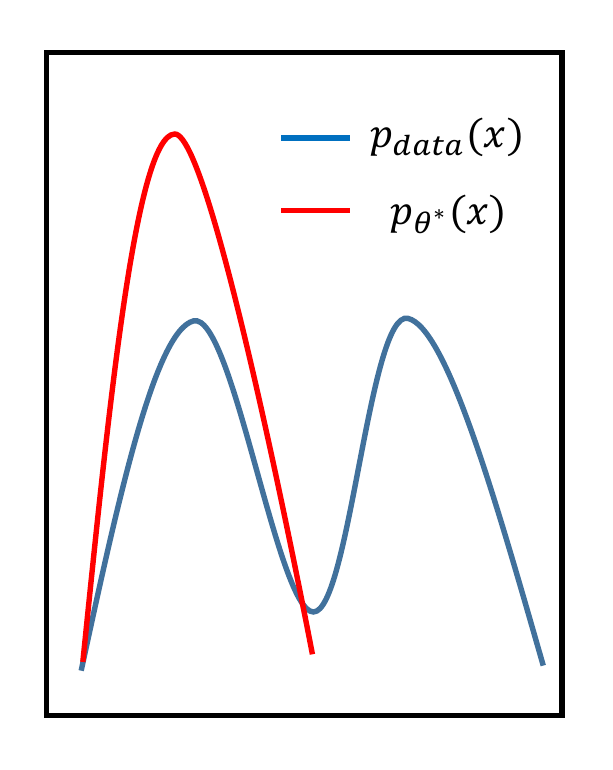}
        \caption{$KLD(p_{\mathrm{g}}||p_{\mathrm{data}})$}
        \label{fig:klb}
    \end{subfigure}
    \caption{Different behavior of asymmetric KLD. Images reproduced from \cite{GOODFELLOWNIPS}}\label{fig:kl}
\end{figure}

Thus, the result of maximizing likelihood is equivalent to minimizing KLD$(p_{\mathrm{data}}||p_{\mathrm{g}})$ given infinite training samples. Because KLD is not symmetrical, minimizing KLD$(p_{\mathrm{g}}||p_{\mathrm{data}})$ gives a different result. Figure~\ref{fig:kl} from \citet{GOODFELLOWNIPS} shows the details of different behaviors of asymmetric KLD where Figure~\ref{fig:kla} shows minimizing KLD$(p_{\mathrm{data}}||p_{\mathrm{g}})$ and Figure~\ref{fig:klb} shows minimizing KLD$(p_{\mathrm{g}}||p_{\mathrm{data}})$ given a mixture of two Gaussian distributions $p_{\mathrm{data}}(x)$ and the single Gaussian distribution $p_{\mathrm{g}}(x)$. $\theta^{\star}$ in each figure denotes the argument minimum of each asymmetric KLD. For Figure~\ref{fig:kla}, the points where $p_{data}\neq 0$ contribute to the value of KLD and the other points at which $p_{data}$ is small rarely affect the KLD. Thus, $p_{\mathrm{g}}$ becomes nonzero on the points where $p_{data}$ is nonzero. Therefore, $p_{\theta^{\star}}(x)$ in Figure~\ref{fig:kla} is averaged for all modes of $p_{\mathrm{data}}(x)$ as $KLD(p_{\mathrm{data}}||p_{\mathrm{g}})$ is more focused on covering all parts of $p_{\mathrm{data}}(x)$. In contrast, for KLD$(p_{\mathrm{g}}||p_{\mathrm{data}})$, the points of which $p_{data}=0$ but $p_{\mathrm{g}}\neq0$ contribute to a high cost. This is why $p_{\theta^{\star}}(x)$ in Figure~\ref{fig:klb} seeks to find an $x$ which is highly likely from $p_{\mathrm{data}}(x)$.  

JSD has an advantage over the two asymmetric KLDs in that it accounts for both mode dropping and sharpness. It never explodes to infinity unlike KLD even though there exists a point $x$ that lies outside of $p_{\mathrm{g}}(x)$'s support which makes $p_{\mathrm{g}}(x)$ equal 0. \citet{GAN} showed that the discriminator $D$ aims to approximate $V(G, D^{\star})=2JSD(p_{\mathrm{data}}||p_{\mathrm{g}}) - 2\log2$ for the fixed generator $G$ between $p_{\mathrm{g}}(x)$ and $p_{\mathrm{data}}(x)$, where $D^{\star}$ is an optimal discriminator and $V(G,D)$ is defined in Equation~\ref{eq:gan}. Concretely, if $D$ is trained well so that it approximates $JSD(p_{\mathrm{data}}||p_{\mathrm{g}}) - 2\log2$ sufficiently, training $G$ minimizes the approximated distance. However, \citet{Towards} reveal mathematically why approximating $V(G, D^{\star})$ does not work well in practice.


%
%
\citet{Towards} proved why training GAN is fundamentally unstable. When the supports of two distributions are disjointed or lie in low-dimensional manifolds, there exists the perfect discriminator which classifies real or fake samples perfectly and thus, the gradient of the discriminator is 0 at the supports of the two distributions. It has been proven empirically and mathematically that $p_{\mathrm{data}}(x)$ and $p_{\mathrm{g}}(x)$ derived from $z$ have a low-dimensional manifold in practice \cite{sample}, and this fact allows $D$'s gradient transferred to $G$ to vanish as $D$ becomes to perfectly classify real and fake samples. Because, in practice, $G$ is optimized with a gradient based optimization method, $D$'s vanishing (or exploding) gradient hinders $G$ from learning enough through $D$'s gradient feedback. Moreover, even with the alternate $-\log D(G(z))$ objective proposed in \cite{GAN}, minimizing an objective function is equivalent to simultaneously trying to minimize KLD$(p_{\mathrm{g}} || p_{\mathrm{data}})$ and maximize JSD$(p_{\mathrm{g}}||p_{\mathrm{data}})$. As these two objectives are opposites, this leads the magnitude and variance of $D$'s gradients to increase as training progresses, causing unstable training and making it difficult to converge to equilibrium. To summarize, training the GAN is theoretically guaranteed to converge if we use an optimal discriminator $D^{\star}$ which approximates JSD, but this theoretical result is not implemented in practice when using gradient based optimization. In addition to the $D$'s improper gradient problem discussed in this paragraph, there are two practical issues as to why GAN training suffers from nonconvergence. 

\subsubsection{Practical issues} \label{pracissue}

First, we represent $G$ and $D$ as deep neural networks to learn parameters rather than directly learning $p_{\mathrm{g}}(x)$ itself. Modeling with deep neural networks such as the multilayer perceptron (MLP) or CNN is advantageous in that the parameters of distributions can be easily learned through gradient descent using back-propagation. This does not require further distribution assumptions to produce an inference; rather, it can generate samples following $p_{\mathrm{g}}$(x) through simple feed-forward. However, this practical implementation causes a gap with theory. \citet{GAN} provide theoretical convergence proof based on the convexity of probability density function in $V(G,D)$. However, as we model $G$ and $D$ with deep neural networks, the convexity does not hold because we now optimize in the parameter space rather than in the function space (where assumed theoretical analysis lies). Therefore, theoretical guarantees do not hold anymore in practice. For a further issue related to parameterized neural network space, \citet{MIXGAN} discussed the existence of the equilibrium of GAN, and showed that a large capacity of $D$ does not guarantee $G$ to generate all real samples perfectly, meaning that an equilibrium may not exist under a certain finite capacity of $D$. 

A second problem is related to an iterative update algorithm suggested in \citet{GAN}. We wish to train $D$ until optimal for fixed $G$, but optimizing $D$ in such a manner is computationally expensive. Naturally, we must train $D$ in certain $k$ steps and that scheme causes confusion as to whether it is solving a minimax problem or a maximin problem, because $D$ and $G$ are updated alternatively by gradient descent in the iterative procedure. Unfortunately, solutions of the minimax and maximin problem are not generally equal as follows:
\begin{equation} \label{maximin}
\min_G \max_D V(G,D) \neq \max_D \min_G V(G,D)
\end{equation}

With a maximin problem, minimizing $G$ lies in the inner loop in the right side of Equation~\ref{maximin}. $G$ is now forced to place its probability mass on the most likely point where the fixed nonoptimal $D$ believes it likely to be real rather than fake. After $D$ is updated to reject the generated fake one, $G$ attempts to move the probability mass to the other most likely point for fixed $D$. In practice, real data distribution is normally a multi modal distribution but in such a maximin training procedure, $G$ does not cover all modes of the real data distribution because $G$ considers that picking only one mode is enough to fool $D$. Empirically, $G$ tends to cover only a single mode or a few modes of real data distribution. This undesirable nonconvergent situation is called a mode collapse. A mode collapse occurs when many modes in the real data distribution are not represented in the generated samples, resulting in a lack of diversity in the generated samples. It can be simply considered as $G$ being trained to be a non one-to-one function which produces a single output value for several input values. 

Furthermore, the problem of the existence of the perfect discriminator we discussed in the above paragraph can be connected to a mode collapse. First, assume $D$ comes to output almost 1 for all real samples and 0 for all fake samples. Then, because $D$ produces values near 1 for all possible modes, there is no need for $G$ to represent all modes of real data probability. The theoretical and practical issues discussed in this section can be summarized as follows.
\begin{itemize}
\item Because the supports of distributions lie on low dimensional manifolds, there exists the perfect discriminator whose gradients vanish on every data point. Optimizing the generator may be difficult because it is not provided with any information from the discriminator.
\item GAN training optimizes the discriminator for the fixed generator and the generator for fixed discriminator simultaneously in one loop, but it sometimes behaves as if solving a maximin problem, not a minimax problem. It critically causes a mode collapse. In addition, the generator and the discriminator optimize the same objective function $V(G,D)$ in opposite directions which is not usual in classical machine learning, and often suffers from oscillations causing excessive training time.
\item The theoretical convergence proof does not apply in practice because the generator and the discriminator are modeled with deep neural networks, so optimization has to occur in the parameter space rather than in learning the probability density function itself.
\end{itemize} 

\subsubsection{Training techniques to improve GAN training} \label{ttitg}
As demonstrated in Section~\ref{theoissue} and \ref{pracissue}, GAN training is highly unstable and difficult because GAN is required to find a Nash equilibrium of a non-convex minimax game with high dimensional parameters but GAN is typically trained with gradient descent \cite{Improved}. In this section, we introduce some techniques to improve training of GAN, to make training more stable and produce better results.
\begin{itemize}
\item Feature matching \cite{Improved}: \\
This technique substitutes the discriminator's output in the objective function (Equation~\ref{eq:gan}) with an activation function's output of an intermediate layer of the discriminator to prevent overfitting from the current discriminator. Feature matching does not aim on the discriminator's output, rather it guides the generator to see the statistics or features of real training data, in an effort to stabilize training.

\item Label smoothing \cite{Improved}: \\
As mentioned previously, $V(G,D)$ is a binary cross entropy loss whose real data label is 1 and its generated data label is 0. However, since a deep neural network classifier tends to output a class probability with extremely high confidence \cite{GOODFELLOWNIPS}, label smoothing encourages a deep neural network classifier to produce a more soft estimation by assigning label values lower than 1. Importantly, for GAN, label smoothing has to be made for labels of real data, not for labels of fake data, since, if not, the discriminator can act incorrectly \cite{GOODFELLOWNIPS}. 

\item Spectral normalization \cite{miyato2018spectral}: \\
As we see in Section~\ref{wagan} and \ref{wgangp}, WGAN and Improved WGAN impose the discriminator to have Lipschitz continuity which constrain the magnitude of function differentiation. Spectral normalization aims to impose a Lipschitz condition for the discriminator in a different manner. Instead of adding a regularizing term or weight clipping, spectral normalization constrains the spectral norm of each layer of the discriminator where the spectral norm is the largest singular value of a given matrix. Since a neural network is a composition of multi layers, spectral normalization normalizes the weight matrices of each layer to make the whole network Lipschitz continuous. In addition, compared to the gradient penalty method proposed in Improved WGAN, spectral normalization is computationally beneficial since gradient penalty regularization directly controls the gradient of the discriminator.

\item PatchGAN \cite{pix2pix}: \\
PatchGAN is not a technique for stabilizing training of GAN. However, PatchGAN greatly helps to generate sharper results in various applications such as image translation \cite{pix2pix,CycleGAN}. Rather than producing a single output from the discriminator, which is a probability for its input's authenticity, PatchGAN \cite{pix2pix} makes the discriminator produce a grid output. For one element of the discriminator's output, its receptive field in the input image should be one small local patch in the input image, so the discriminator aims to distinguish each patch in the input image. To achieve this, one can remove the fully connected layer in the last part of the discriminator in the standard GAN. As a matter of fact, PatchGAN is equivalent to adopting multiple discriminators for every patch of the image, making the discriminator help the generator to represent more sharp images locally.

\end{itemize}

\subsection{Methods to Address Mode Collapse in GAN} \label{modecollapse}

Mode collapse which indicates the failure of GAN to represent various types of real samples is the main catastrophic problem of a GAN. From a perspective of the generative model, mode collapse is a critical obstacle for a GAN to be utilized in many applications, since the diversity of generated data needs to be guaranteed to represent the data manifold concretely. Unless multi modes of real data distribution are not represented by the generative model, such a model would be meaningless to use. 

\begin{figure} [t]
\centering
\includegraphics[width=0.8\textwidth]{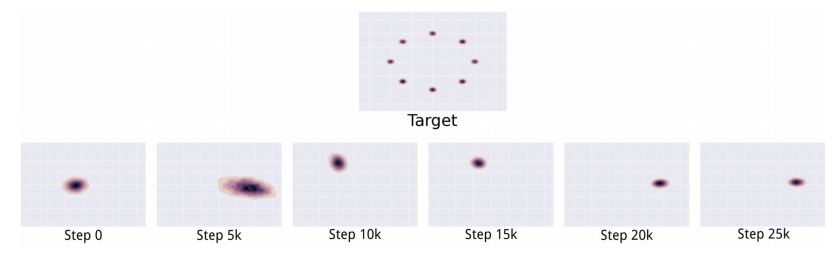}
\caption{\label{fig:modecollapse} An illustration of the mode collapse problem. Images from Unrolled GAN \cite{UnrolledGAN}.}
\end{figure}

Figure~\ref{fig:modecollapse} shows a mode collapse problem for a toy example. A target distribution $p_{\mathrm{data}}$ has multi modes, which is a Gaussian mixture in two-dimensional space \cite{UnrolledGAN}. Figures in the lower row represent learned distribution $p_{\mathrm{g}}$ as the training progresses. As we see in Figure~\ref{fig:modecollapse}, the generator does not cover all possible modes of the target distribution. Rather, the generator covers only a single mode, switching between different modes as the training goes on. The generator learns to produce a single mode, believing that it can fool the discriminator. The discriminator counter acts the generator by rejecting the chosen mode. Then, the generator switches to another mode which is believed to be real. This training behavior keeps proceeding, and thus, the convergence to a distribution covering all the modes is highly difficult.

In this section, we present several studies that suggest methods to overcome the mode collapse problem. In Section~\ref{ofm}, we demonstrate studies that exploit new objective functions to tackle a mode collapse, and in Section~\ref{am}, we introduce studies which propose architecture modifications. Lastly, in Section~\ref{mbd}, we describe mini-batch discrimination which is a notably and practically effective technique for the mode collapse problem.  

\subsubsection{Object function methods} \label{ofm}
Unrolled GAN \cite{UnrolledGAN} manages mode collapse with a surrogate objective function for the generator, which helps the generator predict the discriminator's response by unrolling the discriminator update $k$ steps for the current generator update. As we see in the standard GAN \cite{GAN}, it updates the discriminator first for the fixed generator and then updates the generator for the updated discriminator. Unrolled GAN differs from standard GAN in that it updates the generator based on a $k$ steps updated discriminator given the current generator update, which aims to capture how the discriminator responds to the current generator update. 
We see that when the generator is updated, it unrolls the discriminator's update step to consider the discriminator's $k$ steps future response with respect to the generator's current update while updating the discriminator in the same manner as the standard GAN. Since the generator is given more information about the discriminator's response, the generator spreads its probability mass to make it more difficult for the discriminator to react to the generator's behavior. It can be seen as empowering the generator because only the generator's update is unrolled, but it seems to be fair in that the discriminator can not be trained to be optimal in practice due to an infeasible computational cost while the generator is theoretically assumed to obtain enough information from the optimal discriminator.     

Deep regret analytic GAN (DRAGAN) \cite{DRAGAN} suggests that a mode collapse occurs due to the existence of a spurious local Nash Equilibrium in the nonconvex problem. DRAGAN addresses this issue by proposing constraining gradients of the discriminator around the real data manifold. It adds a gradient penalizing term which biases the discriminator to have a gradient norm of 1 around the real data manifold. This method attempts to create linear functions by making gradients have a norm of 1. Linear functions near the real data manifolds form a convex function space, which imposes a global unique optimum. Note that this gradient penalty method is also applied to WGAN-GP \cite{WGAN-GP}. They differ in that DRAGAN imposes gradient penalty constraints only to local regions around the real data manifold while Improved WGAN imposes gradient penalty constraints to almost everywhere around the generated data manifold and real data manifold, which leads to higher constraints than those of DRAGAN.

In addition, EBGAN proposes a repelling regularizer loss term to the generator, which encourages feature vectors in a mini-batch to be orthogonalized. This term is utilized with cosine similarities at a representation level of an encoder and forces the generator not to produce samples fallen in a few modes. 

\subsubsection{Architecture methods} \label{am}

Multi agent diverse GAN (MAD-GAN) \cite{MAD-GAN} adopts multiple generators for one discriminator to capture the diversity of generated samples as shown in Figure \textcolor{red}{7a}. To induce each generator to move toward different modes, it adopts a cosine similarity value as an additional objective term to make each generator produce dissimilar samples. This technique is inspired from the fact that as images from two different generators become similar, a higher similarity value is produced, thus, by optimizing this objective term, it may make each generator move toward different modes respectively. In addition, because each generator produces different fake samples, the discriminator's objective function adopts a soft-max cross entropy loss term to distinguish real samples from fake samples generated by multiple generators. 

Mode regularized GAN (MRGAN) \cite{MRGAN} assumes that mode collapse occurs because the generator is not penalized for missing modes. To address mode collapse, MRGAN adds an encoder which maps the data space $\mathcal{X}$ into the latent space $\mathcal{Z}$. Motivated from the manifold disjoint mentioned in Section~\ref{theoissue}, MRGAN first tries to match the generated manifold and real data manifold using an encoder. For manifold matching, the discriminator $D_M$ distinguishes real samples $x$ and its reconstruction $G\circ E(x)$, and the generator is trained with $D_M(G\circ E(x))$ with a geometric regularizer $d(x, G\circ E(x))$ where $d$ can be any metric in the data space. A geometric regularizer is used to reduce the geometric distance in the data space, to help the generated manifold move to the real data manifold, and allow the generator and an encoder to learn how to reconstruct real samples.
For penalizing missing modes, MRGAN adopts another discriminator $D_D$ which distinguishes $G(z)$ as fake and $G\circ E(x)$ as real. Since MRGAN matches manifolds in advance with a geometric regularizer, this modes diffusion step can distribute a probability mass even to minor modes of the data space with the help of $G\circ E(x)$. An outline of MRGAN is illustrated in Figure \textcolor{red}{7b}, where $R$ denotes a geometric regularizing term (reconstruction).

\begin{figure} [t]
\centering
\includegraphics[width=0.9\textwidth]{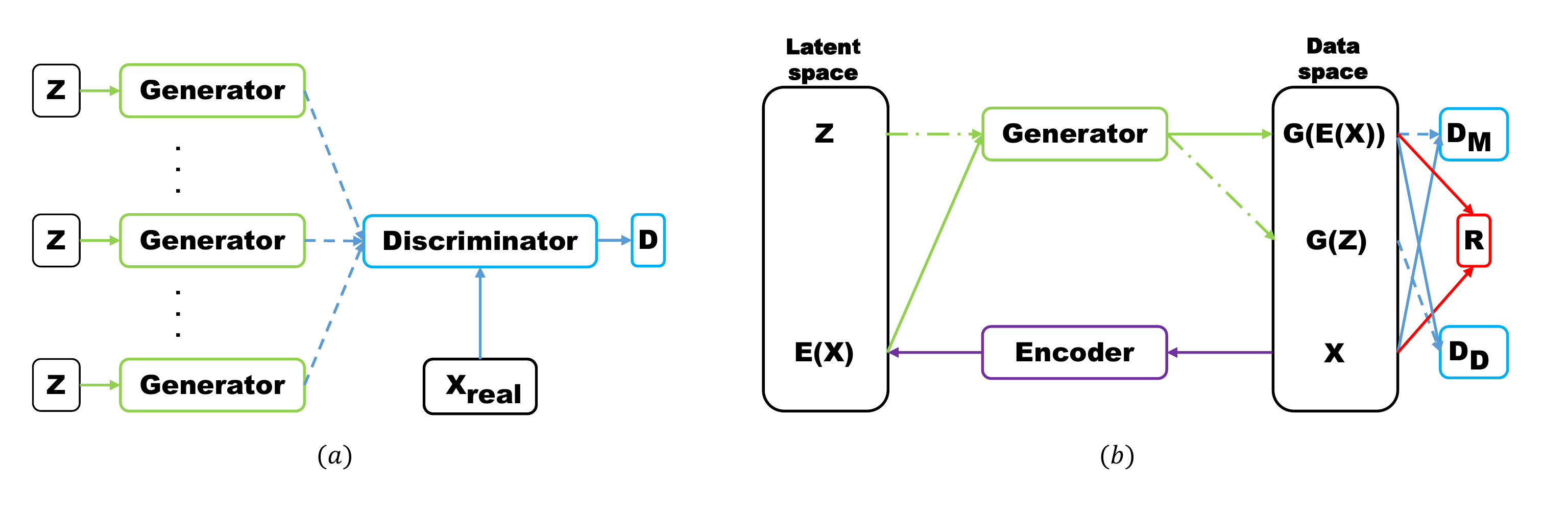}
\caption{\label{fig:mad} Illustrations of (a) MAD-GAN \cite{MAD-GAN} and (b) MRGAN \cite{MRGAN}.}
\end{figure}

\subsubsection{Mini-batch Discrimination} \label{mbd}

Mini-batch discrimination \cite{Improved} allows the discriminator to look at multiple examples in a mini-batch to avoid a mode collapse of the generator. The basic idea is that it encourages the discriminator to allow diversity directly in a mini-batch, not considering independent samples in isolation. To make the discriminator deal with not only each example but the correlation between other examples in a mini-batch simultaneously, it models a mini-batch layer in an intermediate layer of the discriminator, which calculates L1-distance based statistics of samples in a mini-batch. By adding such statistics to the discriminator, each example in a mini-batch can be estimated by how far or close to other examples in a mini-batch they are, and this information can be internally utilized by the discriminator, which helps the discriminator reflect samples' diversity to the output. For the aspect of the generator, it tries to create statistics similar to those of real samples in the discriminator by adversarial learning procedure. 

In addition, Progressive GAN \cite{karras2017progressive} proposed a simplified version of mini-batch discrimination, which uses the mean of the standard deviation for each feature (channel) in each spatial location over the mini-batch. This way do not add trainable parameters which projects statistics of the mini-batch while maintaining its effectiveness for a mode collapse. To summarize, mini-batch discrimination reflects samples' diversity to the discriminator, helping the discriminator determine whether its input batch is real or fake.

\section{Treating the Latent Space} \label{section3}
Latent space, also called an embedding space, is the space in which a compressed representation of data lies. If we wish to change or reflect some attributes of an image (for example, a pose, an age, an expression or even an object of an image), modifying images directly in the image space would be highly difficult because the manifolds where the image distributions lie are high-dimensional and complex. Rather, manipulating in the latent space is more tractable because the latent representation expresses specific features of the input image in a compressed manner.
In this section, we investigate how GAN handles latent space to represent target attributes and how a variational approach can be combined with the GAN framework.

\subsection{Latent Space Decomposition} \label{lsd}

The input latent vector $z$ of the generator is so highly entangled and unstructured that we do not know which vector point contains the specific representations we want. From this point of view, several papers suggest decomposing the input latent space to an input vector $c$, which contains the meaningful information and standard input latent vector $z$, which can be categorized into a supervised method and an unsupervised method.

\subsubsection{Supervised Methods}

Supervised methods require a pair of data and corresponding attributes such as the data's class label. The attributes are generally used as an additional input vector as explained below.

Conditional GAN (CGAN) \cite{CGAN} imposes a condition of additional information such as a class label to control the data generation process in a supervised manner by adding an information vector $c$ to the generator and discriminator. The generator takes not only a latent vector $z$ but also an additional information vector $c$ and the discriminator takes samples and the information vector $c$ so that it distinguishes fake samples given $c$. By doing so, CGAN can control the number of digits to be generated, which is impossible for standard GAN.

\begin{figure} [t]
    \centering
    \includegraphics[width=0.9\textwidth]{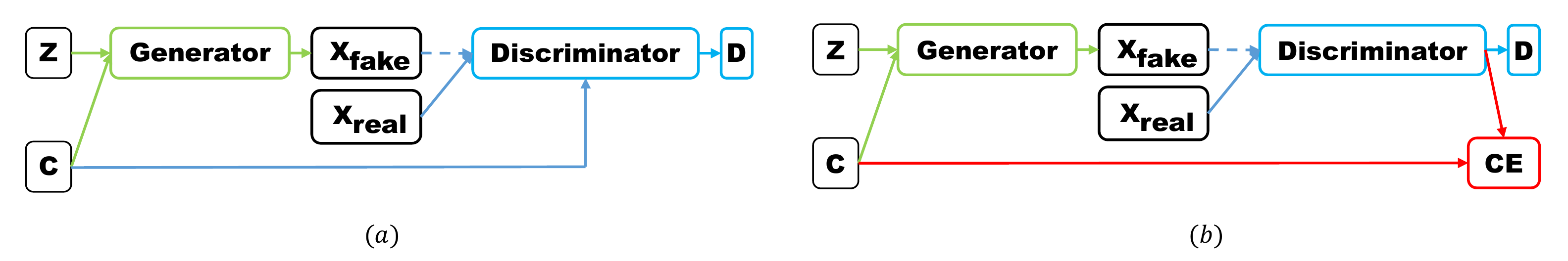}
    \caption{Illustrations of (a) CGAN \cite{CGAN}, (b) CGAN with a projection discriminator \cite{cganICLR} and (c) AC-GAN \cite{AC-GAN}.}\label{fig:Conditional}
\end{figure}

Auxiliary classifier GAN (AC-GAN) \cite{AC-GAN} takes a somewhat different approach than CGAN. It is trained by minimizing the log-likelihood of class labels with the adversarial loss. The discriminator produces not only the probability that the input samples are from the real dataset but also the probability over the class labels. Figure~\ref{fig:Conditional} outlines CGAN and CGAN with a projection discriminator and ACGAN where $CE$ denotes the cross entropy loss for the classification. 

In addition, plug and play generative networks (PPGN) \cite{PPGN} are another type of generative model that produce data under a given condition. Unlike the other methods described above, PPGN does not use the labeled attributes while training the generator. Instead, PPGN learns the auxiliary classifier and the generator producing real-like data via adversarial learning independently. Then, PPGN produces the data for the given condition using the classifier and the generator through an MCMC-based sampler. An important characteristic of PPGN is that they can work as plug and play. When a classifier pretrained with the same data but different labels is given to the generator, the generator can synthesize samples under the condition without further training.

\subsubsection{Unsupervised Methods}

Different from the supervised methods discussed above, unsupervised methods do not exploit any labeled information. Thus, they require an additional algorithm to disentangle the meaningful features from the latent space.

InfoGAN \cite{InfoGAN} decomposes an input noise vector into a standard incompressible latent vector $z$ and another latent variable $c$ to capture salient semantic features of real samples. Then, InfoGAN maximizes the amount of mutual information between $c$ and a generated sample $G(z,c)$ to allow $c$ to capture some noticeable features of real data. In other words, the generator takes the concatenated input $(z,c)$ and maximizes the mutual information, $I(c;G(z,c))$ between a given latent code $c$ and the generated samples $G(z,c)$ to learn meaningful feature representations. However, evaluating mutual information $I(c;G(z,c))$ needs to directly estimate the posterior probability $p(c|x)$, which is intractable. InfoGAN, thus, takes a variational approach which replaces a target value $I(c;G(z,c))$ by maximizing a lower bound. 

Both CGAN and InfoGAN learn conditional probability $p(x|c)$ given a certain condition vector $c$; however, they are dissimilar regarding how they handle condition vector $c$. In CGAN, additional information $c$ is assumed to be semantically known (such as class labels), so we have to provide $c$ to the generator and the discriminator during the training phase. On the other hand, $c$ is assumed to be unknown in InfoGAN, so we take $c$ by sampling from prior distribution $p(c)$ and control the generating process based on $I(c;G(z,c))$. As a result, the automatically inferred $c$ in InfoGAN has much more freedom to capture certain features of real data than $c$ in CGAN, which is restricted to known information.

Semi-supervised InfoGAN (ss-InfoGAN) \cite{ss-InfoGAN} takes advantage of both supervised and unsupervised methods. It introduces some label information in a semi-supervised manner by decomposing latent code $c$ in to two parts, $c=c_{ss}\dot{\bigcup} c_{us}$. Similar to InfoGAN, ss-InfoGAN attempts to learn the semantic representations from the unlabeled data by maximizing the mutual information between the generated data and the unsupervised latent code $c_{us}$. In addition, the semi-supervised latent code $c_{ss}$ is trained to contain features that we want by using labeled data. For InfoGAN, we cannot predict which feature will be learned from the training, while ss-InfoGAN uses labeled data to control the learned feature by maximizing two mutual informations; one between $c_{ss}$ and the labeled real data to guide $c_{ss}$ to encode label information $y$, and the other between the generated data and $c_{ss}$. By combining the supervised and unsupervised methods, ss-InfoGAN learns the latent code representation more easily with a small subset of labeled data than the fully unsupervised methods of InfoGAN. 


\subsubsection{Examples} \label{decexam}



\begin{figure} [t]
    \centering
    \includegraphics[width=0.7\textwidth]{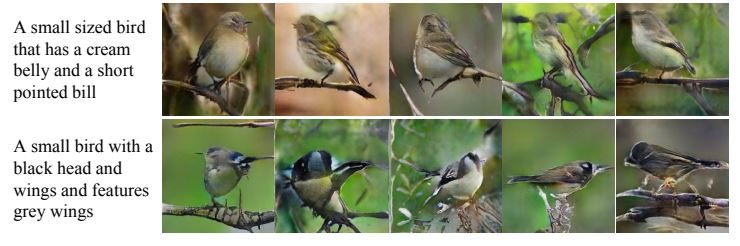}
    \caption{A text to image synthesis of StackGAN \cite{StackGAN}. StackGAN shows higher output diversity than other text to image models. Images from \cite{GOODFELLOWNIPS}.}\label{fig:stackgan}
\end{figure}

Decomposing the latent space into meaningful attributes within the GAN framework has been exploited for various tasks. StackGAN \cite{StackGAN} was proposed for text to image generation that synthesizes corresponding images given text descriptions as shown in Figure~\ref{fig:stackgan}. StackGAN synthesizes images conditioned on text descriptions in a two-stage process: a low-level feature generation (stage 1), and painting details from a given generated image at stage 1 (stage 2). The generators of each stage are trained adversarially given text embedding information $\varphi_t$ from the text $t$. Notably, rather than directly concatenating $\varphi_t$ to $z$ as CGAN does, StackGAN proposes a conditional augmentation technique which samples conditioned text latent vectors $c$ from the Gaussian distribution $\mathcal{N}(\mu(\varphi_t), \sum(\varphi_t))$. By sampling the text embedding vector $c$ from $\mathcal{N}(\mu(\varphi_t), \sum(\varphi_t))$, this technique attempts to augment more training pairs given limited amount of image-text paired data. 


Semantically decomposing GAN (SD-GAN) \cite{SD-GAN} tries to generate a face having different poses by directly decomposing $z$ into identity and pose parts of a face image and then sampling each latent variable separately from the independent latent distributions. Meanwhile, we do not need to restrict the attributes of an image to facial characteristics. Attributes can be not only characteristics of a face but also scenery features such as the weather. \citet{AL-CGAN} synthesized outdoor images having specific scenery attributes using the CGAN framework, and they also concatenated an attribute latent vector to $z$ for the generator. 

\subsection{With an Autoencoder} \label{waa}

In this section, we explore efforts combining an autoencoder structure into the GAN framework. An autoencoder structure consists of two parts: an encoder which compresses data $x$ into latent variable $z$: and a decoder, which reconstructs encoded data into the original data $x$. This structure is suitable for stabilizing GAN because it learns the posterior distribution $p(z|x)$ to reconstruct data $x$, which reduces mode collapse caused by the lack of GAN's inference ability to map data $x$ to $z$. An autoencoder can also help manipulations at the abstract level become possible by learning a latent representation of a complex, high-dimensional data space with an encoder $\mathcal{X}\to \mathcal{Z}$ where $\mathcal{X}$ and $\mathcal{Z}$ denote the data space and the latent space. Learning a latent representation may make it easier to perform complex modifications in the data space through interpolation or conditional concatenation in the latent space. 
We demonstrate how GAN variants learn in the latent space in Section~\ref{latentspaceencoder} and extend our discussion to proposed ideas which combine a VAE framework, another generative model with an autoencoder, with GAN in Section~\ref{vae}.


\subsubsection{Learning the Latent Space} \label{latentspaceencoder}
Adversarially learned inference (ALI) \cite{ALI} and bidirectional GAN (BiGAN) \cite{BiGAN} learn latent representations within the GAN framework combined with an encoder. As seen in Figure \textcolor{red}{10a}, they learn the joint probability distribution of data $x$ and latent $z$ while GAN learns only the data distribution directly. The discriminator receives samples from the joint space of the data $x$ and the latent variable $z$ and discriminates joint pairs $(G(z), z)$ and $(x, E(x))$ where $G$ and $E$ represent a decoder and an encoder, respectively. By training an encoder and a decoder together, they can learn an inference $\mathcal{X}\to \mathcal{Z}$ while still being able to generate sharp, high-quality samples. 

\citet{AGE} proposed a slightly peculiar method in which adversarial learning is run between the generator $G$ and the encoder $E$ instead of the discriminator. They showed that adversarial learning in the latent space using the generator and the encoder theoretically results in the perfect generator. Upon their theorems, the generator minimizes the divergence between $E(G(z))$ and a prior $z$ in the latent space while the encoder maximizes the divergence. By doing so, they implemented adversarial learning together with tractable inference and disentangled latent space without additional computational cost.  

In addition, they added reconstruction loss terms in each space to guarantee each component to be reciprocal. 
These loss terms can be interpreted as imposing each function to be a one to one mapping function so as not to fall into mode collapse as similar to in MRGAN. 
Recall that a geometric regularizing term of MRGAN \cite{MRGAN} also aims to incorporate a supervised training signal which guides the reconstruction process to the correct location.
Figure \textcolor{red}{10b} shows an outline of \citet{AGE} where $R$ in the red rectangle denotes the reconstruction loss term between the original and reconstructed samples.

\begin{figure} [t]
    \centering    
    \includegraphics[width=0.9\textwidth]{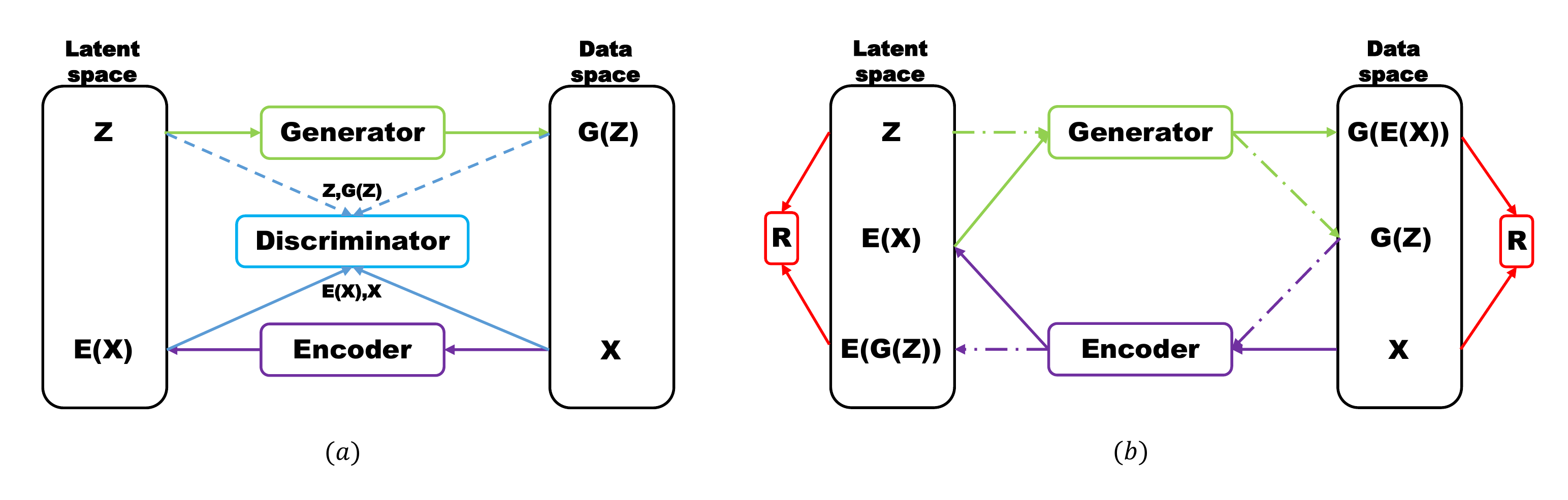}
    \caption{Illustrations of (a) ALI \cite{ALI}, BiGAN \cite{BiGAN}, and (b) AGE \cite{AGE}.}\label{fig:encoder}
    \vspace{-2mm}
\end{figure}

\subsubsection{Variational Autoencoder} \label{vae}
Variational Autoencoder (VAE) \cite{vae} is a popular generative model using an autoencoder framework. Assuming some unobserved latent variable $z$ affects a real sample $x$ in an unknown manner, VAE essentially finds the maximum of the marginal likelihood $p_\theta(x)$ for the model parameter $\theta$. VAE addresses the intractability of $p_\theta(x)$ by introducing a variational lower bound, learning the mapping of $\mathcal{X} \to\mathcal{Z}$ with an encoder and $\mathcal{Z}\to\mathcal{X}$ with a decoder. Specifically, VAE assumes a prior knowledge $p(z)$ and approximated posterior probability modeled by $Q_\phi(z|x)$ to be a standard normal distribution and a normal distribution with diagonal covariance, respectively for the tractability. More explicitly, VAE learns to maximize $p_\theta(x)$ where a variational lower bound of the marginal log-likelihood $\log p_\theta(x)$ can be derived as follows:

\begin{align} 
\log p_\theta(x) &= \int_z Q_\phi(z|x)\log p_\theta(x)dz = \int_z Q_\phi(z|x) \log \left(\frac{p_\theta(x,z)}{p_\theta(z|x)}\frac{Q_\phi(z|x)}{Q_\phi(z|x)}\right)dz \label{marginalfirst} \\
&= \int_z Q_\phi(z|x)\left(\log\left(\frac{Q_\phi(z|x)}{p_\theta(z|x)}\right)+\log\left(\frac{p_\theta(x,z)}{Q_\phi(z|x)}\right)\right)dz \label{marginalsecond} \\
&= KL(Q_\phi(z|x)||p_\theta(z|x))+\E_{Q_\phi(z|x)}\left[\log \frac{ p_\theta(x,z)}{Q_\phi(z|x)}\right] \label{marginallast}
\end{align}
As $KL(Q_\phi(z|x)||p_\theta(z|X))$ is always nonnegative, a variational lower bound $L(\theta, \phi ; x)$ of Equation~\ref{vaelowerlast} can be derived as follows:
\begin{align} 
\log p_\theta(x) &\geq \E_{Q_\phi(z|x)}[\log p_\theta(x,z)-\log Q_\phi(z|x)] \label{vaelowerfirst} \\
&= \E_{Q_\phi(z|x)}[\log p_\theta(z)-\log p_\theta(x|z)-\log Q_\phi(z|x)] \label{vaelowersecond} \\
&= -KL(Q_\phi(z|x)||p_\theta(z)) +\E_{Q_\phi(z|x)}[\log p_\theta(x|z)] \label{vaelowerthird} \\ 
&= L(\theta, \phi;x) \label{vaelowerlast}
\end{align}
where $p_\theta(x|z)$ is a decoder that generates sample $x$ given the latent $z$ and $Q_\phi(z|x)$ is an encoder that generates the latent code $z$ given sample $x$.

Maximizing $L(\theta, \phi;x)$ increases the marginal likelihood $p_\theta(x)$. The first term can be interpreted as leading an encoder $Q_\phi(z|x)$ to be close to a prior probability $p_\theta(z)$. It can be calculated analytically because $Q_\phi(z|x)$ and the prior probability are assumed to follow a Gaussian distribution. The second term can be estimated from the sample using a reparameterization method. To sum up, VAE learns by tuning the parameter of the encoder and the decoder to maximize the lower bound $L(\theta, \phi;x)$.

\paragraph{Hybrid with GAN}

Recently, several approaches to incorporate each advantage of VAE and GAN have been proposed. Although VAE generates blurry images, VAE suffers less from the mode collapse problem because an autoencoder encourages all real samples to be reconstructed. However, GAN generates sharper images than VAE and does not need further constraints on the model while GAN suffers from mode collapse as mentioned in Section~\ref{modecollapse}. In this section, we address two studies which attempt to combine VAE and GAN into one framework.

VAEGAN \cite{VAEGAN} combined VAE with GAN by assigning GAN's generator to a decoder. Its objective function combined VAE's objective function with an adversarial loss term to produce sharp images while maintaining encoding ability for the latent space. Notably, it replaced the reconstruction of $x$ in Equation~\ref{vaelowerlast} with the intermediate features of the discriminator to capture more perceptual similarity of real samples. Figure \textcolor{red}{11a} shows an outline of VAEGAN where the discriminator $D$ takes one real sample and two fake samples where one is sampled from an encoded latent space ($z_{VAE}$) and the other from a prior distribution ($z$).

Variational approaches for autoencoding GAN ($\alpha$-GAN) \cite{alpha-GAN} proposes adopting discriminators for the variational inference and transforms Equation~\ref{vaelowerthird} into a more GAN-like formulation. The most negative aspect of VAE is that we have to constrain a distribution form of $Q_\phi (z|x)$ to analytically calculate $L(\theta, \phi;x)$. $\alpha$-GAN treats the variational posteriori distribution implicitly by using the density ratio technique which can be derived as follows: 
\begin{equation} \label{eq:approxdensityratio}
KL(Q_\phi(z|x)||p_\theta(z))=\E_{Q_\phi(z|x)}\left[\frac{Q_\phi(z|x)}{p_\theta(z)}\right] \approx \E_{Q_\phi(z|x)}\left[\frac{C_\phi(z)}{1-C_\phi(z)}\right]
\end{equation}
where $C_\phi(z) = \frac{Q_\phi(z|x)}{Q_\phi(z|x)+p_\theta(z)}$ which is an optimal solution of the discriminator for a fixed generator in standard GAN~\cite{GAN}. $\alpha$-GAN estimates the KLD term using a learned discriminator from the encoder $Q_\phi(z|x)$ and the prior distribution $p_\theta(z)$ in the latent space. A KLD regularization term of a variational distribution, thus, no longer needs to be calculated analytically.

$\alpha$-GAN also modifies a reconstruction term in Equation~\ref{vaelowerlast} by adopting two techniques. One is using another discriminator which distinguishes real and synthetic samples, and the other adds a normal $l_1$ pixelwise reconstruction loss term in the data space. Specifically, $\alpha$-GAN changes a variational lower bound $L(\theta, \phi;x)$ into a more GAN-like formulation by introducing two discriminators using a density ratio estimation and reconstruction loss to prevent a mode collapse problem. Figure \textcolor{red}{11b} shows an outline of $\alpha$-GAN where $D_L$ is the discriminator of the latent space, and $D_D$ represents the discriminator which acts on the data space and $R$ is the reconstruction loss term.

\begin{figure} [t]
    \centering
    \includegraphics[width=0.9\textwidth]{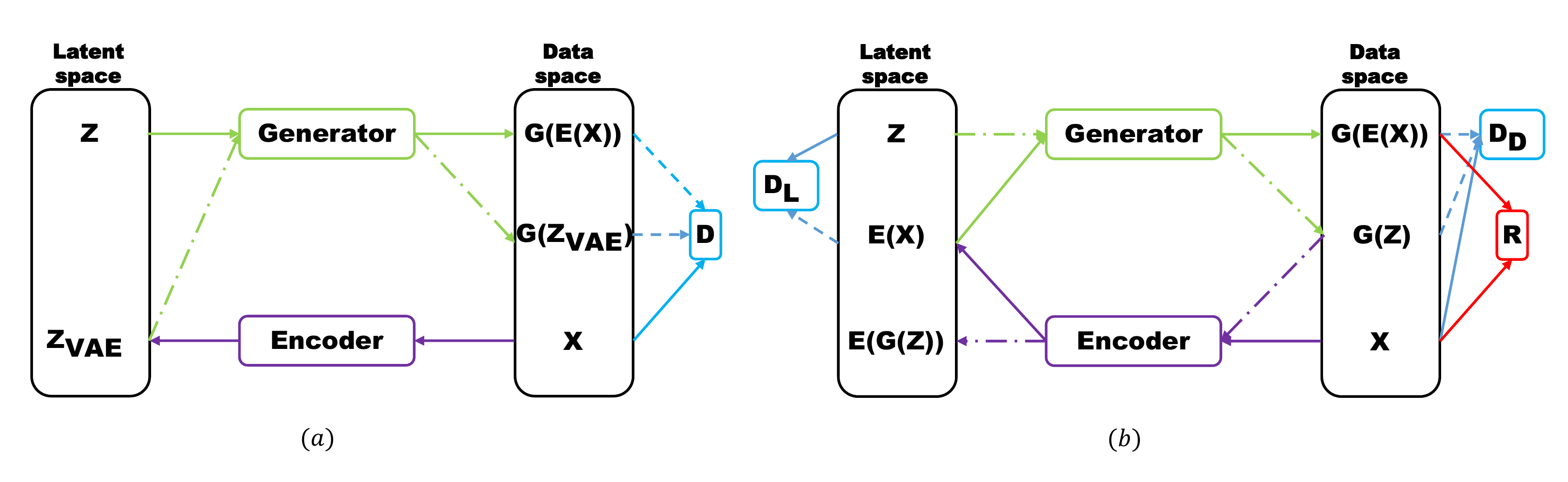}
    \caption{Illustrations of (a) VAEGAN \cite{VAEGAN} and (b) $\alpha$-GAN \cite{alpha-GAN}.}\label{fig:Hybrid with GAN}
\end{figure}

\subsubsection{Examples}
\citet{AGEGAN} proposed an encoder-decoder combined method for the face aging of a person. It produces facial aging of the target image under the given age vector $y$ while maintaining the identity of the target image. The encoder takes a role to output a latent vector $z$ which represents a personal identity to be preserved. However, practically, this CGAN-based approach has a problem in that the generator tends to ignore the latent variable $z$, concerning only conditional information $y$. 

There are some approaches to training $z$ in an unsupervised manner with an auto encoder. Mainly, semi latent GAN (SL-GAN) \cite{SL-GAN} proposed a method for changing a facial image from high-level semantic facial attributes (\ie male/female, skin/hair color), by decomposing the latent space from an encoder into annotated attributes (ground-truth attribute of an image) and data-driven attributes (as $c$ of InfoGAN). Similarly to InfoGAN, SL-GAN tries to maximize the mutual information between the data-driven attribute and the generated image while using unsupervised training for the data-driven attributes. 

In addition to that, disentangled representation GAN (DR-GAN) \cite{DR-GAN} addresses pose-invariant face recognition, which is a difficult problem because of the drastic changes of an image for each different pose, adopting an encoder-decoder structure for the generator. As the purpose of DR-GAN is to generate a face of the same identity given a target pose, it has to learn the identity feature to be invariant regardless of a facial pose. To achieve this, DR-GAN designs an encoder of the generator to represent an identity feature, while a decoder of the generator produces an image under the pose representing vectors and the encoded identity. 



\section{Applications Using GANs} \label{section4}
As discussed in earlier sections, GAN is a very powerful generative model in that it can generate real-like samples with an arbitrary latent vector $z$. We do not need to know an explicit real data distribution nor assume further mathematical conditions. These advantages lead GAN to be applied in various academic and engineering fields. In this section, we discuss applications of GANs in several domains. 
\subsection{Image}
\subsubsection{Image translation}


Image translation involves translating images in one domain $X$ to images in another domain $Y$. Mainly, translated images have the dominant characteristic of domain $Y$ maintaining their attributes in the original images. Image translation can be categorized into supervised and unsupervised techniques such as the classical machine learning.

\paragraph{Paired two domain data} \label{paired}

\begin{figure} [t]
    \centering    
    \includegraphics[width=0.7\textwidth]{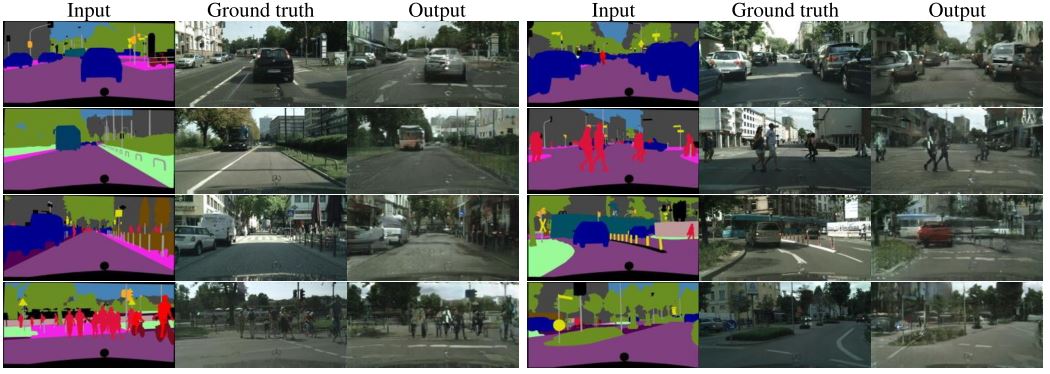}      
    \caption{Paired image translation results proposed by pix2pix \cite{pix2pix}: converting Cityscapes labels into real photo compared to ground truth. Images from pix2pix \cite{pix2pix}.}\label{fig:pix2pix}
\end{figure}

Image translation with paired images can be regarded as supervised image translation in that an input image ${x}$ $\in$ $X$ to be translated always has the target image ${y}$ $\in$ $Y$ where $X$ and $Y$ are two distinctive domains. Pix2pix \cite{pix2pix} suggests an image translation method with paired images using a CGAN framework in which a generator produces a corresponding target image conditioned on an input image as seen in Figure~\ref{fig:pix2pix}. In contrast, Perceptual Adversarial Networks (PAN) \cite{wang2018perceptual} add the perceptual loss between a paired data $(x,y)$ to the generative adversarial loss to transform input image $x$ into ground-truth image $y$. Instead of using the pixelwise loss to push the generated image toward the target image, it uses hidden layer discrepancies of the discriminator between an input image $x$ and ground truth image $y$. It tries to transform $x$ to $y$ to be perceptually similar by minimizing perceptual information discrepancies from the discriminator. 

\paragraph{Unpaired two domain data} \label{unpaired}

Image translation in an unsupervised manner learns a mapping between two domains given unpaired data from two domains. CycleGAN \cite{CycleGAN} and discover cross-domain relations with GAN (DiscoGAN) \cite{DiscoGAN} aim to conduct unpaired image-to-image translations using a cyclic consistent loss term in addition to an adversarial loss term. With a sole translator ${G}$: ${X} \to {Y}$, GAN may learn meaningless translation or mode collapse, resulting in an undesired translation. To reduce the space of mapping of the generator, they adopt another inverse translator {${T}$: ${Y} \to {X}$} and introduce the cyclic consistency loss which encourages ${T}({G}({x})) \approx {x}$ and ${G}({T}({y})) \approx {y}$ so that each translation finds a plausible mapping between the two domains as mentioned in Section~\ref{rof}. Their methods can be interpreted in a similar manner described in Section~\ref{latentspaceencoder} in that they add a supervised signal for reconstruction. 

Attribute guided image translation was also considered to transfer the visual characteristic of an image. Conditional CycleGAN \cite{conditioncycle} utilizes CGAN with a cyclic consistency framework. \citet{conditiondisco} attempted to transfer visual attributes. In addition to the cyclic consistency of an image, they also added an attribute consistency loss which forces the transferred image to have a target attribute of the reference image.

\subsubsection{Super resolution}

Acquiring super resolution images from low resolution images has the fundamental problem that the recovered high-resolution image misses high-level texture details during the upscaling of the image. \citet{SRGAN} adopted a perceptual similarity loss in addition to an adversarial loss, instead of pixelwise mean-squared error loss. It focuses on feature differences from the intermediate layer of the discriminator, not pixelwise because optimizing pixel-wise mean squared error induces the pixelwise average of a plausible solution, leading to perceptually poor smoothed details and it is not robust to drastic pixel value changes. 

\subsubsection{Object detection}

Detecting small objects in an image typically suffers from low-resolution of an object, and thus, it is necessary to train models with images of various scales similar to You Look Only Once (YOLO) \cite{yolo} and Single Shot Detection (SSD) \cite{ssd} methods. Notably, \citet{Perceptual} tries to transform a small object with low resolution into a super resolved large object to make the object more discriminative. They utilized a GAN framework except decomposed the discriminator into two branches, namely, an adversarial branch and a perceptual branch. The generator produces a real-like large-scale object by the typical adversarial branch while the perceptual branch guarantees that the generated large-scale object is useful for the detection.

\citet{SeGAN} proposed another framework to detect objects occluded by other objects in an image. It uses a segmentor, a generator, and a discriminator to extract the entire occluded-object mask and to paint it as a real-like image. The segmentor takes an image and a visible region mask of an occluded object and produces a mask of the entire occluded object. The generator and the discriminator are trained adversarially to produce an object image in which the invisible regions of the object are reconstructed. 

\subsubsection{Object transfiguration}

Object transfiguration is a conditional image generation that replaces an object in an image with a particular condition while the background does not change. \citet{GeneGAN} adopted an encoder-decoder structure to transplant an object, where the encoder decomposes an image into the background feature and the object feature, and the decoder reconstructs the image from the background feature and the object feature we want to transfigure. Importantly, to disentangle the encoded feature space, two separated training sets are required where one is the set of images having the object and the other is the set of images not having the object. 


In addition, the GAN can be applied to an image blending task which implants an object into another image's background and makes the composited copy-paste images look more realistic. Gaussian-Poisson GAN (GP-GAN) \cite{GP-GAN} suggests a high-resolution image blending framework using GAN and a classic image blending gradient-based approach \cite{frankot1988method}. It decomposes images into low-resolution but well-blended images using a GAN and detailed textures and edges using a gradient constraint. Then, GP-GAN attempts to combine the information by optimizing a Gaussian-Poisson equation \cite{poisson} to generate high-resolution well-blended images while maintaining captured high-resolution details. 



\subsubsection{Joint image generation}
The GAN can be utilized to generate multiple domain images at once.
Coupled GAN \cite{CoupledGAN} suggests a method of generating multidomain images jointly by weight sharing techniques among GAN pairs. It first adopts GAN pairs to match the number of domains we want to produce. Then, it shares the weights of some layers of each GAN pair that represents high-level semantics. Therefore, it attempts to learn joint distributions of a multidomain from samples drawn from a marginal domain distribution. It should be noted that because it aims to generate multidomain images which share high-level abstract representations, images from each domain have to be very similar in a broad view.  


\subsubsection{Video generation}

In this paragraph, we discuss GANs generating video. Generally, the video is composed of relatively stationary background scenery and dynamic object motions. Video GAN (VGAN) \cite{VGAN} considers a two-stream generator. A moving foreground generator using 3D CNN predicts plausible future frames while a static background generator using 2D CNN makes the background stationary. Pose-GAN \cite{Pose-GAN} takes a VAE and GAN combining approach. It uses a VAE approach to estimate future object movements conditioned on a current object pose and hidden representations of past poses. With a rendered future pose video and clip image, it uses a GAN framework to generate future frames using a 3D CNN. Recently, motion and content GAN (MoCoGAN) \cite{MocoGAN} proposed to decompose the content part and motion part of the latent space, especially modeling the motion part with RNN to capture the time dependency. 

	
\subsection{Sequential Data Generation} \label{sdg}
GAN variants that generate discrete values mostly borrow a policy gradient algorithm of RL, to circumvent direct back-propagation of discrete values. To output discrete values, the generator, as a function, needs to map the latent variable into the domain where elements are not continuous. However, if we do the back-propagation as another continuous value generating process, the generator is steadily guided to generate real-like data by the discriminator, rather than suddenly jumping to the target discrete values. Thus, such a slight change of the generator cannot easily look for a limited real discrete data domain \cite{SeqGAN}.

In addition, when generating a sequence such as music or language, we need to evaluate a partially generated sequence step-by-step, measuring  the performance of the generator. However, the conventional GAN framework can only evaluate whole generated sequences unless there is a discriminator for each time-step. This, too, can be solved by the policy gradient algorithm, in that RL naturally addresses the sequential decision process of the agent.

\subsubsection{Music}
When we want to generate music, we need to generate the note and a tone of the music step-by-step, and these elements are not continuous values. A simple and direct approach is continuous RNN-GAN (C-RNN-GAN) \cite{C-RNN-GAN}, where it models both the generator and discriminator as an RNN with long-short term memory (LSTM) \cite{lstm}, directly extracting whole sequences of music. However, as mentioned above, we can only evaluate whole sequences, and not a partially generated sequence. Furthermore, its results are not highly satisfactory since it does not consider the discrete property of the music elements.

\begin{figure} [t]
\centering
\includegraphics[width=0.9\textwidth]{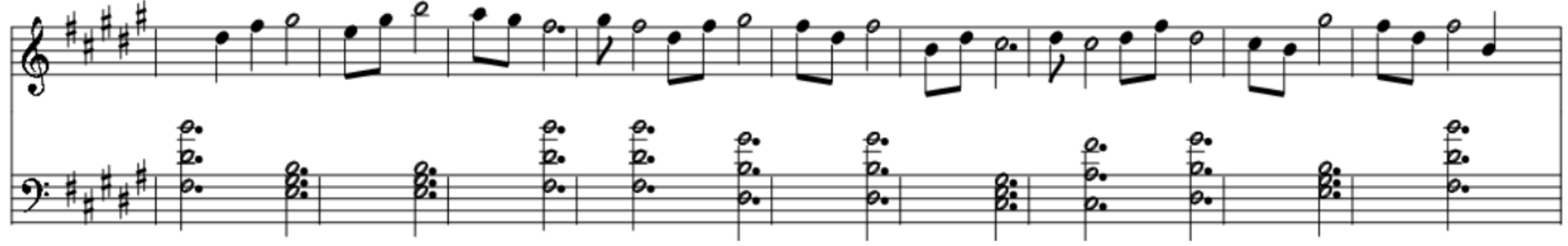}
\caption{\label{fig:music}Polyphonic music sequences generated from SeqGAN \cite{leeseqgan}.}
\end{figure}

In contrast, sequence GAN (SeqGAN) \cite{SeqGAN}, object reinforced GAN (ORGAN) \cite{ORGAN}, and \citet{leeseqgan} employed a policy gradient algorithm, and not generating whole sequences at once. The result of SeqGAN is shown in Figure~\ref{fig:music}. They treat a generator's output as a policy of an agent and take the discriminator's output as a reward. Selecting a reward with the discriminator is a natural choice as the generator acts to obtain a large output (reward) from the discriminator, similar to the agent learning to acquire a large reward in reinforcement learning. In addition, ORGAN is slightly different from SeqGAN, adding a hard-coded objective to the reward function to achieve the specified goal.

\subsubsection{Language and speech}

RankGAN \cite{RANKGAN} suggests language (sentence) generation methods and a ranker instead of a conventional discriminator. In natural language processing, the expression power of natural language needs to be considered in addition to its authenticity. Thus, RankGAN adopts a relative ranking concept between generated sentences and reference sentences which are human-written. The generator tries its generated language sample to be ranked high, while the ranker evaluates the rank score of the human-written sentences higher than the machine-written sentences. As the generator outputs discrete symbols, it similarly adopts a policy gradient algorithm similar to SeqGAN and ORGAN. In RankGAN, the generator can be interpreted as a policy predicting next step symbol and the rank score can be thought of as a value function given a past generated sequence.

Variational autoencoding Wasserstein GAN (VAW-GAN) \cite{VAW-GAN} is a voice conversion system combining GAN and VAE frameworks. The encoder infers a phonetic content $z$ of the source voice, and the decoder synthesizes the converted target voice given a target speaker's information $y$, similar to conditional VAE \cite{cvae}. As discussed in Section~\ref{vae}, VAE suffers from generating sharp results due to the oversimplified assumption of the Gaussian distribution. To address this issue, VAW-GAN incorporates WGAN \cite{WGAN} similarly to VAEGAN \cite{VAEGAN}. By assigning the decoder to the generator, it aims to reconstruct the target voice given the speaker representation. 

\subsection{Semi-Supervised Learning} \label{ssl}

Semi-supervised learning is a learning method that improves the classification performance by using unlabeled data in a situation where there are both labeled and unlabeled data. In the big data era, there exists a common situation where the size of the data is too large to label all the data, or the cost of labeling is expensive. Thus, it is often necessary to train the model with a dataset in which only a small portion of the total data has labels.


\subsubsection{Semi-supervised learning with a discriminator}

The GAN-based semi-supervised learning method \cite{Improved} demonstrates how unlabeled and generated data are available on the GAN framework. The generated data is allocated to a $K+1$ class beyond $1 \dots K$ class for the labeled data. For labeled real data, the discriminator classifies their correct label ($1$ to $K$). For unlabeled real data and generated data, they are trained with a GAN minimax game. Their training objective can be expressed as follows:
\begin{gather}
L = L_{s} + L_{us} \\
L_{s} = -\mathbb{E}_{x,y\sim p_{\mathrm{data}}(x,y)}[\log p_\theta(y|x,y<K+1)] \\
L_{us} = -\mathbb{E}_{x\sim p_{\mathrm{data}}(x)}[1-\log p_\theta(y=K+1|x)] +\mathbb{E}_{x\sim G}[\log p_\theta(y=K+1|x)] 
\end{gather}
where $L_{s}$ and $L_{us}$ stand for the loss functions of the labeled data and the unlabeled data, respectively. It is noted that because only generated data is classified as the $K+1$ class, we could think of $L_{us}$ as a GAN standard minimax game. The unlabeled data and the generated data serve to inform the model of the space where the real data resides. In other words, the unsupervised cost serves to guide the location of the optimum of the supervised cost of the real labeled data.

Categorical GAN (CatGAN) \cite{CatGAN} proposes an algorithm for the robust classification for which the generator regularizes the classifier. The discriminator has no classification head for distinguishing real and fake and is trained with three requirements: a small conditional entropy of $H(y|x)$ to make the correct class assignment for the real data, a large conditional entropy of $H(y|G(z))$ to make the class assignment for the generated data diverse and a large entropy of $H(y)$ to make a uniform marginal distribution with an assumption of a uniform prior $p(y)$ over classes where $x$, $y$ and $G(z)$ are real data, labels, and generated data respectively. The generator, meanwhile, is trained with two requirements: a small conditional entropy of $H(y|G(z))$ to make the class assignment for the generated data certain and a large entropy of $H(y)$ to generate equally distributed samples over classes. The unlabeled data and the generated data help the classification by balancing classes through the adversarial act of the generator, so it helps semi-supervised learning. 

\subsubsection{Semi-supervised learning with an auxiliary classifier}
\begin{figure} [t]
\centering
\includegraphics[width=0.75\textwidth]{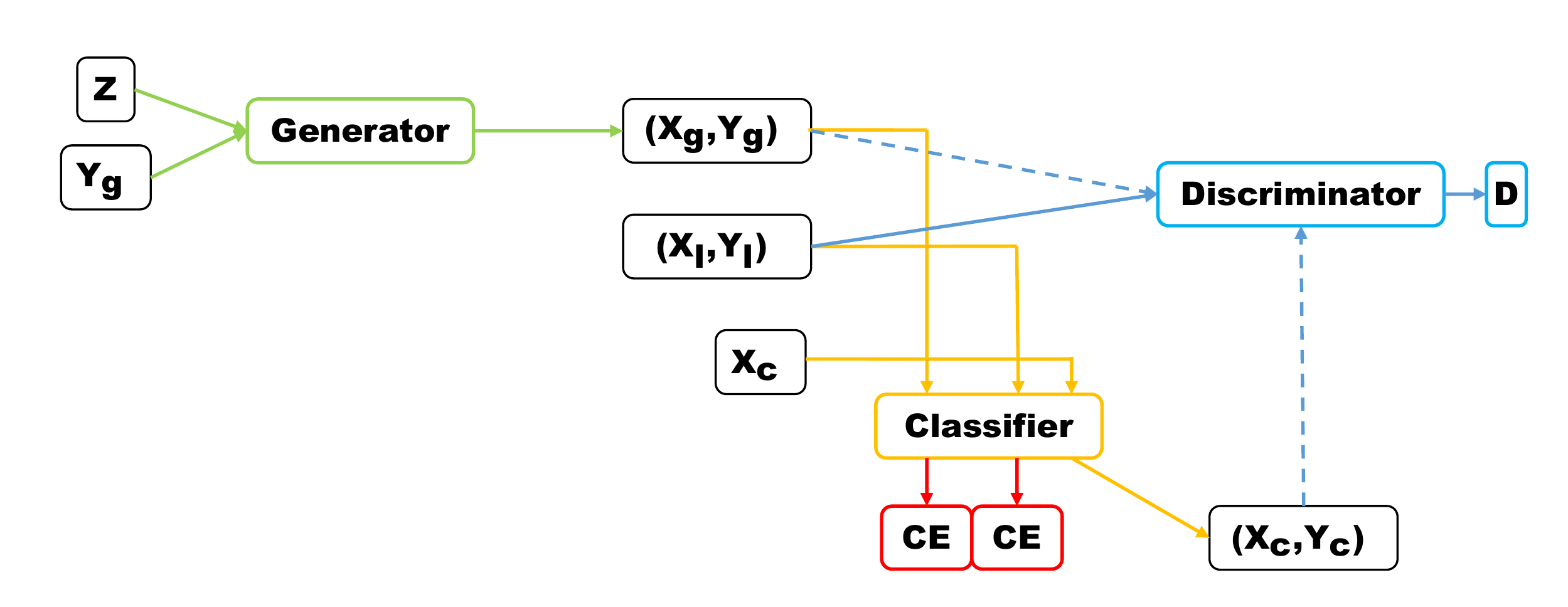}
\caption{\label{fig:triple}An illustration of Triple-GAN \cite{triple}.}
\end{figure}

The above GAN variants in semi-supervised learning have two problems: the first one is that the discriminator has two incompatible convergence points, one for discriminating real and fake data and the other for predicting the class label; and the second problem is that the generator cannot generate data in a specific class. Triple-GAN \cite{triple} addresses the two problems by a three-player formulation: a generator $G$, a discriminator $D$ and a classifier $C$. The model is illustrated in Figure~\ref{fig:triple} where $(X_g, Y_g)\sim p_g(X, Y)$, $(X_l, Y_l) \sim p(X, Y)$, and $(X_c, Y_c) \sim p_c(X, Y)$ refer to the generated data, the labeled data, and the unlabeled data with a predicted label, and $CE$ is the cross entropy loss. To summarize, Triple-GAN adopts an auxiliary classifier which classifies real labeled data and label-conditioned generated data, relieving the discriminator of classifying the labeled data. In addition, Triple-GAN generates data conditioned on $Y_g$, which means that it can generate label-specific data.



\subsection{Domain Adaptation} \label{domainadaptation}

Domain adaptation is a type of transfer learning which tries to adapt data from one domain (\ie the source domain) into another domain (\ie the target domain), while the classification task performance is preserved in the target domain \cite{pan2010survey}. Formally, the unsupervised domain adaptation solves the following problem: with input data $x$ and its label $y$, let a source domain distribution $\mathbb{D}_{S}(x,y)$ and a target domain distribution $\mathbb{D}_{T}(x,y)$ be defined over $X\times Y$ where $X$ and $Y$ are sets of a data space and a label space respectively. Given labeled source domain data {$(x_s, y_s)$ $\in \mathbb{D}_{S}(x,y)$}\ and unlabeled target domain data ${x_t} \in \mathbb{D}_{T}(x)$ (the marginal distribution of $\mathbb{D}_T(x,y)$), the unsupervised domain adaptation aims to learn a function ${h} : {X} \rightarrow {Y}$ which classifies well in the target domain without the label information of target domain data $x_t$. 


\subsubsection{Domain adaptation by feature space alignment via GAN}
The main difficulty in domain adaptation is the difference between the source distribution and the target distribution, called domain shift. This domain shift allows the classifier trained with only the source data to fail in the target domain. One of the methods to address the domain shift is to project each domain data into the common feature space where the distributions of the projected data are similar. There have been a few studies to achieve the common feature space via GAN for the domain adaptation task.

\begin{figure} [t]
    \centering
    \includegraphics[width=0.9\textwidth]{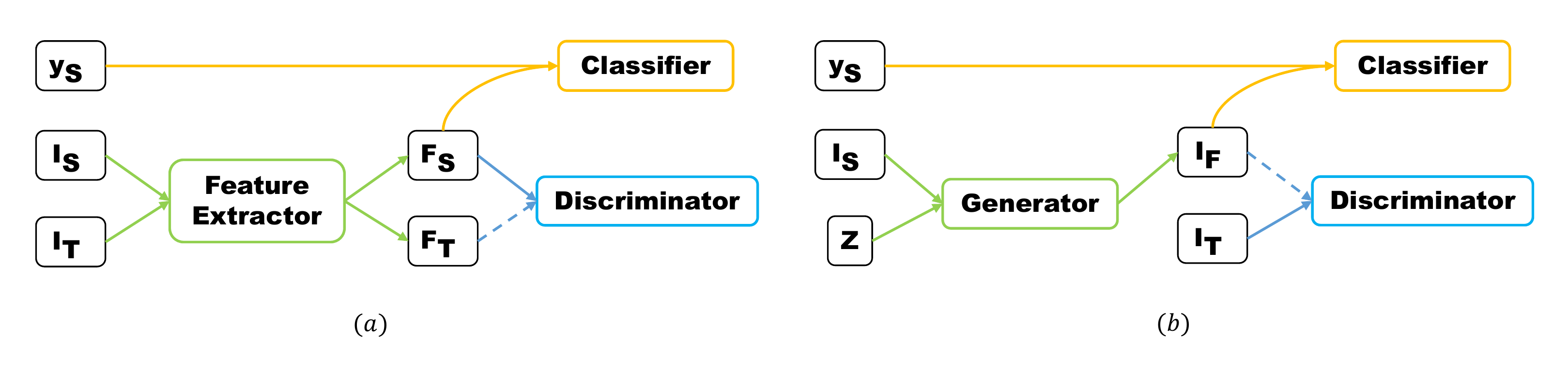}
    \caption{Illustrations of (a) DANN \cite{DANN} and (b) unsupervised pixel-level domain adaptation \cite{Unsupervised}.}\label{fig:DA outline}
\end{figure}

Domain adversarial neural network (DANN) \cite{DANN} first used GAN to obtain domain invariant features by making them indistinguishable as to whether it comes from the source domain or the target domain while still discriminative for the classifying task. There are two components sharing a feature extractor. One is a classifier which classifies the labels of the data. The other is a domain discriminator which discern where the data comes from. The feature generator acts like the generator in GAN producing source-like features from the target domain. To sum up, DANN takes CNN classification networks in addition to the GAN framework to classify data from the target domain without label information by learning the domain invariant features. Figure~\ref{fig:DA outline} shows the overall outline of DANN, ARDA and unsupervised pixel-level domain adaptation where $I_S, I_T$, and $I_f$ stands for the source domain image, target domain image, and fake image, respectively. $F_S$ and $F_T$ means the extracted source features, and the target features, respectively. It should be noted that $Y_s$, which is the label information of $I_s$, is fed into the classifier, training the classifier with the cross entropy loss.



Although DANN may achieve domain invariant marginal feature distributions, if the label of the source feature and that of the target feature do not contain a mismatch, the learned classifier should not work well in the  target domain. Cycle-consistent adversarial domain adaptation (CyCADA)~\cite{CyCADA}, thus, adds a cycle consistency~\cite{CycleGAN,DiscoGAN} to preserve the content of the data which is the crucial characteristic in determining the label while inheriting the most of the architecture of DANN.

\subsubsection{Examples}
\citet{Unsupervised} used the domain adaptation via GAN for the grasping task. They used simulation data as the source domain data and ran the learned model in the real environment. Their method is slightly different than DANN in that it only adapts source images to be seen as if they were drawn from the target domain while DANN tries to obtain features similarly from both domains. It can be understood that the feature space of \cite{Unsupervised} is the target domain space and the feature generator in the target domain is the identity function. In addition, they added a content similarity loss defined as the pixelwise mean-squared error between the source image and the adapted image to preserve the content of the source image. By doing so, they achieved better performance than the supervised method in the grasping task. It should also be noted that the method of \cite{Unsupervised} can check whether the domain adaptation process is working well during the training phase because the transformed data is visible, while \cite{DANN,ARDA} cannot visually check the domain adaptation process because the common representation space cannot be easily visualized.

\begin{figure} [t]
    \centering
    \includegraphics[width=0.7\textwidth]{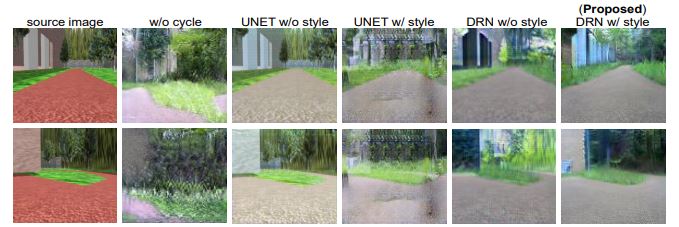}
    \caption{The source images and transferred images by five models proposed by \citet{AUNDA}. The leftmost images are source images made by a simulator, and figures of 2-6 columns are transferred images by five different models. Images from \citet{AUNDA}.}\label{fig:yoo}
\end{figure}

\citet{AUNDA} achieved autonomous navigation without any real labeled data using domain adaptation. As in \cite{Unsupervised}, they also exploited the simulation data as the source data and showed autonomous navigation in a real outdoor environment. They used cycle consistency to preserve the content and style loss term motivated by the style transfer task~\cite{styletransfer} to reduce the domain shift dramatically for the outdoor environment as seen in Figure~\ref{fig:yoo}. By doing so, they showed the applicability of the simulation via domain adaptation into the autonomous navigation tasks where collecting labels for various environments is difficult and expensive.

\subsection{Other Tasks}

Several variants of GAN have also been developed in other academic or practical fields other than the machine learning fields. 

\subsubsection{Medical Image Segmentation}
\citet{Medical} proposed a segmentor-critic structure to segment a medical image. A segmentor generates a predicted segmented image, and a critic maximizes the hierarchical feature differences between the ground-truth and the generated segmentation. This structure leads a segmentor to learn the features of the ground-truth segmentation adversarially similar to the GAN approach. There are also other medical image segmentation algorithms such as the deep image-to-image network (DI2IN) \cite{DI2IN} and structure correcting adversarial network (SCAN) \cite{SCAN}. DI2IN conducts liver segmentation of 3D CT images through adversarial learning. SCAN tries to segment the lung and the heart from chest X-ray images through an adversarial approach with a ground-truth segmentation mask. 

\subsubsection{Steganography}
It is also feasible to use GAN for steganography. Steganography is a technique that conceals secret messages in non-secret containers such as an image. A steganalyzer determines if a container contains a secret message or not. Some studies such as \citet{SteganographyGAN}, and \citet{SSGAN} propose a steganography model with three components: a generator producing real-looking images that are used as containers and two discriminators, one of which classifies whether an image is real or fake, and the other determines whether an image contains a secret message. 

\subsubsection{Continual Learning}
Deep generative replay \cite{Continual} extends a GAN framework to continual learning. Continual learning solves multiple tasks and accumulates new knowledge continually. Continual learning in deep neural networks suffers from catastrophic forgetting which refers to forgetting a previously learned task while learning a new task. Inspired by the brain mechanism, catastrophic forgetting is addressed with a GAN framework called deep generative replay. Deep generative replay trains a scholar model in each task where a scholar model is composed of a generator. The generator produces samples of an old task, and the solver gives a target answer to an old task's sample. By sequentially training scholar models with old task values generated by old scholars, it attempts to overcome catastrophic forgetting while learning new tasks.

\section{Discussion} \label{section5}
We have discussed how GAN and its variants work and how they are applied to various applications. Table~\ref{tab:comparison} compares some famous variants of GAN with respect to model architectures and additional constraints. As we have viewed GAN from a microscopic perspective until now, we are going to discuss the macroscopic view of GAN in this section.

\begin{table*} [t]
\centering
    \caption{\label{tab:comparison}Comparison of GAN variants from some aspects.}
		\begin{tabular}{c|l|l|p{3.6cm}}
			& Generator & Discriminator & Additional loss \par \& Constraint \tabularnewline
			\Xhline{3\arrayrulewidth}
			WGAN-GP	& ReLU MLP & ReLU MLP & gradient penalty \\\hline
			BEGAN & Discriminator decoder &	Autoencoder \scriptsize{(ELU CNN)} & equilibrium measure \\\hline
            ACGAN & Transposed ReLU CNN	& LeakyReLU CNN & classification loss \\\hline
            SeqGAN & LSTM & ReLU CNN & policy gradient \\\hline
            DANN & ReLU CNN & ReLU MLP & classification loss, \par gradient reversal layer
		\end{tabular}
\vspace{-4mm}
\end{table*}

\subsection{Evaluation}

Measuring the performance of GAN is related to capturing the diversity and quality of generated data. As explained in Section~\ref{intro}, the generative models mostly model the likelihood and learn by maximizing it. Thus, it is natural to evaluate the generator using log likelihood. However, GAN generates real-like data without estimating likelihood directly, so evaluation with the likelihood is not quite proper. The most widely accepted evaluation metric for GAN is an Inception score. The Inception score was proposed by \citet{Improved}, and the Inception score is defined as follows:
\begin{equation} \label{inception}
\exp(\E_{x\sim p_{\mathrm{g}}(x)}[KL(p(y|x)||p(y))])
\end{equation}
As seen in Equation~\ref{inception}, it computes the average KLD between the conditional label distribution $p(y|x)$ and the marginal distribution of the generated data's label $p(y)$ with a pretrained classifier such as VGG \cite{vgg} and ImageNet data \cite{krizhevsky2012imagenet}. Since the KLD term between $p(y|x)$ and $p(y)$ in the exponent function is equivalent to their mutual information $I(y;x) = H(y) - H(y|x)$ \cite{barratt2018note}, the high entropy of $p(y|x)$ and low entropy of $p(y)$ leads to a high Inception score. The entropy of $p(y|x)$ measures how the generated data are sharp and clear to be well-classified. The other term $H(y)$ represents whether the generated data are diverse with respect to the generated class. In this way, the Inception score is believed to measure the diversity and visual quality of the generated data.

However, the Inception score has a limitation in that it is unable to resist the mode collapse problem. Even though the trained generator produces only one plausible data for each class, $p(y|x)$ can have low entropy leading to a high Inception score. To address this issue, AC-GAN \cite{AC-GAN} adopts multi-scale structural similarity (MS-SSIM) \cite{msssim}, which evaluates perceptual image similarity, and thus identifies a mode collapse more reliably. In addition, MRGAN \cite{MRGAN} proposed a MODE score based on the Inception score and \citet{independentw} suggested an independent Wasserstein critic which is trained independently for the validation dataset to measure overfitting and mode collapse. Moreover, various object functions such as MMD, total variance and Wasserstein distance can be used as the approximated distance between $p_{\mathrm{g}}(x)$ and $p_{data}(x)$. Theoretically, all metrics should produce the same result under the assumption of the full capacity of the model and an infinite number of training samples. However, \citet{evaluation} showed that minimizing MMD, JSD or KLD results in different optimal points, even for the mixture of Gaussian distributions as a real data distribution, and the convergence in one type of distance does not guarantee the convergence for another type of distance because their object functions do not share a unique global optima.

The method for measuring the performance of GAN is still a disputable subject. Since GAN is naturally unsupervised learning, we cannot measure the accuracy or error rate as in the supervised learning approaches. Evaluation metrics or different distances discussed in the above paragraph still do not measures the performance of GAN exactly, and there are many cases in which images that do not look natural have a high score \cite{barratt2018note}. Thus, there is room to improve the evaluation for GAN.

\subsection{Discrete Structured Data}

Unlike other generative models such as VAE, GAN has an issue handling the discrete data such as text sequences or discretized images. Since discrete data is non-differentiable, gradient descent update via back-propagation cannot directly be applied for a discrete output. The content of this section may overlap with Section~\ref{sdg}, but we will shortly cover this issues in the discussion because it is one of the troublesome issue in GAN. 

To address this issue, some methods adopt a policy gradient algorithm \cite{PG} in reinforcement learning (RL) in which the objective is to maximize the total rewards. By rolling out whole sequences, this method circumvents direct back-propagation for a discrete output. Intuitively, the generator which generates fake data maximizing the discriminator's output can be thought of as a policy agent in RL which is a probability distribution to take action maximizing a reward in a given state. 

Maximum-likelihood augmented discrete GAN (MLADGAN) \cite{MALIGAN} and boundary-seeking GAN (BSGAN) \cite{BSGAN} both take a policy gradient algorithm to generate discrete structured data. They both treat the generator as a policy which outputs a probability for a discrete output. Importantly, they estimate the true distribution as $\tilde{p}_{data}(x)=\frac{1}{Z}p_{\mathrm{g}}(x)\frac{D(x)}{1-D(x)}$ where $Z$ is a normalization factor. $\tilde{p}_{data}(x)$ is motivated from $p_{\mathrm{data}}(x)=\frac{D^*(x)}{1-D^*(x)}p_{\mathrm{g}}(x)$ where $D^*$ is the optimal discriminator, and MLADGAN and BSGAN interpret $\tilde{p}_{data}(x)$ as a reward for a discrete output $x$. Because larger $D(x)$ means greater $\tilde{p}_{data}(x)$, taking the $\tilde{p}_{data}(x)$ value as a reward brings about the same consequence as when $D(x)$ is taken as a reward. They showed some successful performance, but the learning process is highly unstable due to MCMC sampling in the training phase.

Adversarially regularized autoencoders (ARAE) \cite{ARAE} address this issue by combining a discrete autoencoder which encodes a discrete input into the continuous code $c$. ARAE also adopts WGAN \cite{WGAN} acting on the latent space where encoded code lies. To avoid direct access to the discrete structure, the generator produces fake code $\tilde{c}$ from the sampled latent $z$, and the critic evaluates such fake code $\tilde{c}$ and real code $c$. By jointly training an autoencoder with WGAN on the code space, it avoids back-propagation on the discrete space while learning latent representations of discrete structured data.

\subsection{Relationship to Reinforcement Learning} \label{connectiontorl}
Reinforcement learning (RL) is a type of learning theory that focuses on teaching an agent to choose the best action given a current state. A policy $\pi(a|s)$, which is a probability for choosing an action $a$ at state $s$, is learned via an on-policy or off-policy algorithm. RL has a very similar concept to GAN in the aspect of a policy gradient where it is very important to estimate the value function correctly given state $s$ and action $a$.

The policy gradient algorithm \cite{PG} presented an unbiased estimation of a policy gradient under the correct action-state value $Q(s,a)$. Because $Q(s,a)$ is the discounted total reward from state $s$ and action $a$, $Q(s,a)$ must be estimated via an estimator called a critic. A wide variety of policy gradients such as the deep deterministic policy gradient (DDPG) \cite{ddpg}, trust region policy optimization (TRPO) \cite{trpo}, and Q-prop \cite{qprop} can be thought of as estimating $Q(s,a)$ which has low bias and low variance, to correctly estimate the policy gradient. In that regard, they are similar to a GAN framework in that GAN's discriminator estimates the distance between two distributions and the approximated distance needs to be highly unbiased to make $p_{\mathrm{g}}(x)$ closely approximate $p_{data}(x)$. \citet{connectrl} detailed the connection between GAN and the actor-critic methods.

Inverse reinforcement learning (IRL) \cite{irl} is similar to reinforcement learning in that its objective is to find the optimal policy. However, in the IRL framework, the experts' demonstrations are provided instead of a reward. It finds the appropriate reward function that makes the given demonstration as optimal as possible and then produces the optimal policy for the identified reward function. There are many variants in IRL. Maximal entropy IRL \cite{maximum} is one which finds the policy distribution that satisfies the constraints so that the feature expectations of the policy distribution and the given demonstration are the same. To solve such an ill-posed problem, maximal entropy IRL finds the policy distribution with the largest entropy according to the maximal entropy principle \cite{jaynes}. Intuitively, the maximal entropy IRL finds the policy distribution which maximizes the likelihood of a demonstration and its entropy. Its constraint and convexity induce the dual minimax problem. The dual variable can be seen as a reward. The minimax formulation and the fact that it finds the policy which has the largest likelihood of demonstrations gives it a deep connection with the GAN framework. The primal variable is a policy distribution in IRL whereas it can be considered as a data distribution from the generator in GAN. The dual variable is a reward/cost in IRL while it can be seen as the discriminator in GAN. 

\citet{finnirl}, \citet{yooirl}, and \citet{jonathanirl} showed a mathematical connection between IRL and GAN. \citet{jonathanirl} converted IRL to the original GAN by constraining the space of dual variables and \citet{yooirl} showed the relationship between EBGAN \cite{EBGAN} and IRL using approximate inference.

\subsection{Pros and Cons of GAN}
\subsubsection{Pros}
As briefly mentioned in Section~\ref{intro}, the major advantage of GAN is that GAN does not need to define the shape of the probability distribution of the generator model. Thus, GAN naturally avoids concerning tractable density forms which need to represent complex and high-dimensional distributions. Compared to other models using explicitly defined probability density \cite{vae, pixelcnn, pixelrnn, Wavenet}, GAN has following advantages:

\begin{itemize}
\item GAN can parallelize the sampling of the generated data. In the case of PixelCNN \cite{pixelcnn}, PixelRNN \cite{pixelrnn} and WaveNet \cite{Wavenet}, their speed of generation is very slow due to their autoregressive fashion, wherein $p_{\mathrm{g}}(x)$ is decomposed into a product of conditional distributions given previously generated values:
\begin{align}
p_{\mathrm{g}}(x) = \prod_{i=1}^d p_{\mathrm{g}}(x_i | x_{1}, \dots x_{i-1})
\end{align}
where $x$ is the d-dimensional vector. For example, in image generation, autoregressive models generate an image pixel by pixel where the probability distribution of future pixel cannot be inherently computed until the value of the previous pixel is computed. Thus, the generation process is naturally slow, which becomes more severe for high-dimensional data generation such as speech synthesis \cite{Wavenet}.

On the other hand, the generator of GAN is a simple feed-forward network mapping from $\mathcal{Z}$ to $\mathcal{X}$. The generator produces data all at once, not pixel by pixel as autoregressive models. Therefore, GAN can generate samples in parallel, which results in a considerable speed up for sampling, and this property gives more opportunity for GAN to be used in various real applications.

\item GAN does not need to approximate a likelihood by introducing a lower bound, as in VAE. As we mentioned in Section~\ref{vae}, VAE tries to maximize a likelihood by introducing a variational lower bound. The strategy of VAE is to maximize a tractable variational lower bound, guaranteeing it to be at least as high as the lower bound, even when the likelihood is intractable. However, VAE still needs assumptions on a prior and posterior distributions, which do not guarantee the tight bound of Equation~\ref{vaelowerlast}. This strong assumption on distributions makes the approximation to the maximum likelihood biased.

In contrast, GAN does not approximate the likelihood and does not need any probability distribution assumptions. Instead, GAN is designed to solve an adversarial game between the generator and the discriminator, and a Nash equilibrium of the GAN game corresponds to finding the real data distribution \cite{GOODFELLOWNIPS}.

\item GAN is empirically known to produce better and sharper result than other generative models, especially VAE. In VAE, a distribution of pixel values in the reconstructed image is modeled as a conditional Gaussian distribution. This causes the optimization of $\log p_{\mathrm{g}}(x|z)$ to be equivalent to minimizing the Euclidean term of $-\Vert x-Decoder(z)\Vert^2$, which can be interpreted as a regression problem fitting the mean. 

GAN is highly capable of capturing the high-frequency parts of an image. Since the generator tries to fool the discriminator to recover the real data distribution, the generator evolves to lead even the high-frequency parts to deceive the discriminator. In addition, some techniques such as PatchGAN in Section~\ref{ttitg} helps GAN produce and capture sharper results more effectively.
\end{itemize}

\subsubsection{Cons}
GAN was developed to solve the minimax game between the generator and the discriminator. Though several studies discuss the convergence and the existence of the Nash equilibrium of the GAN game, training of GAN is highly unstable and difficult to converge as mentioned in Section~\ref{theoissue} and \ref{pracissue}. GAN solves the minimax game through the gradient descent method iteratively for the generator and the discriminator. In perspective of the cost function: $V(G,D)$ in Equation~\ref{eq:gan}, a solution for the GAN game is the Nash equilibrium which is a point of parameters where the discriminator's cost and the generator's cost is minimum with respect to their parameters. However, the decrease of the discriminator's cost function can cause the increase of the generator's cost function and vice versa. Thus, a convergence of the GAN game may often fail and is prone to be unstable.

Another important issue for GAN is the mode collapse problem. This problem is very detrimental for GAN that is applied in real applications since a mode collapse restricts GAN's ability of diversity. In Equation~\ref{eq:gan}, the generator is only forced to deceive the discriminator, not for representing multimodality of a real data distribution. A mode collapse thus can happen even in a simple experiment \cite{MRGAN}, and this discourages applying GAN due to the low diversity. As mentioned in Section~\ref{modecollapse}, various studies tried to address the mode collapse by using a new object function \cite{MRGAN,DRAGAN}, or adding new components \cite{MAD-GAN,MRGAN}. However, for a highly complex and multimodal real data distribution, the mode collapse still remains a problem GAN has to solve.


\subsubsection{Future research areas}

As GAN has been popular throughout deep learning, the limitations of GAN mentioned above have recently been improved \cite{miyato2018spectral,optimism,ambientgan}.  With the development of GAN, new tasks are steadily conquered using GAN. For instance, CausalGAN \cite{CausalGAN} combines a causal implicit generative model with the conditional GAN to replace conditioning by intervention, which enables the model to generate data with the desired characteristic combinations that do not exist in the dataset. In addition, new types of GANs have been proposed for new applications such as cipher cracking \cite{CipherGAN} and object tracking \cite{song2018vital}. When you design a GAN for a new task, you can first identify the nature of the task and then use Tables~\ref{tab:overview1}, \ref{tab:overview_app}, and \ref{tab:comparison} to determine which model to use as a baseline. A new loss function can be designed using the characteristics of the task. In the future, we anticipate that the limitations of existing GANs will be solved in novel ways, and GAN will remain as an important generative model by conquering areas that existing deep learning-based models cannot effectively solve.

\section{Conclusion} \label{section6}

We discussed how various object functions and architectures affect the behavior of GAN and the applications of GAN such as image translation, image attribute editing, domain adaptation, and other fields. The GAN originated from the theoretical minimax game perspective. In addition to the standard GAN \cite{GAN}, practical trials as well as mathematical approaches have been adopted, resulting in many variants of GAN. Furthermore, the relationship between GAN and other concepts such as imitation learning, and other generative models has been discussed and combined in various studies, resulting in rich theory and numerous application techniques. GAN has the potential to be applied in many application domains including those we have discussed. Despite GAN's significant success, there remain unsolved problems in the theoretical aspects as to whether GAN actually converges and whether it can perfectly overcome mode collapse, as \citet{MIXGAN}, \citet{chekhov}, and \citet{numerics} discussed. However, with the power of deep neural networks and with the utility of learning a highly non-linear mapping from latent space into data space, there remain enormous opportunities to develop GAN further and to apply GAN to various applications and fields.

\bibliographystyle{plainnat}
\bibliography{main}

\begin{thebibliography}{146}
\providecommand{\natexlab}[1]{#1}
\providecommand{\url}[1]{\texttt{#1}}
\expandafter\ifx\csname urlstyle\endcsname\relax
  \providecommand{\doi}[1]{doi: #1}\else
  \providecommand{\doi}{doi: \begingroup \urlstyle{rm}\Url}\fi

\bibitem[Abbeel and Ng(2011)]{irl}
Pieter Abbeel and Andrew~Y Ng.
\newblock Inverse reinforcement learning.
\newblock In \emph{Encyclopedia of machine learning}, pages 554--558. Springer,
  2011.

\bibitem[Ajakan et~al.(2014)Ajakan, Germain, Larochelle, Laviolette, and
  Marchand]{DANN}
Hana Ajakan, Pascal Germain, Hugo Larochelle, Fran{\c{c}}ois Laviolette, and
  Mario Marchand.
\newblock Domain-adversarial neural networks.
\newblock \emph{arXiv preprint arXiv:1412.4446}, 2014.

\bibitem[Antipov et~al.(2017)Antipov, Baccouche, and Dugelay]{AGEGAN}
Grigory Antipov, Moez Baccouche, and Jean-Luc Dugelay.
\newblock Face aging with conditional generative adversarial networks.
\newblock \emph{arXiv preprint arXiv:1702.01983}, 2017.

\bibitem[Arjovsky and Bottou(2017)]{Towards}
Martin Arjovsky and L{\'e}on Bottou.
\newblock Towards principled methods for training generative adversarial
  networks.
\newblock \emph{arXiv preprint arXiv:1701.04862}, 2017.

\bibitem[Arjovsky et~al.(2017)Arjovsky, Chintala, and Bottou]{WGAN}
Martin Arjovsky, Soumith Chintala, and L{\'e}on Bottou.
\newblock Wasserstein gan.
\newblock \emph{arXiv preprint arXiv:1701.07875}, 2017.

\bibitem[Arora et~al.(2017)Arora, Ge, Liang, Ma, and Zhang]{MIXGAN}
Sanjeev Arora, Rong Ge, Yingyu Liang, Tengyu Ma, and Yi~Zhang.
\newblock Generalization and equilibrium in generative adversarial nets (gans).
\newblock \emph{arXiv preprint arXiv:1703.00573}, 2017.

\bibitem[Barratt and Sharma(2018)]{barratt2018note}
Shane Barratt and Rishi Sharma.
\newblock A note on the inception score.
\newblock \emph{arXiv preprint arXiv:1801.01973}, 2018.

\bibitem[Bellemare et~al.(2017)Bellemare, Danihelka, Dabney, Mohamed,
  Lakshminarayanan, Hoyer, and Munos]{CramerGAN}
Marc~G Bellemare, Ivo Danihelka, Will Dabney, Shakir Mohamed, Balaji
  Lakshminarayanan, Stephan Hoyer, and R{\'e}mi Munos.
\newblock The cramer distance as a solution to biased wasserstein gradients.
\newblock \emph{arXiv preprint arXiv:1705.10743}, 2017.

\bibitem[Benaim and Wolf(2017)]{DistanceGAN}
Sagie Benaim and Lior Wolf.
\newblock One-sided unsupervised domain mapping.
\newblock \emph{arXiv preprint arXiv:1706.00826}, 2017.

\bibitem[Berthelot et~al.(2017)Berthelot, Schumm, and Metz]{BEGAN}
David Berthelot, Tom Schumm, and Luke Metz.
\newblock Began: Boundary equilibrium generative adversarial networks.
\newblock \emph{arXiv preprint arXiv:1703.10717}, 2017.

\bibitem[Bora et~al.(2018)Bora, Price, and Dimakis]{ambientgan}
Ashish Bora, Eric Price, and Alexandros~G. Dimakis.
\newblock Ambient{GAN}: Generative models from lossy measurements.
\newblock In \emph{International Conference on Learning Representations}, 2018.
\newblock URL \url{https://openreview.net/forum?id=Hy7fDog0b}.

\bibitem[Bousmalis et~al.(2016)Bousmalis, Silberman, Dohan, Erhan, and
  Krishnan]{Unsupervised}
Konstantinos Bousmalis, Nathan Silberman, David Dohan, Dumitru Erhan, and Dilip
  Krishnan.
\newblock Unsupervised pixel-level domain adaptation with generative
  adversarial networks.
\newblock \emph{arXiv preprint arXiv:1612.05424}, 2016.

\bibitem[Che et~al.(2016)Che, Li, Jacob, Bengio, and Li]{MRGAN}
Tong Che, Yanran Li, Athul~Paul Jacob, Yoshua Bengio, and Wenjie Li.
\newblock Mode regularized generative adversarial networks.
\newblock \emph{arXiv preprint arXiv:1612.02136}, 2016.

\bibitem[Che et~al.(2017)Che, Li, Zhang, Hjelm, Li, Song, and Bengio]{MALIGAN}
Tong Che, Yanran Li, Ruixiang Zhang, R~Devon Hjelm, Wenjie Li, Yangqiu Song,
  and Yoshua Bengio.
\newblock Maximum-likelihood augmented discrete generative adversarial
  networks.
\newblock \emph{arXiv preprint arXiv:1702.07983}, 2017.

\bibitem[Chen et~al.(2016)Chen, Duan, Houthooft, Schulman, Sutskever, and
  Abbeel]{InfoGAN}
Xi~Chen, Yan Duan, Rein Houthooft, John Schulman, Ilya Sutskever, and Pieter
  Abbeel.
\newblock Infogan: Interpretable representation learning by information
  maximizing generative adversarial nets.
\newblock In \emph{Advances in Neural Information Processing Systems}, pages
  2172--2180, 2016.

\bibitem[Dai et~al.(2017)Dai, Doyle, Liang, Zhang, Dong, Li, and Xing]{SCAN}
Wei Dai, Joseph Doyle, Xiaodan Liang, Hao Zhang, Nanqing Dong, Yuan Li, and
  Eric~P Xing.
\newblock Scan: Structure correcting adversarial network for chest x-rays organ
  segmentation.
\newblock \emph{arXiv preprint arXiv:1703.08770}, 2017.

\bibitem[Danihelka et~al.(2017)Danihelka, Lakshminarayanan, Uria, Wierstra, and
  Dayan]{independentw}
Ivo Danihelka, Balaji Lakshminarayanan, Benigno Uria, Daan Wierstra, and Peter
  Dayan.
\newblock Comparison of maximum likelihood and gan-based training of real nvps.
\newblock \emph{arXiv preprint arXiv:1705.05263}, 2017.

\bibitem[Dash et~al.(2017)Dash, Gamboa, Ahmed, Afzal, and Liwicki]{TAC-GAN}
Ayushman Dash, John Cristian~Borges Gamboa, Sheraz Ahmed, Muhammad~Zeshan
  Afzal, and Marcus Liwicki.
\newblock Tac-gan-text conditioned auxiliary classifier generative adversarial
  network.
\newblock \emph{arXiv preprint arXiv:1703.06412}, 2017.

\bibitem[Daskalakis et~al.(2018)Daskalakis, Ilyas, Syrgkanis, and
  Zeng]{optimism}
Constantinos Daskalakis, Andrew Ilyas, Vasilis Syrgkanis, and Haoyang Zeng.
\newblock Training {GAN}s with optimism.
\newblock In \emph{International Conference on Learning Representations}, 2018.
\newblock URL \url{https://openreview.net/forum?id=SJJySbbAZ}.

\bibitem[De~Maesschalck et~al.(2000)De~Maesschalck, Jouan-Rimbaud, and
  Massart]{maha}
Roy De~Maesschalck, Delphine Jouan-Rimbaud, and D{\'e}sir{\'e}~L Massart.
\newblock The mahalanobis distance.
\newblock \emph{Chemometrics and intelligent laboratory systems}, 50\penalty0
  (1):\penalty0 1--18, 2000.

\bibitem[Denton et~al.(2016)Denton, Gross, and Fergus]{ssl-gan}
Emily Denton, Sam Gross, and Rob Fergus.
\newblock Semi-supervised learning with context-conditional generative
  adversarial networks.
\newblock \emph{arXiv preprint arXiv:1611.06430}, 2016.

\bibitem[Doersch(2016)]{vae}
Carl Doersch.
\newblock Tutorial on variational autoencoders.
\newblock \emph{arXiv preprint arXiv:1606.05908}, 2016.

\bibitem[Donahue et~al.(2017)Donahue, Balsubramani, McAuley, and
  Lipton]{SD-GAN}
Chris Donahue, Akshay Balsubramani, Julian McAuley, and Zachary~C Lipton.
\newblock Semantically decomposing the latent spaces of generative adversarial
  networks.
\newblock \emph{arXiv preprint arXiv:1705.07904}, 2017.

\bibitem[Donahue et~al.(2016)Donahue, Kr{\"a}henb{\"u}hl, and Darrell]{BiGAN}
Jeff Donahue, Philipp Kr{\"a}henb{\"u}hl, and Trevor Darrell.
\newblock Adversarial feature learning.
\newblock \emph{arXiv preprint arXiv:1605.09782}, 2016.

\bibitem[Dumoulin and Visin(2016)]{transposedc}
Vincent Dumoulin and Francesco Visin.
\newblock A guide to convolution arithmetic for deep learning.
\newblock \emph{arXiv preprint arXiv:1603.07285}, 2016.

\bibitem[Dumoulin et~al.(2016)Dumoulin, Belghazi, Poole, Lamb, Arjovsky,
  Mastropietro, and Courville]{ALI}
Vincent Dumoulin, Ishmael Belghazi, Ben Poole, Alex Lamb, Martin Arjovsky,
  Olivier Mastropietro, and Aaron Courville.
\newblock Adversarially learned inference.
\newblock \emph{arXiv preprint arXiv:1606.00704}, 2016.

\bibitem[Durugkar et~al.(2016)Durugkar, Gemp, and
  Mahadevan]{durugkar2016generative}
Ishan Durugkar, Ian Gemp, and Sridhar Mahadevan.
\newblock Generative multi-adversarial networks.
\newblock \emph{arXiv preprint arXiv:1611.01673}, 2016.

\bibitem[Ehsani et~al.(2017)Ehsani, Mottaghi, and Farhadi]{SeGAN}
Kiana Ehsani, Roozbeh Mottaghi, and Ali Farhadi.
\newblock Segan: Segmenting and generating the invisible.
\newblock \emph{arXiv preprint arXiv:1703.10239}, 2017.

\bibitem[Fenchel(1949)]{fenchel}
Werner Fenchel.
\newblock On conjugate convex functions.
\newblock \emph{Canad. J. Math}, 1\penalty0 (73-77), 1949.

\bibitem[Finn et~al.(2016)Finn, Christiano, Abbeel, and Levine]{finnirl}
Chelsea Finn, Paul Christiano, Pieter Abbeel, and Sergey Levine.
\newblock A connection between generative adversarial networks, inverse
  reinforcement learning, and energy-based models.
\newblock \emph{arXiv preprint arXiv:1611.03852}, 2016.

\bibitem[Frankot and Chellappa(1988)]{frankot1988method}
Robert~T. Frankot and Rama Chellappa.
\newblock A method for enforcing integrability in shape from shading
  algorithms.
\newblock \emph{IEEE Transactions on pattern analysis and machine
  intelligence}, 10\penalty0 (4):\penalty0 439--451, 1988.

\bibitem[Gatys et~al.(2016)Gatys, Ecker, and Bethge]{styletransfer}
Leon~A Gatys, Alexander~S Ecker, and Matthias Bethge.
\newblock Image style transfer using convolutional neural networks.
\newblock In \emph{Computer Vision and Pattern Recognition (CVPR), 2016 IEEE
  Conference on}, pages 2414--2423. IEEE, 2016.

\bibitem[Ghosh et~al.(2017)Ghosh, Kulharia, Namboodiri, Torr, and
  Dokania]{MAD-GAN}
Arnab Ghosh, Viveka Kulharia, Vinay Namboodiri, Philip~HS Torr, and Puneet~K
  Dokania.
\newblock Multi-agent diverse generative adversarial networks.
\newblock \emph{arXiv preprint arXiv:1704.02906}, 2017.

\bibitem[Gomez et~al.(2018)Gomez, Huang, Zhang, Li, Osama, and
  Kaiser]{CipherGAN}
Aidan~N. Gomez, Sicong Huang, Ivan Zhang, Bryan~M. Li, Muhammad Osama, and
  Lukasz Kaiser.
\newblock Unsupervised cipher cracking using discrete {GAN}s.
\newblock In \emph{International Conference on Learning Representations}, 2018.
\newblock URL \url{https://openreview.net/forum?id=BkeqO7x0-}.

\bibitem[Goodfellow(2016)]{GOODFELLOWNIPS}
Ian Goodfellow.
\newblock Nips 2016 tutorial: Generative adversarial networks.
\newblock \emph{arXiv preprint arXiv:1701.00160}, 2016.

\bibitem[Goodfellow et~al.(2014)Goodfellow, Pouget-Abadie, Mirza, Xu,
  Warde-Farley, Ozair, Courville, and Bengio]{GAN}
Ian Goodfellow, Jean Pouget-Abadie, Mehdi Mirza, Bing Xu, David Warde-Farley,
  Sherjil Ozair, Aaron Courville, and Yoshua Bengio.
\newblock Generative adversarial nets.
\newblock In \emph{Advances in neural information processing systems}, pages
  2672--2680, 2014.

\bibitem[Gorijala and Dukkipati(2017)]{ViGAN}
Mahesh Gorijala and Ambedkar Dukkipati.
\newblock Image generation and editing with variational info generative
  adversarialnetworks.
\newblock \emph{arXiv preprint arXiv:1701.04568}, 2017.

\bibitem[Graves et~al.(2013)Graves, Mohamed, and Hinton]{rnn}
Alex Graves, Abdel-rahman Mohamed, and Geoffrey Hinton.
\newblock Speech recognition with deep recurrent neural networks.
\newblock In \emph{Acoustics, speech and signal processing (icassp), 2013 ieee
  international conference on}, pages 6645--6649. IEEE, 2013.

\bibitem[Grnarova et~al.(2017)Grnarova, Levy, Lucchi, Hofmann, and
  Krause]{chekhov}
Paulina Grnarova, Kfir~Y Levy, Aurelien Lucchi, Thomas Hofmann, and Andreas
  Krause.
\newblock An online learning approach to generative adversarial networks.
\newblock \emph{arXiv preprint arXiv:1706.03269}, 2017.

\bibitem[Gu et~al.(2016)Gu, Lillicrap, Ghahramani, Turner, and Levine]{qprop}
Shixiang Gu, Timothy Lillicrap, Zoubin Ghahramani, Richard~E Turner, and Sergey
  Levine.
\newblock Q-prop: Sample-efficient policy gradient with an off-policy critic.
\newblock \emph{arXiv preprint arXiv:1611.02247}, 2016.

\bibitem[Guimaraes et~al.(2017)Guimaraes, Sanchez-Lengeling, Farias, and
  Aspuru-Guzik]{ORGAN}
Gabriel~Lima Guimaraes, Benjamin Sanchez-Lengeling, Pedro Luis~Cunha Farias,
  and Al{\'a}n Aspuru-Guzik.
\newblock Objective-reinforced generative adversarial networks (organ) for
  sequence generation models.
\newblock \emph{arXiv preprint arXiv:1705.10843}, 2017.

\bibitem[Gulrajani et~al.(2017)Gulrajani, Ahmed, Arjovsky, Dumoulin, and
  Courville]{WGAN-GP}
Ishaan Gulrajani, Faruk Ahmed, Martin Arjovsky, Vincent Dumoulin, and Aaron
  Courville.
\newblock Improved training of wasserstein gans.
\newblock \emph{arXiv preprint arXiv:1704.00028}, 2017.

\bibitem[Hanin(1992)]{kantorovich}
Leonid~G Hanin.
\newblock Kantorovich-rubinstein norm and its application in the theory of
  lipschitz spaces.
\newblock \emph{Proceedings of the American Mathematical Society}, 115\penalty0
  (2):\penalty0 345--352, 1992.

\bibitem[Hjelm et~al.(2017)Hjelm, Jacob, Che, Cho, and Bengio]{BSGAN}
R~Devon Hjelm, Athul~Paul Jacob, Tong Che, Kyunghyun Cho, and Yoshua Bengio.
\newblock Boundary-seeking generative adversarial networks.
\newblock \emph{arXiv preprint arXiv:1702.08431}, 2017.

\bibitem[Ho and Ermon(2016)]{jonathanirl}
Jonathan Ho and Stefano Ermon.
\newblock Generative adversarial imitation learning.
\newblock In \emph{Advances in Neural Information Processing Systems}, pages
  4565--4573, 2016.

\bibitem[Hochreiter and Schmidhuber(1997)]{lstm}
Sepp Hochreiter and J{\"u}rgen Schmidhuber.
\newblock Long short-term memory.
\newblock \emph{Neural computation}, 9\penalty0 (8):\penalty0 1735--1780, 1997.

\bibitem[Hoffman et~al.(2017)Hoffman, Tzeng, Park, Zhu, Isola, Saenko, Efros,
  and Darrell]{CyCADA}
Judy Hoffman, Eric Tzeng, Taesung Park, Jun-Yan Zhu, Phillip Isola, Kate
  Saenko, Alexei~A Efros, and Trevor Darrell.
\newblock Cycada: Cycle-consistent adversarial domain adaptation.
\newblock \emph{arXiv preprint arXiv:1711.03213}, 2017.

\bibitem[Hsu et~al.(2017)Hsu, Hwang, Wu, Tsao, and Wang]{VAW-GAN}
Chin-Cheng Hsu, Hsin-Te Hwang, Yi-Chiao Wu, Yu~Tsao, and Hsin-Min Wang.
\newblock Voice conversion from unaligned corpora using variational
  autoencoding wasserstein generative adversarial networks.
\newblock \emph{arXiv preprint arXiv:1704.00849}, 2017.

\bibitem[Huang et~al.(2016)Huang, Li, Poursaeed, Hopcroft, and
  Belongie]{StackedGAN}
Xun Huang, Yixuan Li, Omid Poursaeed, John Hopcroft, and Serge Belongie.
\newblock Stacked generative adversarial networks.
\newblock \emph{arXiv:1612.04357}, 2016.
\newblock URL \url{https://arxiv.org/abs/1612.04357}.

\bibitem[Huang et~al.(2017)Huang, Li, Poursaeed, Hopcroft, and
  Belongie]{huang2017stacked}
Xun Huang, Yixuan Li, Omid Poursaeed, John Hopcroft, and Serge Belongie.
\newblock Stacked generative adversarial networks.
\newblock In \emph{IEEE Conference on Computer Vision and Pattern Recognition
  (CVPR)}, volume~2, page~4, 2017.

\bibitem[Im et~al.(2016)Im, Kim, Jiang, and Memisevic]{GRAN}
Daniel~Jiwoong Im, Chris~Dongjoo Kim, Hui Jiang, and Roland Memisevic.
\newblock Generating images with recurrent adversarial networks.
\newblock \emph{arXiv preprint arXiv:1602.05110}, 2016.

\bibitem[Isola et~al.(2016)Isola, Zhu, Zhou, and Efros]{pix2pix}
Phillip Isola, Jun-Yan Zhu, Tinghui Zhou, and Alexei~A Efros.
\newblock Image-to-image translation with conditional adversarial networks.
\newblock \emph{arXiv preprint arXiv:1611.07004}, 2016.

\bibitem[Jaynes(1957)]{jaynes}
Edwin~T Jaynes.
\newblock Information theory and statistical mechanics.
\newblock \emph{Physical review}, 106\penalty0 (4):\penalty0 620, 1957.

\bibitem[Juefei-Xu et~al.(2017)Juefei-Xu, Boddeti, and Savvides]{GoGAN}
Felix Juefei-Xu, Vishnu~Naresh Boddeti, and Marios Savvides.
\newblock Gang of gans: Generative adversarial networks with maximum margin
  ranking.
\newblock \emph{arXiv preprint arXiv:1704.04865}, 2017.

\bibitem[Karacan et~al.(2016)Karacan, Akata, Erdem, and Erdem]{AL-CGAN}
Levent Karacan, Zeynep Akata, Aykut Erdem, and Erkut Erdem.
\newblock Learning to generate images of outdoor scenes from attributes and
  semantic layouts.
\newblock \emph{arXiv preprint arXiv:1612.00215}, 2016.

\bibitem[Karras et~al.(2017)Karras, Aila, Laine, and
  Lehtinen]{karras2017progressive}
Tero Karras, Timo Aila, Samuli Laine, and Jaakko Lehtinen.
\newblock Progressive growing of gans for improved quality, stability, and
  variation.
\newblock \emph{arXiv preprint arXiv:1710.10196}, 2017.

\bibitem[Kim et~al.(2017{\natexlab{a}})Kim, Cha, Kim, Lee, and Kim]{DiscoGAN}
Taeksoo Kim, Moonsu Cha, Hyunsoo Kim, Jungkwon Lee, and Jiwon Kim.
\newblock Learning to discover cross-domain relations with generative
  adversarial networks.
\newblock \emph{arXiv preprint arXiv:1703.05192}, 2017{\natexlab{a}}.

\bibitem[Kim et~al.(2017{\natexlab{b}})Kim, Kim, Cha, and Kim]{conditiondisco}
Taeksoo Kim, Byoungjip Kim, Moonsu Cha, and Jiwon Kim.
\newblock Unsupervised visual attribute transfer with reconfigurable generative
  adversarial networks.
\newblock \emph{arXiv preprint arXiv:1707.09798}, 2017{\natexlab{b}}.

\bibitem[Kim et~al.(2017{\natexlab{c}})Kim, Zhang, Rush, LeCun, et~al.]{ARAE}
Yoon Kim, Kelly Zhang, Alexander~M Rush, Yann LeCun, et~al.
\newblock Adversarially regularized autoencoders for generating discrete
  structures.
\newblock \emph{arXiv preprint arXiv:1706.04223}, 2017{\natexlab{c}}.

\bibitem[Kocaoglu et~al.(2017)Kocaoglu, Snyder, Dimakis, and
  Vishwanath]{CausalGAN}
Murat Kocaoglu, Christopher Snyder, Alexandros~G Dimakis, and Sriram
  Vishwanath.
\newblock Causalgan: Learning causal implicit generative models with
  adversarial training.
\newblock \emph{arXiv preprint arXiv:1709.02023}, 2017.

\bibitem[Kodali et~al.(2017)Kodali, Abernethy, Hays, and Kira]{DRAGAN}
Naveen Kodali, Jacob Abernethy, James Hays, and Zsolt Kira.
\newblock How to train your dragan.
\newblock \emph{arXiv preprint arXiv:1705.07215}, 2017.

\bibitem[Krizhevsky et~al.(2012{\natexlab{a}})Krizhevsky, Sutskever, and
  Hinton]{cnn}
Alex Krizhevsky, Ilya Sutskever, and Geoffrey~E Hinton.
\newblock Imagenet classification with deep convolutional neural networks.
\newblock In \emph{Advances in neural information processing systems}, pages
  1097--1105, 2012{\natexlab{a}}.

\bibitem[Krizhevsky et~al.(2012{\natexlab{b}})Krizhevsky, Sutskever, and
  Hinton]{krizhevsky2012imagenet}
Alex Krizhevsky, Ilya Sutskever, and Geoffrey~E Hinton.
\newblock Imagenet classification with deep convolutional neural networks.
\newblock In \emph{Advances in neural information processing systems}, pages
  1097--1105, 2012{\natexlab{b}}.

\bibitem[Larsen et~al.(2015)Larsen, S{\o}nderby, Larochelle, and
  Winther]{VAEGAN}
Anders Boesen~Lindbo Larsen, S{\o}ren~Kaae S{\o}nderby, Hugo Larochelle, and
  Ole Winther.
\newblock Autoencoding beyond pixels using a learned similarity metric.
\newblock \emph{arXiv preprint arXiv:1512.09300}, 2015.

\bibitem[Ledig et~al.(2016)Ledig, Theis, Husz{\'a}r, Caballero, Cunningham,
  Acosta, Aitken, Tejani, Totz, Wang, et~al.]{SRGAN}
Christian Ledig, Lucas Theis, Ferenc Husz{\'a}r, Jose Caballero, Andrew
  Cunningham, Alejandro Acosta, Andrew Aitken, Alykhan Tejani, Johannes Totz,
  Zehan Wang, et~al.
\newblock Photo-realistic single image super-resolution using a generative
  adversarial network.
\newblock \emph{arXiv preprint arXiv:1609.04802}, 2016.

\bibitem[Lee et~al.(2017)Lee, Hwang, Min, and Yoon]{leeseqgan}
Sang-gil Lee, Uiwon Hwang, Seonwoo Min, and Sungroh Yoon.
\newblock A seqgan for polyphonic music generation.
\newblock \emph{arXiv preprint arXiv:1710.11418}, 2017.

\bibitem[Li et~al.(2017{\natexlab{a}})Li, Xu, Zhu, and Zhang]{triple}
Chongxuan Li, Kun Xu, Jun Zhu, and Bo~Zhang.
\newblock Triple generative adversarial nets.
\newblock \emph{arXiv preprint arXiv:1703.02291}, 2017{\natexlab{a}}.

\bibitem[Li et~al.(2017{\natexlab{b}})Li, Chang, Cheng, Yang, and
  P{\'o}czos]{MMDGAN}
Chun-Liang Li, Wei-Cheng Chang, Yu~Cheng, Yiming Yang, and Barnab{\'a}s
  P{\'o}czos.
\newblock Mmd gan: Towards deeper understanding of moment matching network.
\newblock \emph{arXiv preprint arXiv:1705.08584}, 2017{\natexlab{b}}.

\bibitem[Li et~al.(2017{\natexlab{c}})Li, Liang, Wei, Xu, Feng, and
  Yan]{Perceptual}
Jianan Li, Xiaodan Liang, Yunchao Wei, Tingfa Xu, Jiashi Feng, and Shuicheng
  Yan.
\newblock Perceptual generative adversarial networks for small object
  detection.
\newblock In \emph{IEEE CVPR}, 2017{\natexlab{c}}.

\bibitem[Li et~al.(2015)Li, Swersky, and Zemel]{GMMN}
Yujia Li, Kevin Swersky, and Rich Zemel.
\newblock Generative moment matching networks.
\newblock In \emph{Proceedings of the 32nd International Conference on Machine
  Learning (ICML-15)}, pages 1718--1727, 2015.

\bibitem[Lillicrap et~al.(2015)Lillicrap, Hunt, Pritzel, Heess, Erez, Tassa,
  Silver, and Wierstra]{ddpg}
Timothy~P Lillicrap, Jonathan~J Hunt, Alexander Pritzel, Nicolas Heess, Tom
  Erez, Yuval Tassa, David Silver, and Daan Wierstra.
\newblock Continuous control with deep reinforcement learning.
\newblock \emph{arXiv preprint arXiv:1509.02971}, 2015.

\bibitem[Lim and Ye(2017)]{GeometricGAN}
Jae~Hyun Lim and Jong~Chul Ye.
\newblock Geometric gan.
\newblock \emph{arXiv preprint arXiv:1705.02894}, 2017.

\bibitem[Lin et~al.(2017)Lin, Li, He, Zhang, and Sun]{RANKGAN}
Kevin Lin, Dianqi Li, Xiaodong He, Zhengyou Zhang, and Ming-Ting Sun.
\newblock Adversarial ranking for language generation.
\newblock \emph{arXiv preprint arXiv:1705.11001}, 2017.

\bibitem[Liu and Tuzel(2016)]{CoupledGAN}
Ming-Yu Liu and Oncel Tuzel.
\newblock Coupled generative adversarial networks.
\newblock In \emph{Advances in neural information processing systems}, pages
  469--477, 2016.

\bibitem[Lu et~al.(2017)Lu, Tai, and Tang]{conditioncycle}
Yongyi Lu, Yu-Wing Tai, and Chi-Keung Tang.
\newblock Conditional cyclegan for attribute guided face image generation.
\newblock \emph{arXiv preprint arXiv:1705.09966}, 2017.

\bibitem[Mao et~al.(2016)Mao, Li, Xie, Lau, Wang, and Smolley]{LSGAN}
Xudong Mao, Qing Li, Haoran Xie, Raymond~YK Lau, Zhen Wang, and Stephen~Paul
  Smolley.
\newblock Least squares generative adversarial networks.
\newblock \emph{arXiv preprint ArXiv:1611.04076}, 2016.

\bibitem[Mardani et~al.(2017)Mardani, Gong, Cheng, Vasanawala, Zaharchuk,
  Alley, Thakur, Han, Dally, Pauly, et~al.]{GANCS}
Morteza Mardani, Enhao Gong, Joseph~Y Cheng, Shreyas Vasanawala, Greg
  Zaharchuk, Marcus Alley, Neil Thakur, Song Han, William Dally, John~M Pauly,
  et~al.
\newblock Deep generative adversarial networks for compressed sensing automates
  mri.
\newblock \emph{arXiv preprint arXiv:1706.00051}, 2017.

\bibitem[Mescheder et~al.(2017)Mescheder, Nowozin, and Geiger]{numerics}
Lars Mescheder, Sebastian Nowozin, and Andreas Geiger.
\newblock The numerics of gans.
\newblock \emph{arXiv preprint arXiv:1705.10461}, 2017.

\bibitem[Metz et~al.(2016)Metz, Poole, Pfau, and Sohl-Dickstein]{UnrolledGAN}
Luke Metz, Ben Poole, David Pfau, and Jascha Sohl-Dickstein.
\newblock Unrolled generative adversarial networks.
\newblock \emph{arXiv preprint arXiv:1611.02163}, 2016.

\bibitem[Mirza and Osindero(2014)]{CGAN}
Mehdi Mirza and Simon Osindero.
\newblock Conditional generative adversarial nets.
\newblock \emph{arXiv preprint arXiv:1411.1784}, 2014.

\bibitem[Miyato and Koyama(2018)]{cganICLR}
Takeru Miyato and Masanori Koyama.
\newblock cgans with projection discriminator.
\newblock \emph{arXiv preprint arXiv:1802.05637}, 2018.

\bibitem[Miyato et~al.(2018)Miyato, Kataoka, Koyama, and
  Yoshida]{miyato2018spectral}
Takeru Miyato, Toshiki Kataoka, Masanori Koyama, and Yuichi Yoshida.
\newblock Spectral normalization for generative adversarial networks.
\newblock \emph{arXiv preprint arXiv:1802.05957}, 2018.

\bibitem[Mogren(2016)]{C-RNN-GAN}
Olof Mogren.
\newblock C-rnn-gan: Continuous recurrent neural networks with adversarial
  training.
\newblock \emph{arXiv preprint arXiv:1611.09904}, 2016.

\bibitem[Mroueh and Sercu(2017)]{FisherGAN}
Youssef Mroueh and Tom Sercu.
\newblock Fisher gan.
\newblock \emph{arXiv preprint arXiv:1705.09675}, 2017.

\bibitem[Mroueh et~al.(2017)Mroueh, Sercu, and Goel]{McGAN}
Youssef Mroueh, Tom Sercu, and Vaibhava Goel.
\newblock Mcgan: Mean and covariance feature matching gan.
\newblock \emph{arXiv preprint arXiv:1702.08398}, 2017.

\bibitem[Narayanan and Mitter(2010)]{sample}
Hariharan Narayanan and Sanjoy Mitter.
\newblock Sample complexity of testing the manifold hypothesis.
\newblock In \emph{Advances in Neural Information Processing Systems}, pages
  1786--1794, 2010.

\bibitem[Nguyen et~al.(2016{\natexlab{a}})Nguyen, Dosovitskiy, Yosinski, Brox,
  and Clune]{dcmam}
Anh Nguyen, Alexey Dosovitskiy, Jason Yosinski, Thomas Brox, and Jeff Clune.
\newblock Synthesizing the preferred inputs for neurons in neural networks via
  deep generator networks.
\newblock In \emph{Advances in Neural Information Processing Systems}, pages
  3387--3395, 2016{\natexlab{a}}.

\bibitem[Nguyen et~al.(2016{\natexlab{b}})Nguyen, Yosinski, Bengio,
  Dosovitskiy, and Clune]{PPGN}
Anh Nguyen, Jason Yosinski, Yoshua Bengio, Alexey Dosovitskiy, and Jeff Clune.
\newblock Plug \& play generative networks: Conditional iterative generation of
  images in latent space.
\newblock \emph{arXiv preprint arXiv:1612.00005}, 2016{\natexlab{b}}.

\bibitem[Nowozin et~al.(2016)Nowozin, Cseke, and Tomioka]{f-GAN}
Sebastian Nowozin, Botond Cseke, and Ryota Tomioka.
\newblock f-gan: Training generative neural samplers using variational
  divergence minimization.
\newblock In \emph{Advances in Neural Information Processing Systems}, pages
  271--279, 2016.

\bibitem[Odena et~al.(2016)Odena, Olah, and Shlens]{AC-GAN}
Augustus Odena, Christopher Olah, and Jonathon Shlens.
\newblock Conditional image synthesis with auxiliary classifier gans.
\newblock \emph{arXiv preprint arXiv:1610.09585}, 2016.

\bibitem[Oord et~al.(2016{\natexlab{a}})Oord, Dieleman, Zen, Simonyan, Vinyals,
  Graves, Kalchbrenner, Senior, and Kavukcuoglu]{Wavenet}
Aaron van~den Oord, Sander Dieleman, Heiga Zen, Karen Simonyan, Oriol Vinyals,
  Alex Graves, Nal Kalchbrenner, Andrew Senior, and Koray Kavukcuoglu.
\newblock Wavenet: A generative model for raw audio.
\newblock \emph{arXiv preprint arXiv:1609.03499}, 2016{\natexlab{a}}.

\bibitem[Oord et~al.(2016{\natexlab{b}})Oord, Kalchbrenner, and
  Kavukcuoglu]{pixelrnn}
Aaron van~den Oord, Nal Kalchbrenner, and Koray Kavukcuoglu.
\newblock Pixel recurrent neural networks.
\newblock \emph{arXiv preprint arXiv:1601.06759}, 2016{\natexlab{b}}.

\bibitem[Pan and Yang(2010)]{pan2010survey}
Sinno~Jialin Pan and Qiang Yang.
\newblock A survey on transfer learning.
\newblock \emph{IEEE Transactions on knowledge and data engineering},
  22\penalty0 (10):\penalty0 1345--1359, 2010.

\bibitem[Perarnau et~al.(2016)Perarnau, van~de Weijer, Raducanu, and
  {\'A}lvarez]{ICGAN}
Guim Perarnau, Joost van~de Weijer, Bogdan Raducanu, and Jose~M {\'A}lvarez.
\newblock Invertible conditional gans for image editing.
\newblock \emph{arXiv preprint arXiv:1611.06355}, 2016.

\bibitem[P{\'e}rez et~al.(2003)P{\'e}rez, Gangnet, and Blake]{poisson}
Patrick P{\'e}rez, Michel Gangnet, and Andrew Blake.
\newblock Poisson image editing.
\newblock In \emph{ACM Transactions on graphics (TOG)}, volume~22, pages
  313--318. ACM, 2003.

\bibitem[Pestov et~al.(2008)Pestov, Wang, Ariunbold, Murawski, Sautenkov,
  Dogariu, Sokolov, and Scully]{ssd}
Dmitry Pestov, Xi~Wang, Gombojav~O Ariunbold, Robert~K Murawski, Vladimir~A
  Sautenkov, Arthur Dogariu, Alexei~V Sokolov, and Marlan~O Scully.
\newblock Single-shot detection of bacterial endospores via coherent raman
  spectroscopy.
\newblock \emph{Proceedings of the National Academy of Sciences}, 105\penalty0
  (2):\penalty0 422--427, 2008.

\bibitem[Pfau and Vinyals(2016)]{connectrl}
David Pfau and Oriol Vinyals.
\newblock Connecting generative adversarial networks and actor-critic methods.
\newblock \emph{arXiv preprint arXiv:1610.01945}, 2016.

\bibitem[Qi(2017)]{LS-GAN}
Guo-Jun Qi.
\newblock Loss-sensitive generative adversarial networks on lipschitz
  densities.
\newblock \emph{arXiv preprint arXiv:1701.06264}, 2017.

\bibitem[Rachev et~al.(1990)]{kr}
Svetlozar~Todorov Rachev et~al.
\newblock Duality theorems for kantorovich-rubinstein and wasserstein
  functionals.
\newblock 1990.

\bibitem[Radford et~al.(2015)Radford, Metz, and Chintala]{DCGAN}
Alec Radford, Luke Metz, and Soumith Chintala.
\newblock Unsupervised representation learning with deep convolutional
  generative adversarial networks.
\newblock \emph{arXiv preprint arXiv:1511.06434}, 2015.

\bibitem[Redmon et~al.(2016)Redmon, Divvala, Girshick, and Farhadi]{yolo}
Joseph Redmon, Santosh Divvala, Ross Girshick, and Ali Farhadi.
\newblock You only look once: Unified, real-time object detection.
\newblock In \emph{Proceedings of the IEEE conference on computer vision and
  pattern recognition}, pages 779--788, 2016.

\bibitem[Rosca et~al.(2017)Rosca, Lakshminarayanan, Warde-Farley, and
  Mohamed]{alpha-GAN}
Mihaela Rosca, Balaji Lakshminarayanan, David Warde-Farley, and Shakir Mohamed.
\newblock Variational approaches for auto-encoding generative adversarial
  networks.
\newblock \emph{arXiv preprint arXiv:1706.04987}, 2017.

\bibitem[Rosenblatt(1956)]{clt}
Murray Rosenblatt.
\newblock A central limit theorem and a strong mixing condition.
\newblock \emph{Proceedings of the National Academy of Sciences}, 42\penalty0
  (1):\penalty0 43--47, 1956.

\bibitem[Salimans et~al.(2016)Salimans, Goodfellow, Zaremba, Cheung, Radford,
  and Chen]{Improved}
Tim Salimans, Ian Goodfellow, Wojciech Zaremba, Vicki Cheung, Alec Radford, and
  Xi~Chen.
\newblock Improved techniques for training gans.
\newblock In \emph{Advances in Neural Information Processing Systems}, pages
  2234--2242, 2016.

\bibitem[Salimans et~al.(2017)Salimans, Karpathy, Chen, and Kingma]{pixelcnn}
Tim Salimans, Andrej Karpathy, Xi~Chen, and Diederik~P Kingma.
\newblock Pixelcnn++: Improving the pixelcnn with discretized logistic mixture
  likelihood and other modifications.
\newblock \emph{arXiv preprint arXiv:1701.05517}, 2017.

\bibitem[Sch{\"o}lkopf and Smola(2002)]{SVM}
Bernhard Sch{\"o}lkopf and Alexander~J Smola.
\newblock \emph{Learning with kernels: support vector machines, regularization,
  optimization, and beyond}.
\newblock MIT press, 2002.

\bibitem[Schulman et~al.(2015)Schulman, Levine, Abbeel, Jordan, and
  Moritz]{trpo}
John Schulman, Sergey Levine, Pieter Abbeel, Michael Jordan, and Philipp
  Moritz.
\newblock Trust region policy optimization.
\newblock In \emph{Proceedings of the 32nd International Conference on Machine
  Learning (ICML-15)}, pages 1889--1897, 2015.

\bibitem[Shen et~al.(2017)Shen, Qu, Zhang, and Yu]{ARDA}
Jian Shen, Yanru Qu, Weinan Zhang, and Yong Yu.
\newblock Adversarial representation learning for domain adaptation.
\newblock \emph{arXiv preprint arXiv:1707.01217}, 2017.

\bibitem[Shi et~al.(2017)Shi, Dong, Wang, Qian, and Zhang]{SSGAN}
Haichao Shi, Jing Dong, Wei Wang, Yinlong Qian, and Xiaoyu Zhang.
\newblock Ssgan: Secure steganography based on generative adversarial networks.
\newblock \emph{arXiv preprint arXiv:1707.01613}, 2017.

\bibitem[Shin et~al.(2017)Shin, Lee, Kim, and Kim]{Continual}
Hanul Shin, Jung~Kwon Lee, Jaehong Kim, and Jiwon Kim.
\newblock Continual learning with deep generative replay.
\newblock In \emph{Advances in Neural Information Processing Systems}, pages
  2990--2999, 2017.

\bibitem[Shu et~al.(2018)Shu, Bui, Narui, and Ermon]{dirt-t}
Rui Shu, Hung~H Bui, Hirokazu Narui, and Stefano Ermon.
\newblock A dirt-t approach to unsupervised domain adaptation.
\newblock \emph{arXiv preprint arXiv:1802.08735}, 2018.

\bibitem[Simonyan and Zisserman(2014)]{vgg}
Karen Simonyan and Andrew Zisserman.
\newblock Very deep convolutional networks for large-scale image recognition.
\newblock \emph{arXiv preprint arXiv:1409.1556}, 2014.

\bibitem[Smolensky(1986)]{smolensky1986information}
Paul Smolensky.
\newblock Information processing in dynamical systems: Foundations of harmony
  theory.
\newblock Technical report, COLORADO UNIV AT BOULDER DEPT OF COMPUTER SCIENCE,
  1986.

\bibitem[Song et~al.(2018)Song, Ma, Wu, Gong, Bao, Zuo, Shen, Lau, and
  Yang]{song2018vital}
Yibing Song, Chao Ma, Xiaohe Wu, Lijun Gong, Linchao Bao, Wangmeng Zuo, Chunhua
  Shen, Rynson Lau, and Ming-Hsuan Yang.
\newblock Vital: Visual tracking via adversarial learning.
\newblock \emph{arXiv preprint arXiv:1804.04273}, 2018.

\bibitem[Springenberg(2015)]{CatGAN}
Jost~Tobias Springenberg.
\newblock Unsupervised and semi-supervised learning with categorical generative
  adversarial networks.
\newblock \emph{arXiv preprint arXiv:1511.06390}, 2015.

\bibitem[Spurr et~al.(2017)Spurr, Aksan, and Hilliges]{ss-InfoGAN}
Adrian Spurr, Emre Aksan, and Otmar Hilliges.
\newblock Guiding infogan with semi-supervision.
\newblock \emph{arXiv preprint arXiv:1707.04487}, 2017.

\bibitem[Sriperumbudur et~al.(2009)Sriperumbudur, Fukumizu, Gretton,
  Sch{\"o}lkopf, and Lanckriet]{IPMf-divergence}
Bharath~K Sriperumbudur, Kenji Fukumizu, Arthur Gretton, Bernhard
  Sch{\"o}lkopf, and Gert~RG Lanckriet.
\newblock On integral probability metrics,$\backslash$phi-divergences and
  binary classification.
\newblock \emph{arXiv preprint arXiv:0901.2698}, 2009.

\bibitem[Sutton et~al.(2000)Sutton, McAllester, Singh, and Mansour]{PG}
Richard~S Sutton, David~A McAllester, Satinder~P Singh, and Yishay Mansour.
\newblock Policy gradient methods for reinforcement learning with function
  approximation.
\newblock In \emph{Advances in neural information processing systems}, pages
  1057--1063, 2000.

\bibitem[Theis et~al.(2015)Theis, Oord, and Bethge]{evaluation}
Lucas Theis, A{\"a}ron van~den Oord, and Matthias Bethge.
\newblock A note on the evaluation of generative models.
\newblock \emph{arXiv preprint arXiv:1511.01844}, 2015.

\bibitem[Torchinsky and Wang(1990)]{holder}
Alberto Torchinsky and Shilin Wang.
\newblock A note on the marcinkiewicz integral.
\newblock In \emph{Colloquium Mathematicae}, volume~1, pages 235--243, 1990.

\bibitem[Tran et~al.(2017)Tran, Yin, and Liu]{DR-GAN}
Luan Tran, Xi~Yin, and Xiaoming Liu.
\newblock Representation learning by rotating your faces.
\newblock \emph{arXiv preprint arXiv:1705.11136}, 2017.

\bibitem[Tulyakov et~al.(2017)Tulyakov, Liu, Yang, and Kautz]{MocoGAN}
Sergey Tulyakov, Ming-Yu Liu, Xiaodong Yang, and Jan Kautz.
\newblock Mocogan: Decomposing motion and content for video generation.
\newblock \emph{arXiv preprint arXiv:1707.04993}, 2017.

\bibitem[Ulyanov et~al.(2017)Ulyanov, Vedaldi, and Lempitsky]{AGE}
Dmitry Ulyanov, Andrea Vedaldi, and Victor Lempitsky.
\newblock Adversarial generator-encoder networks.
\newblock \emph{arXiv preprint arXiv:1704.02304}, 2017.

\bibitem[Volkhonskiy et~al.(2017)Volkhonskiy, Nazarov, Borisenko, and
  Burnaev]{SteganographyGAN}
Denis Volkhonskiy, Ivan Nazarov, Boris Borisenko, and Evgeny Burnaev.
\newblock Steganographic generative adversarial networks.
\newblock \emph{arXiv preprint arXiv:1703.05502}, 2017.

\bibitem[Vondrick et~al.(2016)Vondrick, Pirsiavash, and Torralba]{VGAN}
Carl Vondrick, Hamed Pirsiavash, and Antonio Torralba.
\newblock Generating videos with scene dynamics.
\newblock In \emph{Advances In Neural Information Processing Systems}, pages
  613--621, 2016.

\bibitem[Walker et~al.(2017)Walker, Marino, Gupta, and Hebert]{Pose-GAN}
Jacob Walker, Kenneth Marino, Abhinav Gupta, and Martial Hebert.
\newblock The pose knows: Video forecasting by generating pose futures.
\newblock \emph{arXiv preprint arXiv:1705.00053}, 2017.

\bibitem[Wang et~al.(2018)Wang, Xu, Wang, and Tao]{wang2018perceptual}
Chaoyue Wang, Chang Xu, Chaohui Wang, and Dacheng Tao.
\newblock Perceptual adversarial networks for image-to-image transformation.
\newblock \emph{IEEE Transactions on Image Processing}, 27\penalty0
  (8):\penalty0 4066--4079, 2018.

\bibitem[Wang et~al.(2017)Wang, Cully, Chang, and Demiris]{MAGAN}
Ruohan Wang, Antoine Cully, Hyung~Jin Chang, and Yiannis Demiris.
\newblock Magan: Margin adaptation for generative adversarial networks.
\newblock \emph{arXiv preprint arXiv:1704.03817}, 2017.

\bibitem[Wang et~al.(2004)Wang, Bovik, Sheikh, and Simoncelli]{msssim}
Zhou Wang, Alan~C Bovik, Hamid~R Sheikh, and Eero~P Simoncelli.
\newblock Image quality assessment: from error visibility to structural
  similarity.
\newblock \emph{IEEE transactions on image processing}, 13\penalty0
  (4):\penalty0 600--612, 2004.

\bibitem[Welling(2005)]{FLDA}
Max Welling.
\newblock Fisher linear discriminant analysis.
\newblock \emph{Department of Computer Science, University of Toronto},
  3\penalty0 (1), 2005.

\bibitem[Wilson and Hilferty(1931)]{chi}
Edwin~B Wilson and Margaret~M Hilferty.
\newblock The distribution of chi-square.
\newblock \emph{Proceedings of the National Academy of Sciences}, 17\penalty0
  (12):\penalty0 684--688, 1931.

\bibitem[Wu et~al.(2017)Wu, Zheng, Zhang, and Huang]{GP-GAN}
Huikai Wu, Shuai Zheng, Junge Zhang, and Kaiqi Huang.
\newblock Gp-gan: Towards realistic high-resolution image blending.
\newblock \emph{arXiv preprint arXiv:1703.07195}, 2017.

\bibitem[Wu et~al.(2016)Wu, Zhang, Xue, Freeman, and Tenenbaum]{3DGAN}
Jiajun Wu, Chengkai Zhang, Tianfan Xue, Bill Freeman, and Josh Tenenbaum.
\newblock Learning a probabilistic latent space of object shapes via 3d
  generative-adversarial modeling.
\newblock In \emph{Advances in Neural Information Processing Systems}, pages
  82--90, 2016.

\bibitem[Xue et~al.(2017)Xue, Xu, Zhang, Long, and Huang]{Medical}
Yuan Xue, Tao Xu, Han Zhang, Rodney Long, and Xiaolei Huang.
\newblock Segan: Adversarial network with multi-scale $ l\_1 $ loss for medical
  image segmentation.
\newblock \emph{arXiv preprint arXiv:1706.01805}, 2017.

\bibitem[Yan et~al.(2016)Yan, Yang, Sohn, and Lee]{cvae}
Xinchen Yan, Jimei Yang, Kihyuk Sohn, and Honglak Lee.
\newblock Attribute2image: Conditional image generation from visual attributes.
\newblock In \emph{European Conference on Computer Vision}, pages 776--791.
  Springer, 2016.

\bibitem[Yang et~al.(2017)Yang, Xiong, Xu, Huang, Liu, Zhou, Xu, Park, Chen,
  Tran, et~al.]{DI2IN}
Dong Yang, Tao Xiong, Daguang Xu, Qiangui Huang, David Liu, S~Kevin Zhou,
  Zhoubing Xu, JinHyeong Park, Mingqing Chen, Trac~D Tran, et~al.
\newblock Automatic vertebra labeling in large-scale 3d ct using deep
  image-to-image network with message passing and sparsity regularization.
\newblock In \emph{International Conference on Information Processing in
  Medical Imaging}, pages 633--644. Springer, 2017.

\bibitem[Yi et~al.(2017)Yi, Zhang, Gong, et~al.]{DualGAN}
Zili Yi, Hao Zhang, Ping~Tan Gong, et~al.
\newblock Dualgan: Unsupervised dual learning for image-to-image translation.
\newblock \emph{arXiv preprint arXiv:1704.02510}, 2017.

\bibitem[Yin et~al.(2017)Yin, Fu, Sigal, and Xue]{SL-GAN}
Weidong Yin, Yanwei Fu, Leonid Sigal, and Xiangyang Xue.
\newblock Semi-latent gan: Learning to generate and modify facial images from
  attributes.
\newblock \emph{arXiv preprint arXiv:1704.02166}, 2017.

\bibitem[Yoo et~al.(2017{\natexlab{a}})Yoo, Ha, Yi, Ryu, Kim, Ha, Kim, and
  Yoon]{yooirl}
Jaeyoon Yoo, Heonseok Ha, Jihun Yi, Jongha Ryu, Chanju Kim, Jung-Woo Ha,
  Young-Han Kim, and Sungroh Yoon.
\newblock Energy-based sequence gans for recommendation and their connection to
  imitation learning.
\newblock \emph{arXiv preprint arXiv:1706.09200}, 2017{\natexlab{a}}.

\bibitem[Yoo et~al.(2017{\natexlab{b}})Yoo, Hong, and Yoon]{AUNDA}
Jaeyoon Yoo, Yongjun Hong, and Sungroh Yoon.
\newblock Autonomous uav navigation with domain adaptation.
\newblock \emph{arXiv preprint arXiv:1712.03742}, 2017{\natexlab{b}}.

\bibitem[Yu et~al.(2017)Yu, Zhang, Wang, and Yu]{SeqGAN}
Lantao Yu, Weinan Zhang, Jun Wang, and Yong Yu.
\newblock Seqgan: Sequence generative adversarial nets with policy gradient.
\newblock In \emph{AAAI}, pages 2852--2858, 2017.

\bibitem[Zhang et~al.(2016)Zhang, Xu, Li, Zhang, Huang, Wang, and
  Metaxas]{StackGAN}
Han Zhang, Tao Xu, Hongsheng Li, Shaoting Zhang, Xiaolei Huang, Xiaogang Wang,
  and Dimitris Metaxas.
\newblock Stackgan: Text to photo-realistic image synthesis with stacked
  generative adversarial networks.
\newblock \emph{arXiv preprint arXiv:1612.03242}, 2016.

\bibitem[Zhao et~al.(2016)Zhao, Mathieu, and LeCun]{EBGAN}
Junbo Zhao, Michael Mathieu, and Yann LeCun.
\newblock Energy-based generative adversarial network.
\newblock \emph{arXiv preprint arXiv:1609.03126}, 2016.

\bibitem[Zhou et~al.(2017)Zhou, Xiao, Yang, Feng, He, and He]{GeneGAN}
Shuchang Zhou, Taihong Xiao, Yi~Yang, Dieqiao Feng, Qinyao He, and Weiran He.
\newblock Genegan: Learning object transfiguration and attribute subspace from
  unpaired data.
\newblock \emph{arXiv preprint arXiv:1705.04932}, 2017.

\bibitem[Zhu et~al.(2017)Zhu, Park, Isola, and Efros]{CycleGAN}
Jun-Yan Zhu, Taesung Park, Phillip Isola, and Alexei~A Efros.
\newblock Unpaired image-to-image translation using cycle-consistent
  adversarial networks.
\newblock \emph{arXiv preprint arXiv:1703.10593}, 2017.

\bibitem[Ziebart et~al.(2008)Ziebart, Maas, Bagnell, and Dey]{maximum}
Brian~D Ziebart, Andrew~L Maas, J~Andrew Bagnell, and Anind~K Dey.
\newblock Maximum entropy inverse reinforcement learning.
\newblock In \emph{AAAI}, volume~8, pages 1433--1438. Chicago, IL, USA, 2008.

\end{thebibliography}

\end{document}